\documentclass[10pt,twocolumn,letterpaper]{article}

\usepackage[accsupp]{axessibility}
\usepackage[pagenumbers]{cvpr_toc} \usepackage{epsfig}
\usepackage[tight]{minitoc}
\usepackage{graphicx}
\usepackage{amsmath}
\usepackage{amssymb}
\usepackage{booktabs}
\usepackage{enumitem}
\usepackage{lib_cref}
\usepackage{xcolor,colortbl}
\usepackage{cite}
\usepackage{alias}
\usepackage{multicol}
\usepackage{nicefrac}
\usepackage{pifont}
\usepackage{balance}
\usepackage{wrapfig}

\noptcrule 

\mtcsetdepth{parttoc}{5}

\usepackage[labelfont=bf,font=small,skip=3pt]{caption}
\newcommand{\std}{$\pm$}

\usepackage[pagebackref=true,breaklinks=true,letterpaper=true,colorlinks,citecolor=blue,bookmarks=false,linkcolor=purple]{hyperref}

\usepackage[capitalize]{cleveref}
\Crefname{section}{Section}{Sections}
\Crefname{equation}{Eq.}{Eqns.}
\Crefname{table}{Table}{Tables}
\Crefname{figure}{Figure}{Figures}

\def\cvprPaperID{2974} \def\confName{CVPR}
\def\confYear{2022}

\newcommand\blfootnote[1]{\begingroup
  \renewcommand\thefootnote{}\footnote{#1}\addtocounter{footnote}{-1}\endgroup
}

\begin{document}
\doparttoc

\faketableofcontents

\title{Human Hands as Probes for Interactive Object Understanding}

\author{Mohit Goyal \qquad Sahil Modi \qquad Rishabh Goyal \qquad Saurabh Gupta \\
University of Illinois Urbana-Champaign \\
{\tt \small \{mohit, smodi9, rgoyal6, saurabhg\}@illinois.edu}
}

\maketitle

\blfootnote{Project website: \url{https://s-gupta.github.io/hands-as-probes/}.}

\begin{abstract}

Interactive object understanding, or what we can do to objects 
and how is a long-standing goal of computer vision.
In this paper, we tackle this problem through observation of human hands
in in-the-wild egocentric videos.
We demonstrate that observation of what human hands interact with and how can
provide both the relevant data and the necessary supervision.
Attending to hands, readily localizes and stabilizes active objects for
learning and reveals places where interactions with objects
occur. Analyzing the hands shows what we can do to objects and how.
We apply these basic principles on the \epic dataset, and
successfully learn state-sensitive features, and object
affordances (regions of interaction and afforded grasps), purely by observing
hands in egocentric videos.

 \end{abstract}

\section{Introduction}
\seclabel{intro}

Consider the cupboard in \figref{teaser}. Merely localizing and naming it is
insufficient for a robot to successfully interact with it. To enable
interaction, we need to identify what are plausible sites for interaction,
how should we interact with each site, and what would happen when we do.  The
goal of this paper is to acquire such an understanding about objects.
Specifically, we formulate it as
a) learning a feature space that is sensitive to the {\it state} of the
object (and thus indicative of what we can do with it) rather than just its
{\it category}; and b) identifying what hand-grasps do objects afford and
where. 
These together provide an interactive understanding of
objects, and could aid learning policies for robots. For instance,
distance in a state-sensitive feature space can be used as reward
functions for manipulation tasks~\cite{sermanet2016unsupervised,
zakka2021xirl, schmeckpeper2020reinforcement}.  Similarly, hand-grasps afforded
by objects and their locations provide priors for
exploration~\cite{mandikal2020dexterous, mandikal2021dexvip}.

While we have made large strides in building models for how objects look (the
various object recognition problems), the same recipe of collecting large-scale
labeled datasets for training doesn't quite apply for understanding how objects
work. First of all, no large-scale labeled datasets already exist for such 
tasks. 
Second, manually annotating these aspects on static images is challenging. 
For instance, objects states are highly contextual:
the same object (\eg cupboard in \figref{teaser}) can exist in many different
states (closed, full, on-top-of, has-handle, in-contact-with-hand) at the same
time, depending on the interaction we want to conduct.  
Similarly, consciously annotating where and how one can touch an object can
suffer from biases, leading to data that may not be indicative of how people
{\it actually} use objects during normal daily conduct. While one might
annotate that we pull on the handle to open the cupboard; in real life we may
very often just flick it open by sliding our fingers in between
the cupboard door and its frame.

\begin{figure}[t]
\centering
\insertWL{1.0}{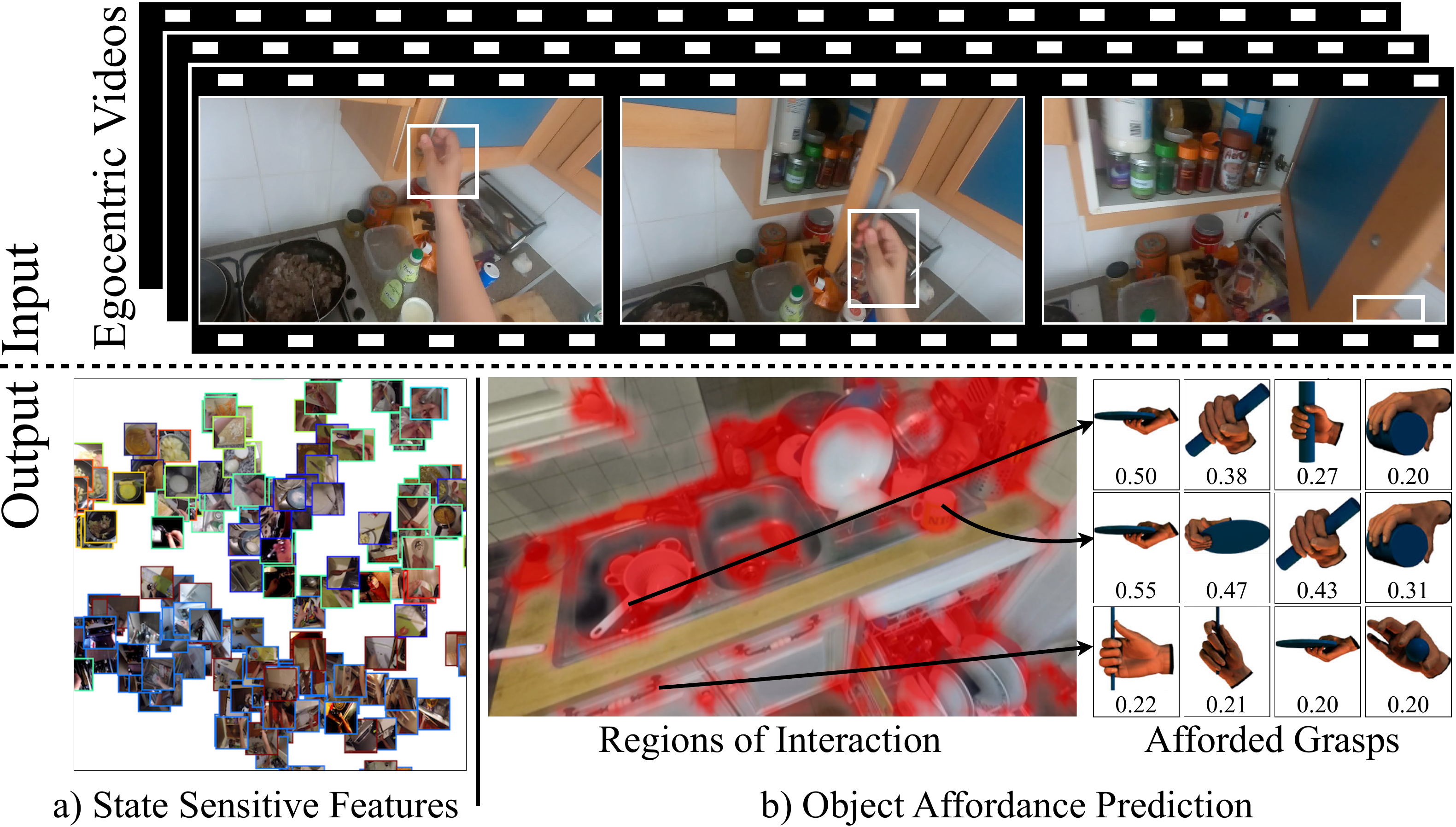}
\caption{Human hands reveal information about objects as they interact with
them. They tell us where and how we can interact with an object (the handle of
the cupboard via an adducted thumb grasp), and what happens when we do (cupboard
opens to reveal many more objects within). This paper develops techniques to
extract an interactive understanding of objects through the observation of 
hands in a corpus of egocentric videos. Specifically, we produce a)
features that are indicative of object states, and b) object affordances (\ie regions of interaction, and afforded grasps).}
\figlabel{teaser}
\end{figure}

Motivated by these challenges, we pursue learning directly 
from the {\it natural} ways in which people interact with objects in egocentric videos.
Since, egocentric data focuses upon hand-object interaction, it solves both the data and the supervision problem.
Egocentric observation of human hands reveals information about the objects they are interacting with. 
Attending to locations that hands attend to, localizes and stabilizes active 
objects in the scene for learning. 
It shows where all hands can interact in the scene. 
Analyzing what the hand is doing reveals information about the state of the
object, and also how to interact with it. 
Thus, observation of human hands in egocentric videos can provide the necessary data and supervision for obtaining an interactive understanding of objects.

To realize these intuitions, we design novel techniques that extract 
an understanding of objects from the understanding of hands as 
obtained from off-the-shelf models. We
apply this approach to the two aspects of interactive object understanding: a)
learning state-sensitive features, and b) inferring object affordances 
(identifying what hand-grasps do objects placed in scenes afford and where).

For the former goal of learning state-sensitive features, we {\it hand-stabilize}
the object-of-interaction. We exploit the appearance and motion of the hand
as it interacts with the object to derive supervision for the object state. This is done 
through contrastive learning where we encourage objects associated with similar 
hand appearance and motion, to be similar to one another.
This leads to features that are more state-sensitive than those obtained from alternate
forms of self-supervision, and even direct semantic supervision.

For the latter goal of predicting regions-of-interaction and applicable grasps,
we additionally use hand grasp-type predictions. As the hand is directly
visible when the interaction is happening, the challenge here is to 
get the model to focus on the object to make its predictions, rather than the
hand.
For this, we design a
context prediction task: we mask-out the hand and train a model to predict the
location and grasp-type from the surrounding context. We find that modern
models can successfully learn to make such contextual predictions. This enables
us to identify the places where humans interact in scenes. We better recall
small interaction sites such as knobs and handles, and also make more specific
predictions when interaction sites are localized to specific regions on the
objects (\eg knobs for stoves). We are also able to successfully learn
hand-grasps applicable to different objects.

For both these aspects, deriving supervision from hands sidesteps the need for
and possible pitfalls of semantic supervision. We are able to conduct learning
without having to define a complete taxonomy of object states, or suffer from
inherent ambiguity in defining action classes.

 \section{Related Work}
\seclabel{related}
We survey research on understanding human hands, using humans or
their hands as cues, interactive object understanding, and self-supervision. 

\noindent \textbf{Understanding hands.}
Several works have sought to build a data-driven understanding of
human hands and how they manipulate objects from \rgb
images~\cite{yang2015grasp}, \rgbd images~\cite{rogez2015understanding}, egocentric data~\cite{Kwon_2021_ICCV},
videos~\cite{garcia2018first} and other sensors~\cite{brahmbhatt2019contactdb, tekin2019h+}.
This understanding can take different forms: grasp type classification
\cite{yang2015grasp, rogez2015understanding, cai2016understanding} from a hand-defined
taxonomy~\cite{feix2015grasp}, hand keypoint and pose
estimation~\cite{garcia2018first},
understanding gestures~\cite{ginosar2019learning}, detecting hands, their
states and objects of interaction~\cite{shan2020understanding, shan2021contrastive}, 3D
reconstruction of the hand and the object of
interaction~\cite{hasson2019learning, cao2020reconstructing}, or even
estimating forces being applied by the hand onto the
object~\cite{ehsani2020use}.  We refer the reader to the survey paper from
Bandini and Zariffa~\cite{bandini2020analysis} for an analysis of hand
understanding in context of egocentric data. Our goals are different:
we build upon the understanding of hands to better understand objects.

\begin{figure*}
\centering
\insertWL{1.0}{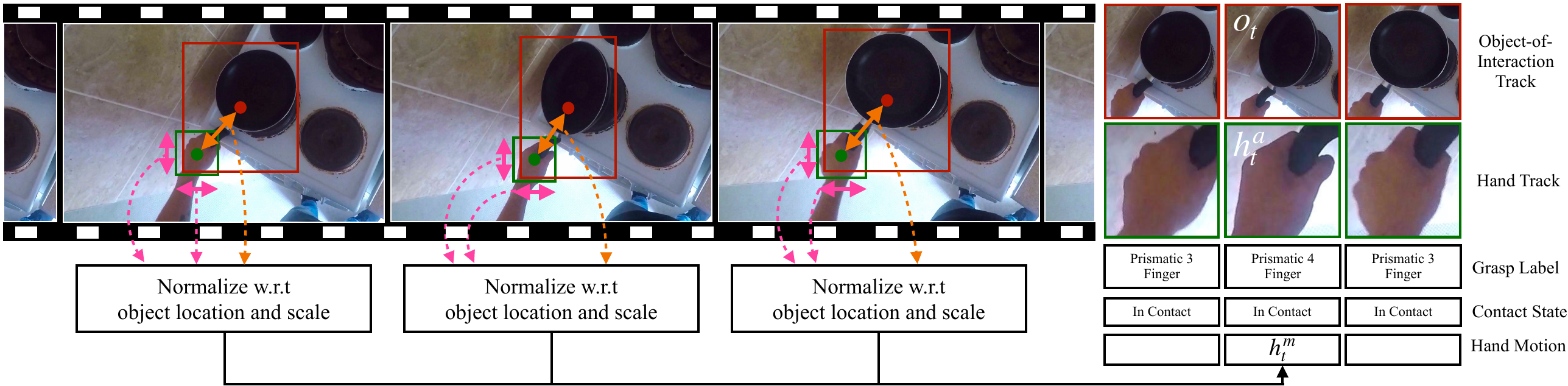}
\caption{\textbf{Data preparation.} 
Given egocentric videos from the \epic
dataset~\cite{damen2020collection}, we obtain per-frame detections for
hand, object-of-interaction, and contact-state from~\cite{shan2020understanding}.
These detections are strung together over time to form paired object-hand tracks.
We represent the motion of the hand $h^m_t$ around the object by stacking the hand box
location and scale relative to the object over 3 adjacent frames.
\textbf{Object-of-interaction
tracks, hand tracks and hand motion} are together used to learn state sensitive
feature spaces (\secref{state-features}). We also obtain hand grasp labels
through a classifier trained on the GUN-71 dataset~\cite{rogez2015understanding}. \textbf{The
detected hand and object-of-interaction pairs along with these hand grasp labels} are used for
learning regions of interaction and the grasps afforded by these regions
(\secref{hotspots}).}
\figlabel{data-gen}
\end{figure*}
 
\noindent \textbf{Using humans or their hands as probes.} The most relevant
research to our work is that of using humans and hands as probes for
understanding objects, scenes and other humans. \cite{fouhey2012people,wang2017binge, sun2020hidden} learn about scene affordance by watching how
people interact with scenes in videos from YouTube, sitcoms and self-driving
cars. Brahmbhatt \etal~\cite{brahmbhatt2019contactdb} learn 
task-oriented grasping regions by analyzing where people touch
objects using thermal imaging. Wang \etal~\cite{Wang2020novelobj} use humans as
visual cues for detecting novel objects. Mandikal and
Grauman~\cite{mandikal2020dexterous} extend work from
\cite{brahmbhatt2019contactdb} to learn policies for object
manipulation using predicted contact regions. Ng \etal~\cite{ng2020you2me}
use body pose of another person to predict the self-pose in egocentric videos. 
Unlike these past works, we focus upon observation of hands (and not full
humans) in unscripted in-the-wild \rgb egocentric videos (rather than in lab or
with specialized sensors), to learn fine-grained aspects of object affordance
(rather than scene affordance).
Concurrent work from Nagarajan \etal~\cite{nagarajan2021shaping} works in
a similar setting but focuses on learning activity-context priors.

\noindent \textbf{Interactive Object Understanding.}
Observing hands interact with objects is not the only way to learn about how to
interact with objects. Researchers have used other forms of supervision (strong
supervision, weak supervision, imitation learning, reinforcement learning,
inverse reinforcement learning) to build interactive understanding of objects. This can be in the form of learning 
a) where and how to grasp~\cite{Mahler2017DexNet2D, jiang2021graspTTA, mousavian2019graspnet, 
corona2020ganhand, karunratanakul2020graspingFL, hamer2010handposeprior, sergey2018handeye,
pinto2016SupersizingSL, lenz2015robgrasp, Kokic2020LearningTG},
b) state classifiers~\cite{isola2015discovering}, 
c) interaction hotspots~\cite{nagarajan2019grounded, demo2vec, thermos2021affordancevideo, mo2021where2act},
d) spatial priors for action sites~\cite{nagarajan2020ego},
e) object articulation modes~\cite{li2020category, damen2016you},
f) reward functions~\cite{kitani2012activity, rhinehart2017first, petrik2020learning, koppula2014physically},
g) functional correspondences~\cite{lai2021functional}.
While our work pursues similar goals, we differ in our supervision source
(observation of human hands interacting with objects in egocentric videos).

\noindent \textbf{Self-supervision.} Our techniques are inspired by work in
self-supervision where the goal is to learn without semantic labels~\cite{doersch2015unsupervised, chen2020improved, Chen2021ExploringSS, 
grill2020byol, caron2020unsupervised, zhang2017split}. 
Specifically, our work builds upon recent use of 
context prediction~\cite{doersch2015unsupervised, pathakCVPR16context} 
and contrastive learning~\cite{chen2020simple, sermanet2018time} 
for self-supervision. We design novel sources of supervision in the context of
egocentric videos to enable interactive object understanding.

 \section{Approach}
\seclabel{method}

\begin{figure*}
\centering
\insertWL{1.0}{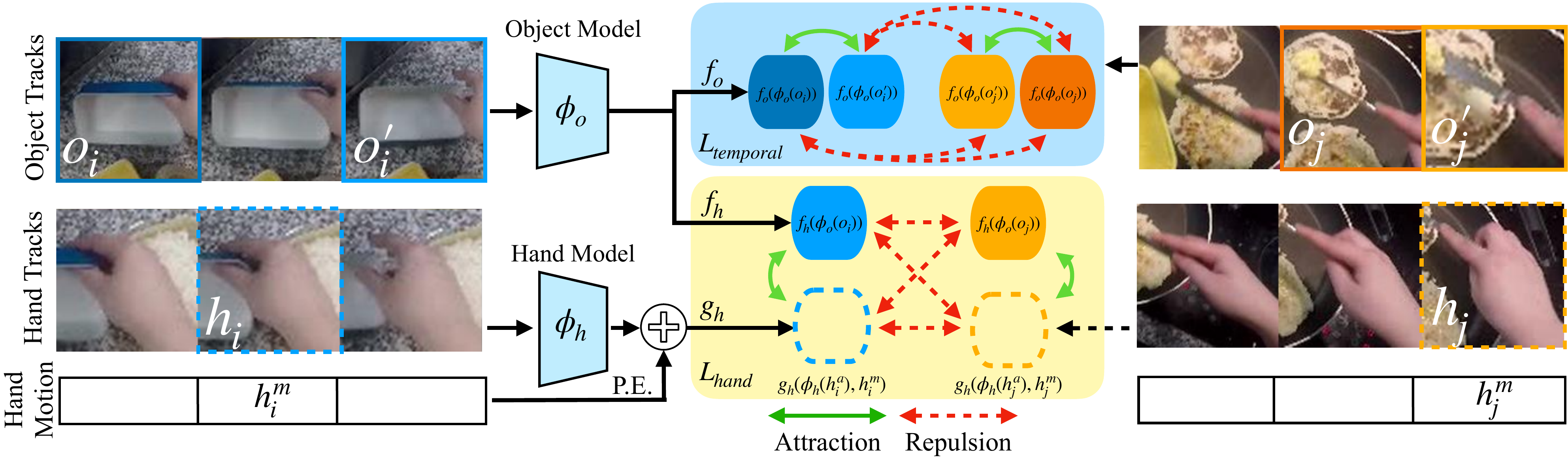}
\caption{\textbf{Temporal SimCLR with Object-Hand Consistency (TSC+OHC).}
Given batches of quadruples containing object crops pairs $o_i, o_i'$,
alongside the corresponding hand crop $h^a_i$ and hand motion $h^m_i$, TSC+OHC
employs two losses $\Lt$ and $\Lh$. $\Lt$ encourages object crops close in time
to be close to one another, while being far away from other object crops. $\Lh$
encourages corresponding object and hands to be close to one another, while
being far away from other objects and hands. Different encoders are used for
objects and hands ($\phi_o$ and $\phi_h$), and different heads ($f_o$ and $f_h$) 
are used for objects for $\Lt$ and $\Lh$. Best viewed in color.}
\figlabel{states-method}
\end{figure*}
 
We work with the challenging \epic dataset from Damen
\etal~\cite{damen2020collection}, and use the hand and object-of-interaction
detector from Shan \etal~\cite{shan2020understanding} 
This detector provides
per-frame detection boxes for both hands and the
objects undergoing interaction, along with the hand contact state (whether
the hands are touching something or not). 
We further obtain predictions for hand
grasp-types for the detected hands, using a model trained on the 71-way
grasp-type classification dataset from Rogez~\etal\cite{rogez2015understanding}. 
We string together detected hands and objects-of-interaction in consecutive
frames to form object-of-interaction and hand tracks as shown in \figref{data-gen}.
We use these tracks for learning state-sensitive features (\secref{state-features}).
Affordances (where and how hands interact with objects)
are learned using per-frame predictions (\secref{hotspots}).

\subsection{State Sensitive Features via Temporal and Hand Consistency}
\seclabel{state-features}

Our formulation builds upon two key ideas: consistency of object
states {\it in time} and {\it with hand pose}. 
Our training objective encourages object crops, that are close in time or are
associated with similar hand appearance and motion, to be similar to one
another; while being far from random other object crops in the dataset. We
realize this intuition through contrastive learning and propose a joint loss: $\Lt + \lambda
\Lh$.
$\Lt$ encourages {\it temporal} 
consistency by sampling naturally occurring temporal augmentations as 
additional transforms. $\Lh$ uses hands as contrasting examples; positives
being the hands that temporally correspond to the object crop, and negatives
being other randomly sampled hands. $\Lh$ indirectly encourages similarity
between {\it different} objects that are similarly interacted by hands, and so
are likely to be in similar states.

\begin{figure*}
\centering
\insertWL{1.0}{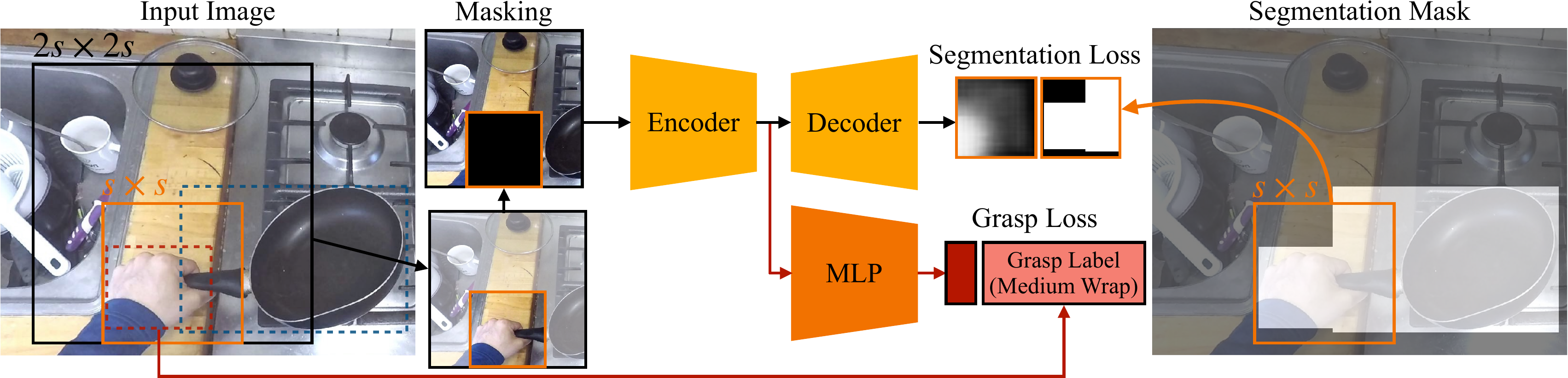}
\caption{\textbf{Affordances via Context Prediction (\acp)}. We sample a patch
(orange) from the input image around the detected hand (shown on the left). We
then consider a context region (black) of twice the size around the sampled
patch containing parts of the object being interacted with. We mask out the
sampled patch (Masking) to hide the hand. 
Our model uses the surrounding context to make predictions for probability of
interaction and grasps afforded in the masked region.
We paste the hand and object boxes to generate supervision for interaction
regions. Supervision for grasp prediction branch comes by running a network
trained on GUN71 dataset~\cite{rogez2015understanding} on the hand crop.}
\figlabel{hotspot-data}
\end{figure*}

We construct batches for contrastive learning by sampling an object crop $o_i$
and a temporally close hand crop $h^a_i$ from tracks shown in
\figref{data-gen}.
We also encode the hand motion $h^m_i$, by concatenating the location and scale
of the hand box relative to the object box over three neighboring frames.
$h^a_i$ and $h^m_i$ jointly represent the hand: $h^a_i$ describes the
appearance and $h^m_i$ describes the motion.  We sample another frame $o'_i$
from the same object track, as a temporal augmentation of $o_i$.

Given $N$ such quadruples $(o_i, o_i', h^a_i, h^m_i)$, we construct positive and contrasting
negative pairs as shown in \figref{states-method}. In $\Lt$, for each $o_i$, $o'_i$ is positive and all
other objects $o_j$s and $o'_j$s are negatives. In $\Lh$, for each $o_i$, $[h^a_i, h^m_i]$ (hand appearance and motion)
serves as the positive while all other objects $o'_j$s and hands $[h^a_j, h^m_j]$s
are negatives; and for each $[h^a_i, h^m_i]$, $o_i$ is positive and all other
objects $o_j$s and hands $[h^a_j, h^m_j]$s are negatives. 
All crops $o_i, o'_i, h^a_i$ are transformed using the 
standard SimCLR augmentations.

We setup contrastive losses by passing object and
hand crops through convolutional trunks $\phi_o$ and $\phi_h$,
respectively. We use a projection head $f_o$ for $\Lt$, and 2 projection heads
$f_h, g_h$ (for object and hand crops, respectively) for $\Lh$. $h^m_i$ is encoded
via positional encoding and appended to $\phi_h(h^a_i)$ before being fed into
projection head $g_h$. We use cosine similarity, and the normalized temperature scaled cross-entropy
loss ({\it NT-Xent}) following SimCLR~\cite{chen2020simple}.

We call our full formulation with both these loss terms as Temporal SimCLR with
Object-Hand Consistency or TSC+OHC. We also experiment with Temporal SimCLR
or TSC that only uses the temporal term (\ie setting $\lambda$ to $0$).
The output of these formulations is $\phi_o$,
which is our state-sensitive feature representation. 
In \secref{expts-features}, we evaluate the quality of $\phi_o$ on an object state classification task.

\subsection{Object Affordances via Context Prediction}
\seclabel{hotspots}
The next aspect of interactive object understanding that we tackle is to infer
what interactions do objects placed in scenes afford and where, which we refer
to jointly as object affordances. Specifically, we want to infer a) the Regions
of Interaction (ROI) in the scene (\ie pixels that are likely to be interacted with
when undertaking some common actions), and b) the hand-grasp type that is
applicable at that region. 

Information about both these aspects is directly available in egocentric
videos. As hands interact with objects, we observe where they touch
and via what grasp. However, learning models from such data as-is is hard; wherever 
we have the hand for supervision, we also have the same hand that trivially reveals 
the information that we want to predict. As a result, a naively trained model won't
learn anything about the underlying object.  To circumvent this issue, we
propose a context prediction task: prediction of the hand locations and grasp
type from image patches around the hand, but with hands \textit{masked out}.
Our context prediction task encourages the model to use the context around an
object to predict regions of interaction. For instance in \figref{hotspot-data},
the model can predict the region-of-interaction (location of the
handle) from part of the pan visible in the context region.
We call our model {\it Affordances via Context Prediction (ACP)}.

\noindent \textbf{Data Generation.}
Our data generation process, shown in \figref{hotspot-data}, assumes detections
for hands, object-of-interaction, contact state, and grasp-type (see \figref{data-gen}). 
Starting with the hand that are in
contact state, we sample a $s \times s$ patch around the
detected hand. We crop out a $2s \times 2s$ {\it asymmetric} context region around
this patch, with the $s \times s$ hand patch being at the bottom center of this
context region. We mask out the $s \times s$ hand patch
to obtain a masked context region that serves as the input to our model. The
goal for the model is to predict a) the segmentation mask for the hand (and
optionally also the object-of-interaction) inside the masked region, and b) the
grasp-type exhibited by the hand. Supervision for these comes from the
detections and the grasp predictions as described above. As the detector
from~\cite{shan2020understanding} only outputs boxes, we derive an
approximate segmentation mask by pasting the detection boxes.  We also
sample additional positives from around the object-of-interaction detections
and negatives from the remaining image.  
We sample patches at varying scale and reshape them to $128 \times 128$ before
feeding them into our network. 

\noindent \textbf{Model Architecture and Training.} The masked context region
is processed through a \RS{50} encoder, followed by two separate heads to
predict the segmentation mask and the grasp type. The segmentation head uses a
deconvolutional decoder to produce $64\times 64$ segmentation masks, and is
trained using binary cross-entropy loss with the positive class weighed by a factor of 4. The grasp-type prediction uses 2
fully-connected layers to predict the applicable grasp types. As more than one
grasp is applicable, we model it as a multi-label problem and train using
independent binary cross-entropy losses for each grasp-type. For each example,
the highest scoring class from GUN71 model is treated as positive, lowest 15
are treated as negatives, and the remaining are not used for computing loss.

\noindent \textbf{Inference.}
For inference, we sample patches densely at 3 different scales.  We reshape
them to $128\times 128$ and mask out the $64\times 64$ bottom center region,
before feeding them into our model. Predictions from the patches are pasted
back onto the original image to generate per-pixel probability for a)
interaction, and b) afforded hand grasps.

Though we only considered predicting coarse segmentation and grasp-types our
contextual prediction framework is more general. Given appropriate pre-trained
models, ACP can be trained for richer hand representations such as
fine-grained segmentation, 2D or 3D hand pose.

  \section{Experiments}
\seclabel{experiments} We train our models on in-the-wild videos from
\epic~\cite{damen2020collection}. Our experiments test the
different aspects of interactive object understanding that we pursue:
state-sensitive features (\secref{expts-features}), and object affordance
prediction (\ie identifying regions-of-interaction (\secref{expts-hotspots})
and predicting hand grasps afforded by objects (\secref{expts-grasps})). 
We focus on comparing different sources of 
supervision, and on evaluating our design choices. As we pursue
relatively new tasks, we collect two labeled datasets on top of \epic to
support the evaluation: \SD for state-sensitive feature learning and \RoID for
regions-of-interaction. We adapt the \ycb benchmark~\cite{corona2020ganhand}
for afforded hand-grasp prediction.

All our experiments are conducted in the challenging setting where there is
{\it no overlap between training and testing participants} for \epic
experiments,\footnote{Note that the detector from~\cite{shan2020understanding} was trained
on 18K labeled frames from the \epic dataset. To ensure that our trainings only
see realistic predictions, we use {\it leave one out} predictions
from~\cite{shan2020understanding}: we split the train set into 5 parts by
participants, retrain~\cite{shan2020understanding} on 4, use
predictions on the 5\textsuperscript{th} (\ie unseen participants); and
repeat this 5 times over.} 
and {\it no overlap in objects} for experiments on \ycb.

\subsection{State Sensitive Features for Objects}
\seclabel{expts-features}
We measure the state sensitivity of our learned feature space $\phi_o$, by
testing its performance for fine-grained object state classification. We design
experiments to measure the effectiveness of focusing on the hands to derive a)
data and b) supervision for learning; and our choice of learning
method. We also compare the quality of our self-supervised features to existing
methods for learning such features via: action classification on \epic and 
state classification on Internet data~\cite{isola2015discovering}.

\noindent \textbf{Object State Classification Task and Dataset.}
For evaluation, we design and collect \SD, a labeled object state
classification dataset. \SD builds upon the raw data in the \epic dataset and
consists of 10 state categories: \ttt{open, close, inhand, outofhand, whole,
cut, raw, cooked, peeled, unpeeled}. We selected these state categories as they
are defined somewhat unambiguously and had enough examples in the \epic
dataset. \SD consists of 14,346 object bounding boxes from the \epic dataset
(2018 version), each labeled with 10 binary labels corresponding to the 10 state
classes. We split the dataset into training, validation, and testing sets based
on the participants, \ie boxes from same participant are in the same split.

To maximally isolate impact of pre-training, we only train a linear 
classifier on representations learned by the different methods. 
We report the mean average precision across these 10 
binary state classification tasks. We also consider two
settings to further test generalization: a) low training data (only using
12.5\% of the \SD train set), and b) testing on novel object categories (by
holding out objects from \SD train set).

\noindent \textbf{Implementation Details.}
{\it Object-of-Interaction Tracks.} We construct tracks by linking together
hand-associated object detections with IoU $\geq$ 0.4 in temporally adjacent frames.
We median filter the
object box sizes to minimize jumps due to inaccurate detections.
This resulted in 61K object tracks
(on average 2.2s long) for training. We extract patches at 10 fps from these
tracks.
\\ {\it Model Architecture.} All models use the \rs~\cite{he2016deep} backbone
initialized with ImageNet pre-training. We average pooled the
$4\times 4$ output from the \rs backbone and introduced 2 fully connected
layers to arrive at a 512 dimensional embedding for all models. \\
{\it Self-supervision Hyper-parameters.} Our proposed models (TSC, TSC+OHC) use standard data augmentations: color jitter,
grayscale, resized crop, horizontal flip, and Gaussian blur. Temporal
augmentation frames $o'_i$ were within one fourth of the track length. For the
TSC+OHC model: hand boxes within 0.3s from the object boxes were considered as
corresponding and $h^m_i$ was computed using 3 consecutive frames. See other
details in Supplementary.

\renewcommand{\arraystretch}{1.1}
\begin{table}
\setlength{\tabcolsep}{5pt}
\centering
\vspace{10pt}
\caption{Mean average precision for object state classification on the \SD test set ($\mu \pm \sigma$ over 3 pre-training seeds). 
Our self-supervised features outperform features from ImageNet-pretraining, other self-supervision (TCN, SimCLR), and even 
semantic supervision across all settings. Performance boost is larger in harder settings: low-data and generalization to novel objects.}
\tablelabel{state}
\resizebox{\linewidth}{!}
{
\begin{tabular}{lcccc}
\toprule
& \multicolumn{2}{c}{\textbf{Novel Objects}} & \multicolumn{2}{c}{\textbf{All Objects}}\\
\cmidrule(lr){2-3} \cmidrule(lr){4-5}
Linear classifier training data & 12.5\% & 100\% & 12.5\% & 100\% \\ 
\midrule
ImageNet Pre-trained                            & 70.2 \std 0.0 & 74.5 \std 0.0 & 78.2 \std 0.0 & 83.1 \std 0.0 \\
TCN~\cite{sermanet2018time}                     & 56.1 \std 1.9 & 63.9 \std 1.1 & 62.5 \std 0.8 & 73.4 \std 1.4 \\
SimCLR~\cite{chen2020simple}                    & 71.9 \std 0.2 & 77.1 \std 1.0 & 77.4 \std 1.0 & 81.0 \std 0.9 \\
SimCLR + TCN                                    & 63.7 \std 0.3 & 68.4 \std 1.6 & 72.9 \std 1.3 & 77.4 \std 1.2 \\
Semantic supervision\\
$\;$ via EPIC action classification          & 70.9 \std 1.9 & 77.0 \std 0.9 & 72.1 \std 0.8 & 77.9 \std 1.3 \\
$\;$ via MIT States dataset~\cite{isola2015discovering} & 70.1 \std 1.4 & 73.9 \std 0.8 & 76.4 \std 0.6 & 81.5 \std 1.3 \\ 
Ours [TSC]                         & 74.5 \std 0.9 & 80.2 \std 0.4 & 81.4 \std 1.0 & 84.2 \std 1.0 \\
Ours [TSC+OHC]                     & \textbf{79.7 \std 0.6} & \textbf{81.8 \std 0.4} & \textbf{82.6 \std 0.2} & \textbf{84.8 \std 0.4} \\
\bottomrule
\end{tabular}}
\end{table}
 
\begin{figure}
\centering
\insertWL{1.0}{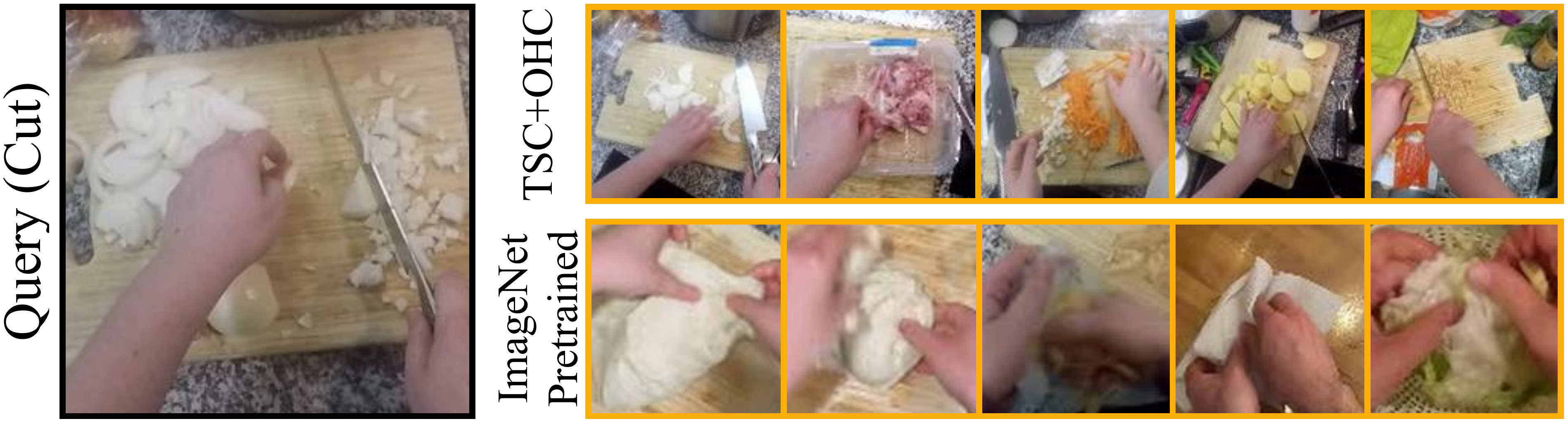}
\caption{\textbf{Object in similar states}.
Nearest neighbors in our learned feature space exhibit similar state.
} 
\figlabel{object-object-nn}
\end{figure}
 
\noindent \textbf{Results.}
\tableref{state} reports the mean average precision (higher is better) for
object state classification on the \SD test set. We also report the standard
deviation across 3 pre-training runs. We compare among our models
and against a) ImageNet pre-training (\ie no further self-supervised
pre-training), b) non-temporal self-supervision via
SimCLR~\cite{chen2020simple}, c) an alternate temporal self-supervision method
(Time Contrastive Networks, TCN~\cite{sermanet2018time}), and d) semantic
supervision from action classification on \epic and state classification on
MIT States dataset~\cite{isola2015discovering}. We describe these comparison
points as we discuss our key takeaways. 

\noindent {\bf Features from TSC and TSC+OHC are more state-sensitive than
ImageNet features.} ImageNet pre-trained features provide a strong baseline
with an mAP of 83.1\%. TSC and TSC+OHC boost performance
to 84.2\% and 84.8\%, respectively. Improvements get amplified in
the challenging low-data and novel category settings for all models, with our
full model TSC+OHC improving upon ImageNet features by 4.4\% and 9.5\%,
respectively. These trends are also borne out when we visualize nearest
neighbors in the learned feature spaces in \figref{object-object-nn}.

\noindent {\bf TSC and TSC+OHC outperform other competing self-supervision
schemes.} Temporal SimCLR, even by itself, is more effective than vanilla
SimCLR that has access to the same crops but ignores the temporal information.
We also outperform TCN~\cite{sermanet2018time}, a leading method for temporal
self-supervision, and TCN combined with SimCLR. TCN uses negatives from
the same track. These are harder to identify in \epic because of the large
variability in time-scales at which changes occur (\eg \textsc{open} \vs
\textsc{chop} action). 

\noindent {\bf Supervision from object-hand consistency improves performance.}
TSC+OHC improves over just TSC by 0.6\%
with larger gains (of up to 5.2\%) in the more challenging novel category and
limited data settings. This confirms our hypothesis that
observation of what hands are doing, aids the understanding of object states.
\figref{hand-object-nn} shows some nearest neighbors retrievals that further
support this.

\noindent {\bf TSC and TSC+OHC models outperform semantically supervised models.}
Conventional wisdom would have suggested pre-training a model on images
gathered from the Internet for this or related tasks. MIT States dataset from
Isola \etal~\cite{isola2015discovering} is the largest such dataset with 32,915
training images labeled with applicable adjectives. Surprisingly, our
self-supervised models outperform features learned through supervised
training on this dataset by 3.3 to 9.6\%, perhaps due to the domain gap
between Internet and egocentric data. 

Another common belief is to equate action classification to video understanding. We assess this by comparing against features from the action
classification task on \epic. This model was trained on our tracks using the
most common 32 action labels along with their temporal extent, available as
part of the \epic dataset. Both TSC and TSC+OHC features outperform action classification
features by 3 to 10\%. 
Thus, while the action classification task is useful for
many applications, it fails to learn good state-sensitive features.

\begin{figure}
\centering
\insertWL{1.0}{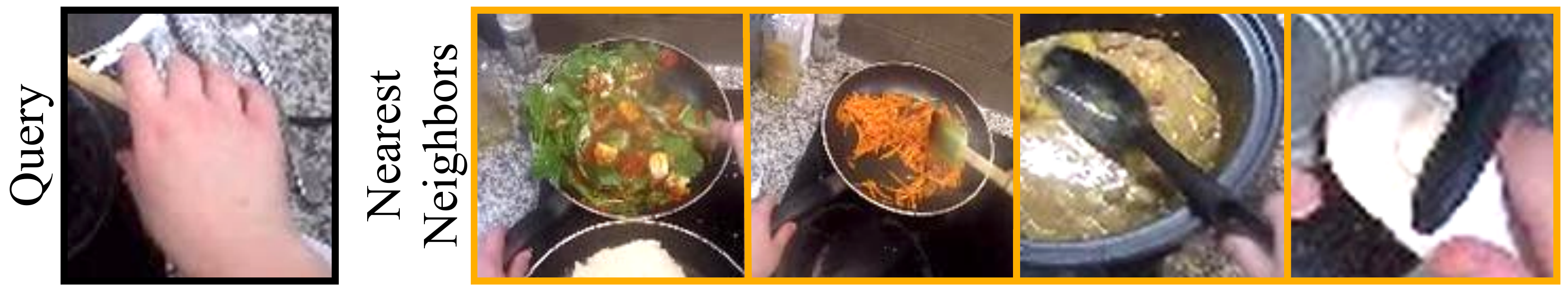}
\caption{\textbf{Objects affording similar hands.} We retrieve objects that are associated with hands having features similar to the query hand. 
Objects that are being interacted with similarly get retrieved.}
\figlabel{hand-object-nn}
\end{figure}

\noindent \textbf{Ablations.} In Supplementary, we compare alternate ways of
obtaining tracks when learning with TSC. We ablate two aspects: what we track
(background crop, background object, object-of-interaction) and how we track it
(no tracking, off-the-shelf tracker~\cite{li2019siamrpn++}, hand-context).
Ablations reveal the utility of object-of-interaction tracks particularly as they
enable use of hand consistency. We also study the role of appearance and motion
individually for representing the hand. We found both to be useful over TSC
with motion being more important than appearance.
 
\subsection{Regions of Interaction}
\seclabel{expts-hotspots}
\begin{figure}
\centering
\insertWL{1.0}{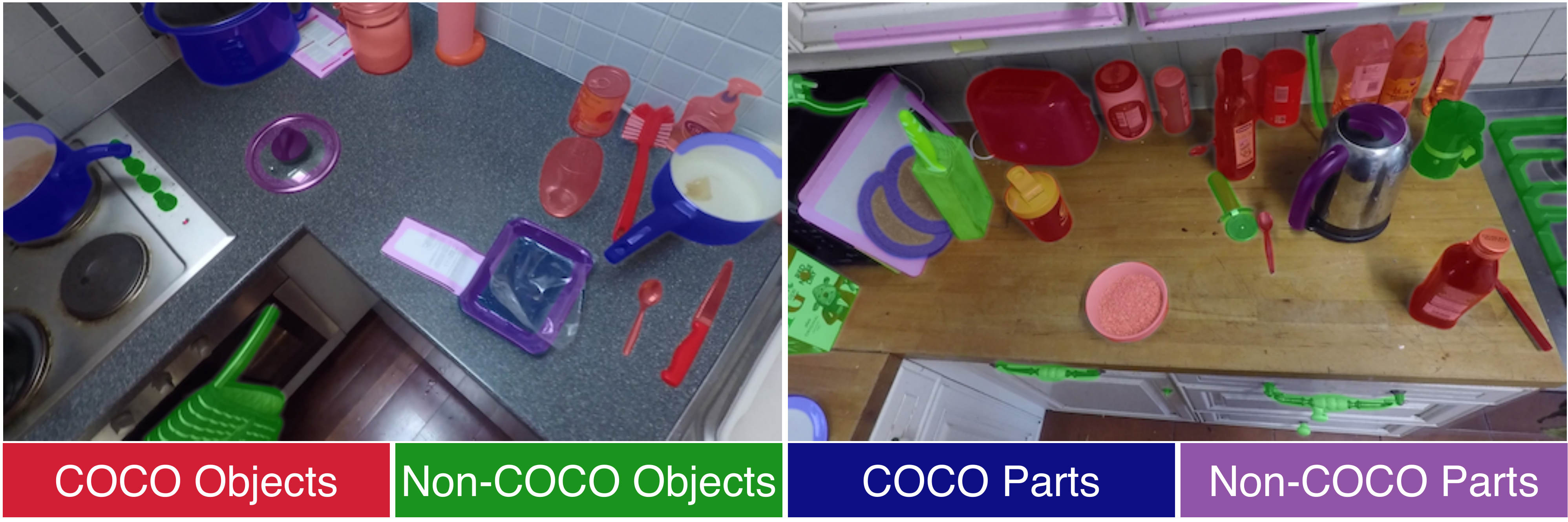}
\caption{\textbf{Images from the proposed \RoID dataset}. Each image is annotated for regions of interaction \ie where the human participants frequently interact with. Every annotation is also labeled with one of four attributes: COCO objects, Non-COCO objects, COCO parts, or Non-COCO parts.}
\figlabel{roi-dataset}
\end{figure}
 
\noindent \textbf{Regions-of-Interaction Task and Dataset.}
We design and collect \RoID, a labeled region-of-interaction dataset.  \RoID
builds on top of the \epic dataset, and consists of 103 diverse images with
pixel-level annotations for regions where human hands {\it frequently} touch in
everyday interaction. Specifically, image regions that afford any of the most
frequent actions: \ttt{take, open, close, press, dry, turn, peel} are
considered as positives. We manually watched video for multiple participants to
define a) object categories, and b) specific regions within each category where
participants interacted while conducting any of the 7 selected actions.
These 103 images were sampled from across 9 different kitchens (7 to 15 images
with minimal overlap, from each kitchen). \RoID is only used for evaluation, and
contains 32 val images and 71 test images. Images from the same
kitchen are in the same split. 
The Regions-of-Interaction task is to score each pixel in the image 
with the probability of a hand interacting with it. Performance is measured 
using average precision. 

To enable detailed analysis, each annotated region is assigned two binary
attributes: a) Is-COCO-object (if region is on an object that is included in the
COCO dataset), b) Is-whole-object (if region covers the whole object). 
This results in 4 sub-classes (see \figref{roi-dataset}), allowing evaluation on more
challenging aspects: \eg small objects that are not
typically represented in object detection datasets such as knobs (Non-COCO Object), 
or when interaction is localized to a specific object part such as the pan-handle 
(COCO Part) or the cutting-board-edges (Non-COCO Part).
We also evaluate in the \ttt{1\% slack} setting where regions within 
20 pixels (1\% of image width) of the segmentation boundaries is ignored 
to discount small leakage in predictions.

\renewcommand{\arraystretch}{1.0}
\begin{table*}
\setlength{\tabcolsep}{14pt}
\centering
\vspace{10pt}
\caption{\textbf{Average Precision for Region-of-Interaction Prediction.} We
report the overall AP and AP across the different types of interaction regions. 
We also report AP with 1\% \ttt{slack} at the boundaries where we don't penalize any 
leakage at regions within 20 pixels (1\% of image width) of the mask boundaries.
Without training on segmentation masks,
our method outperforms methods based on objectness (SalGAN and DeepGaze2),
action classification (Interaction Hotspots), and are able to come close to
Mask~RCNN that is trained with supervised segmentation masks.
We achieve the strongest performance across all categories when combined with Mask~RCNN. 
Highest numbers are boldfaced and the second highest are italicized.}
\tablelabel{hotspots}
\resizebox{\linewidth}{!}{
\begin{tabular}{lggccccccccc}
\toprule
& \multicolumn{2}{g}{\small{\bf Overall}}
& \multicolumn{2}{c}{\small{\bf COCO Objects}}
& \multicolumn{2}{c}{\small{\bf Non-COCO Objects}}
& \multicolumn{2}{c}{\small{\bf COCO Parts}}
& \multicolumn{2}{c}{\small{\bf Non-COCO Parts}}
\\
\cmidrule(lr){2-3} \cmidrule(lr){4-5} \cmidrule(lr){6-7} \cmidrule(lr){8-9} \cmidrule(lr){10-11}
Slack at segment boundaries & \ttt{0\%} & \ttt{1\%} & \ttt{0\%} & \ttt{1\%} & \ttt{0\%} & \ttt{1\%} & \ttt{0\%} & \ttt{1\%} & \ttt{0\%} & \ttt{1\%} \\
\midrule
Mask~RCNN [all]      & 41.9      & 46.7      & 40.6      & 45.0      & 13.3      & 16.0      & 11.5      & 14.4      & 3.6      & 4.3\\
Mask~RCNN [relevant] & \bf 64.0  & \bf 70.0  & \bf 72.2  & \bf 78.1  & 22.8      & 28.7      & \bf 31.0  & \bf 39.6  & 6.8      & 9.1 \\
Interaction Hotspots~\cite{nagarajan2019grounded} & 43.8      & 52.0      & 26.5      & 33.9      & 23.0      & 29.5      & 12.2      & 16.7      & 6.9      & 9.6 \\
SalGAN~\cite{salgan}                              & 48.7      & 56.4      & 40.8      & 49.1      & 24.5      & 31.0      & 11.4      & 15.7      & 4.7      & 6.4 \\
DeepGaze2~\cite{deepgazeii}                       & 55.7      & 64.6      & 44.8      & 55.1      & \bf 35.8  & \bf 45.4  & 11.4      & 16.5      & \bf {7.4}  & \it{10.8} \\
ACP ({\it{Ours}}) & \it{57.4} & \it{67.3} & \it{49.6} & \it{60.8} & \it{33.7} & \it{44.7} & \it{14.7} & \it{22.5} & \it{7.2} & \bf{11.3} \\
\midrule
Mask~RCNN+DeepGaze2                                & \it{66.6} & \it{72.9} & \it{74.4} & \it{80.1} & \it{26.2} & \it{33.5} & \it{31.7} & \it{40.6} & \it{7.1} & \it{9.8}\\
Mask~RCNN+ACP ({\it{Ours}}) & \bf 68.7 & \bf 76.4 & \bf 76.2 & \bf 83.0 & \bf 31.1 & \bf 41.9 & \bf 32.5 &\bf 43.7 & \bf 7.4 & \bf 11.4 \\
\bottomrule 
\end{tabular}}

\end{table*}

\noindent \textbf{Implementation Details.}
We train our model on 250 videos from the 2018 \epic dataset. We exclude 
videos from the 9 kitchens used for evaluation in \RoID. 
Details of the grasp classification branch are in \secref{expts-grasps}.

\noindent \textbf{Results.} \tableref{hotspots} reports the average precision. 
We compare to three classes of
methods: a) objectness based approaches SalGAN~\cite{salgan} and
DeepGaze2~\cite{deepgazeii}, trained using human gaze data / manual
labels; b) instance segmentation based approaches that use
Mask~RCNN~\cite{he2017mask} predicted masks for all / relevant classes; and c)
interaction hotspots from Nagarajan \etal~\cite{nagarajan2019grounded} that
derives supervision from manually annotated object bounding boxes and action labels in the \epic dataset. Given the strong performance of Mask~RCNN-based methods, we also
report the performance by aggregating predictions from Mask~RCNN with \acp 
and next most competitive baseline, DeepGaze2. 
Aggregation is done using a weighted summation of predictions (weight selected 
using validation performance).

Overall, Mask-RCNN when restricted to relevant categories, performs
the best. This is not surprising as it is supervised using over 1 million object 
segmentation masks. However, its performance suffers on non-COCO objects or their parts.
Methods that utilize more general supervision start to do better. And in spite
of not being trained on any segmentation masks at all, \acp ({\it Ours}) is able to
outperform past methods. It starts to approach the performance of Mask~RCNN,
particularly in the 1\% \ttt{slack} setting.

When combined with Mask~RCNN, \acp achieves the strongest performance
across all categories. It improves upon the Mask~RCNN based method by 4.7\%,
indicating that our method is able to effectively learn about objects not
typically included in detection datasets (\eg stove knobs), and object
parts (\eg handle for fridges and drawers). Furthermore, our method provides a
more complete interactive understanding by also predicting afforded grasps as
we discuss in \secref{expts-grasps} and show in \figref{vis-hotspots}.

\noindent \textbf{Ablations.} Experiments in Supplementary study the effects of
variations in network input (not hiding hands, symmetric context, not filtering
based on contact), model architecture, and data sampling and supervision (using
just objects, or using just hands, or using hand masks rather than boxes). We
find that all design choices contribute to ACP's performance. Further
improvements can be had from richer hand understanding (segmentation
masks \vs box masks).

\begin{figure*}[t]
\centering
  \insertWL{1.00}{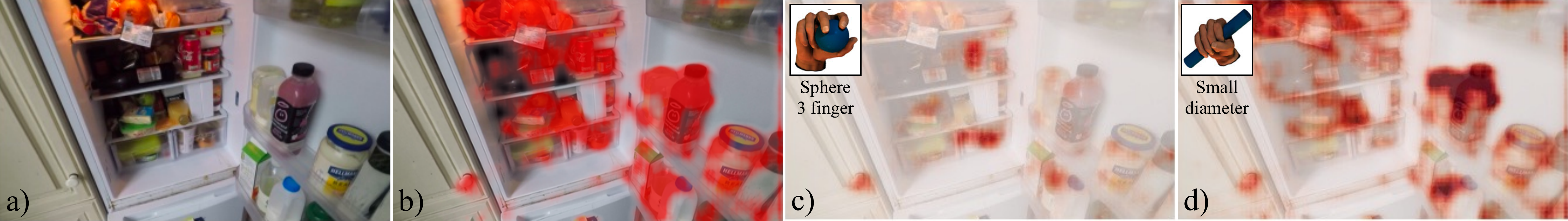}
\caption{{\bf Object Affordance Predictions.} For the input image (shown in
(a)), we show the predicted regions of interaction in (b). 
Our method successfully detects multiple possible regions of interaction:
bottles, jars, general objects in the fridge, and door knobs. We also 
visualize the per-pixel probability of affording the sphere 3 finger grasp in
(c), and small diameter grasp in (d). The sphere 3 finger grasp is predicted for the
door knob, bottle caps and cans; while the small diameter grasp is predicted for
bottles, jars and cans. Thumbnails visualizing hand grasps reproduced
from~\cite{feix2015grasp}.}
\figlabel{vis-hotspots}
\end{figure*}
  
\subsection{Hand Grasps Afforded by Objects}
\seclabel{expts-grasps}

\noindent \textbf{Grasps Afforded by Objects (\gao) Task and Dataset.} We use
the \ycb dataset~\cite{corona2020ganhand} to evaluate performance at the Grasps
Afforded by Objects (\gao) task.  The dataset annotates objects in the scenes
from~\cite{xiang2018posecnn} with all applicable grasps from a 33-class
taxonomy~\cite{feix2015grasp}.  We split the dataset into training (110K
images, 776K grasps, used only to obtain a supervised ceiling), validation (60
images, 230 grasps) and testing (180 images, 760 grasps). Val and test sets
contain {\it novel} objects not present in the training set. Given an image
with a segmentation mask for the object under consideration, the \gao task is
to predict the grasps afforded by the object. As multiple grasps are applicable
to each object, we measure AP for each grasp independently and report mAP
across the 7 (of 33) grasps present in the val and test sets.

\noindent \textbf{Implementation Details.} 
The grasp prediction branch in \acp is trained on predictions from a grasp
classification model trained on the \gun dataset. We only use the 33 classes
relevant to the task on the \ycb dataset from the 71-way output. To test our
grasp affordance prediction on \ycb objects, we average the spatial grasp
scores over the pixels belonging to the object mask.

We found it useful to adapt the \gun classifier to \epic to
generate good supervision. This was done via self-supervision
by using an additional $\Lt$ loss on the \epic~{\it hand} tracks 
(analogous to one used for objects in \secref{state-features}) while
training on \gun.

\noindent \textbf{Results.} 
For reference, chance performance is 30.2\%, and supervised performance is
56.8\%. The supervised method is trained on \ycb using ground truth annotations
for afforded hand grasps on the 110K training images for the 15 training
objects.  Our method achieves an mAP of 38.1\%. It reduces the gap between
chance performance and the supervised method by 30\%. 
Adaptation using $\Lt$ on hands helped (34.3\% \vs 38.1\%).

  \balance
\section{Discussion and Limitations}
We have shown that observation and analysis of human hands interacting 
with the environment is a rich source of information 
for learning about objects and how to interact with them; even when
using a relatively crude understanding of the hand via 2D boxes.
A richer understanding of hands (through segmentation, fine-grained 2D / 3D pose, 
and even 3D reconstruction) would enable a richer understanding of objects in the future.
Our work relies on off-the-shelf models for generating data and supervision,
and is limited by the quality of their output. 

The \acp model in \secref{hotspots}, doesn't look at the pixels it is making
predictions on. This causes our predictions to not be as well-localized.
Our \RoID task requires fine-grained reasoning for large objects (\eg
microwaves), but not as much for small objects because of subjectivity in
annotation.  Collecting fine-grained datasets for where we can interact with
small objects in scenes will enable better evaluation. Similarly, large-scale
in-the-wild datasets for evaluation of grasps afforded by objects can help.
Finally, we tackled different aspects of interactive object understanding in
isolation, a joint formulation could do better.

\noindent \textbf{Ethical considerations, bias, and potential 
negative societal impact}: Egocentric data is of sensitive nature. We relied on 
existing public data from \epic dataset (which obtained necessary consent 
and adopted best practices). Though, our self-supervised techniques mitigate bias
introduced during annotation, we acknowledge that our models inherit and suffer
from bias (\eg what objects are present, their appearance and usage) present in the 
raw videos in \epic dataset. 
As with all AI research, we acknowledge potential for negative 
societal impact. Interactive object understanding can enable many  
useful applications (\eg building assistive systems), but could also be used for
large-scale automation which, if not thought carefully about, 
could have negative implications.
 
\noindent \textbf{Acknowledgements:} We thank David Forsyth, Anand Bhattad, and
Shaowei Liu for useful discussion. We also thank Ashish Kumar and Aditya
Prakash for feedback on the paper draft.  This material is based upon work
supported by NSF (IIS-2007035), DARPA (Machine Common Sense program), and an
Amazon Research Award.

{\small
\bibliographystyle{ieee_fullname}
\bibliography{biblioLong, refs}

\begin{thebibliography}{10}\itemsep=-1pt

\bibitem{bandini2020analysis}
Andrea Bandini and Jos{\'e} Zariffa.
\newblock Analysis of the hands in egocentric vision: A survey.
\newblock {\em IEEE Transactions on Pattern Analysis and Machine Intelligence
  (TPAMI)}, 2020.

\bibitem{brahmbhatt2019contactdb}
Samarth Brahmbhatt, Cusuh Ham, Charles~C Kemp, and James Hays.
\newblock Contactdb: Analyzing and predicting grasp contact via thermal
  imaging.
\newblock In {\em Proceedings of the IEEE Conference on Computer Vision and
  Pattern Recognition (CVPR)}, pages 8709--8719, 2019.

\bibitem{cai2016understanding}
Minjie Cai, Kris~M Kitani, and Yoichi Sato.
\newblock Understanding hand-object manipulation with grasp types and object
  attributes.
\newblock In {\em Robotics: Science and Systems (RSS)}, volume~3. Ann Arbor,
  Michigan;, 2016.

\bibitem{cao2020reconstructing}
Zhe Cao, Ilija Radosavovic, Angjoo Kanazawa, and Jitendra Malik.
\newblock Reconstructing hand-object interactions in the wild.
\newblock In {\em Proceedings of the IEEE International Conference on Computer
  Vision (ICCV)}, 2021.

\bibitem{caron2020unsupervised}
Mathilde Caron, Ishan Misra, Julien Mairal, Priya Goyal, Piotr Bojanowski, and
  Armand Joulin.
\newblock Unsupervised learning of visual features by contrasting cluster
  assignments.
\newblock In {\em Advances in Neural Information Processing Systems (NeurIPS)},
  2020.

\bibitem{chen2020simple}
Ting Chen, Simon Kornblith, Mohammad Norouzi, and Geoffrey Hinton.
\newblock A simple framework for contrastive learning of visual
  representations.
\newblock In {\em Proceedings of the International Conference on Machine
  Learning (ICML)}, pages 1597--1607. PMLR, 2020.

\bibitem{chen2020improved}
Xinlei Chen, Haoqi Fan, Ross Girshick, and Kaiming He.
\newblock Improved baselines with momentum contrastive learning.
\newblock {\em arXiv preprint arXiv:2003.04297}, 2020.

\bibitem{Chen2021ExploringSS}
Xinlei Chen and Kaiming He.
\newblock Exploring simple siamese representation learning.
\newblock In {\em Proceedings of the IEEE Conference on Computer Vision and
  Pattern Recognition (CVPR)}, 2021.

\bibitem{corona2020ganhand}
Enric Corona, Albert Pumarola, Guillem Alenya, Francesc Moreno-Noguer, and
  Gregory Rogez.
\newblock Ganhand: Predicting human grasp affordances in multi-object scenes.
\newblock In {\em Proceedings of the IEEE Conference on Computer Vision and
  Pattern Recognition (CVPR)}, June 2020.

\bibitem{damen2020collection}
Dima Damen, Hazel Doughty, Giovanni~Maria Farinella, Sanja Fidler, Antonino
  Furnari, Evangelos Kazakos, Davide Moltisanti, Jonathan Munro, Toby Perrett,
  Will Price, and Michael Wray.
\newblock The epic-kitchens dataset: Collection, challenges and baselines.
\newblock {\em IEEE Transactions on Pattern Analysis and Machine Intelligence
  (TPAMI)}, 2020.

\bibitem{damen2016you}
Dima Damen, Teesid Leelasawassuk, and Walterio Mayol-Cuevas.
\newblock {You-Do, I-Learn}: Egocentric unsupervised discovery of objects and
  their modes of interaction towards video-based guidance.
\newblock {\em Computer Vision and Image Understanding}, 149:98--112, 2016.

\bibitem{doersch2015unsupervised}
Carl Doersch, Abhinav Gupta, and Alexei~A Efros.
\newblock Unsupervised visual representation learning by context prediction.
\newblock In {\em Proceedings of the IEEE International Conference on Computer
  Vision (ICCV)}, pages 1422--1430, 2015.

\bibitem{ehsani2020use}
Kiana Ehsani, Shubham Tulsiani, Saurabh Gupta, Ali Farhadi, and Abhinav Gupta.
\newblock Use the force, luke! learning to predict physical forces by
  simulating effects.
\newblock In {\em Proceedings of the IEEE Conference on Computer Vision and
  Pattern Recognition (CVPR)}, pages 224--233, 2020.

\bibitem{demo2vec}
Kuan Fang, Te-Lin Wu, Daniel Yang, Silvio Savarese, and Joseph~J. Lim.
\newblock Demo2vec: Reasoning object affordances from online videos.
\newblock In {\em Proceedings of the IEEE Conference on Computer Vision and
  Pattern Recognition (CVPR)}, June 2018.

\bibitem{feix2015grasp}
Thomas Feix, Javier Romero, Heinz-Bodo Schmiedmayer, Aaron~M Dollar, and Danica
  Kragic.
\newblock The grasp taxonomy of human grasp types.
\newblock {\em IEEE Transactions on human-machine systems}, 46(1):66--77, 2015.

\bibitem{fouhey2012people}
David~F Fouhey, Vincent Delaitre, Abhinav Gupta, Alexei~A Efros, Ivan Laptev,
  and Josef Sivic.
\newblock People watching: Human actions as a cue for single view geometry.
\newblock In {\em Proceedings of the European Conference on Computer Vision
  (ECCV)}, pages 732--745. Springer, 2012.

\bibitem{garcia2018first}
Guillermo Garcia-Hernando, Shanxin Yuan, Seungryul Baek, and Tae-Kyun Kim.
\newblock First-person hand action benchmark with rgb-d videos and 3d hand pose
  annotations.
\newblock In {\em Proceedings of the IEEE Conference on Computer Vision and
  Pattern Recognition (CVPR)}, pages 409--419, 2018.

\bibitem{ginosar2019learning}
Shiry Ginosar, Amir Bar, Gefen Kohavi, Caroline Chan, Andrew Owens, and
  Jitendra Malik.
\newblock Learning individual styles of conversational gesture.
\newblock In {\em Proceedings of the IEEE Conference on Computer Vision and
  Pattern Recognition (CVPR)}, pages 3497--3506, 2019.

\bibitem{grill2020byol}
Jean-Bastien Grill, Florian Strub, Florent Altch\'{e}, Corentin Tallec, Pierre
  Richemond, Elena Buchatskaya, Carl Doersch, Bernardo Avila~Pires, Zhaohan
  Guo, Mohammad Gheshlaghi~Azar, Bilal Piot, koray kavukcuoglu, Remi Munos, and
  Michal Valko.
\newblock Bootstrap your own latent - a new approach to self-supervised
  learning.
\newblock In {\em Advances in Neural Information Processing Systems (NeurIPS)},
  2020.

\bibitem{hamer2010handposeprior}
Henning Hamer, Juergen Gall, Thibaut Weise, and Luc Van~Gool.
\newblock An object-dependent hand pose prior from sparse training data.
\newblock In {\em Proceedings of the IEEE Conference on Computer Vision and
  Pattern Recognition (CVPR)}, pages 671--678, 2010.

\bibitem{hasson2019learning}
Yana Hasson, Gul Varol, Dimitrios Tzionas, Igor Kalevatykh, Michael~J Black,
  Ivan Laptev, and Cordelia Schmid.
\newblock Learning joint reconstruction of hands and manipulated objects.
\newblock In {\em Proceedings of the IEEE Conference on Computer Vision and
  Pattern Recognition (CVPR)}, pages 11807--11816, 2019.

\bibitem{he2017mask}
Kaiming He, Georgia Gkioxari, Piotr Doll{\'a}r, and Ross Girshick.
\newblock {Mask R-CNN}.
\newblock In {\em Proceedings of the IEEE Conference on Computer Vision and
  Pattern Recognition (CVPR)}, pages 2961--2969, 2017.

\bibitem{he2016deep}
Kaiming He, Xiangyu Zhang, Shaoqing Ren, and Jian Sun.
\newblock Deep residual learning for image recognition.
\newblock In {\em Proceedings of the IEEE Conference on Computer Vision and
  Pattern Recognition (CVPR)}, pages 770--778, 2016.

\bibitem{isola2015discovering}
Phillip Isola, Joseph~J Lim, and Edward~H Adelson.
\newblock Discovering states and transformations in image collections.
\newblock In {\em Proceedings of the IEEE Conference on Computer Vision and
  Pattern Recognition (CVPR)}, pages 1383--1391, 2015.

\bibitem{jiang2021graspTTA}
Hanwen Jiang, Shaowei Liu, Jiashun Wang, and Xiaolong Wang.
\newblock Hand-object contact consistency reasoning for human grasps
  generation.
\newblock In {\em Proceedings of the IEEE International Conference on Computer
  Vision (ICCV)}, 2021.

\bibitem{karunratanakul2020graspingFL}
Korrawe Karunratanakul, Jinlong Yang, Yan Zhang, Michael~J. Black, Krikamol
  Muandet, and Siyu Tang.
\newblock Grasping field: Learning implicit representations for human grasps.
\newblock {\em International Conference on 3D Vision (3DV)}, pages 333--344,
  2020.

\bibitem{kitani2012activity}
Kris~M Kitani, Brian~D Ziebart, James~Andrew Bagnell, and Martial Hebert.
\newblock Activity forecasting.
\newblock In {\em Proceedings of the European Conference on Computer Vision
  (ECCV)}, pages 201--214. Springer, 2012.

\bibitem{Kokic2020LearningTG}
Mia Kokic, Danica Kragic, and Jeannette Bohg.
\newblock Learning task-oriented grasping from human activity datasets.
\newblock {\em IEEE Robotics and Automation Letters}, 5:3352--3359, 2020.

\bibitem{koppula2014physically}
Hema~S Koppula and Ashutosh Saxena.
\newblock Physically grounded spatio-temporal object affordances.
\newblock In {\em Proceedings of the European Conference on Computer Vision
  (ECCV)}, pages 831--847. Springer, 2014.

\bibitem{deepgazeii}
Matthias Kummerer, Thomas S.~A. Wallis, Leon~A. Gatys, and Matthias Bethge.
\newblock Understanding low- and high-level contributions to fixation
  prediction.
\newblock In {\em Proceedings of the IEEE International Conference on Computer
  Vision (ICCV)}, Oct 2017.

\bibitem{Kwon_2021_ICCV}
Taein Kwon, Bugra Tekin, Jan St\"uhmer, Federica Bogo, and Marc Pollefeys.
\newblock H2o: Two hands manipulating objects for first person interaction
  recognition.
\newblock In {\em Proceedings of the IEEE/CVF International Conference on
  Computer Vision (ICCV)}, pages 10138--10148, October 2021.

\bibitem{lai2021functional}
Zihang Lai, Senthil Purushwalkam, and Abhinav Gupta.
\newblock The functional correspondence problem.
\newblock In {\em Proceedings of the IEEE Conference on Computer Vision and
  Pattern Recognition (CVPR)}, pages 15772--15781, 2021.

\bibitem{lenz2015robgrasp}
Ian Lenz, Honglak Lee, and Ashutosh Saxena.
\newblock Deep learning for detecting robotic grasps.
\newblock {\em The International Journal of Robotics Research},
  34(4-5):705--724, 2015.

\bibitem{sergey2018handeye}
Sergey Levine, Peter Pastor, Alex Krizhevsky, Julian Ibarz, and Deirdre
  Quillen.
\newblock Learning hand-eye coordination for robotic grasping with deep
  learning and large-scale data collection.
\newblock {\em The International Journal of Robotics Research},
  37(4-5):421--436, 2018.

\bibitem{li2019siamrpn++}
Bo Li, Wei Wu, Qiang Wang, Fangyi Zhang, Junliang Xing, and Junjie Yan.
\newblock Siamrpn++: Evolution of siamese visual tracking with very deep
  networks.
\newblock In {\em Proceedings of the IEEE Conference on Computer Vision and
  Pattern Recognition (CVPR)}, 2019.

\bibitem{li2020category}
Xiaolong Li, He Wang, Li Yi, Leonidas~J Guibas, A~Lynn Abbott, and Shuran Song.
\newblock Category-level articulated object pose estimation.
\newblock In {\em Proceedings of the IEEE Conference on Computer Vision and
  Pattern Recognition (CVPR)}, pages 3706--3715, 2020.

\bibitem{Mahler2017DexNet2D}
Jeffrey Mahler, Jacky Liang, Sherdil Niyaz, Michael Laskey, Richard Doan, Xinyu
  Liu, Juan~Aparicio Ojea, and Ken Goldberg.
\newblock Dex-net 2.0: Deep learning to plan robust grasps with synthetic point
  clouds and analytic grasp metrics.
\newblock In {\em Robotics: Science and Systems (RSS)}, 2017.

\bibitem{mandikal2020dexterous}
Priyanka Mandikal and Kristen Grauman.
\newblock Dexterous robotic grasping with object-centric visual affordances.
\newblock In {\em Proceedings of the IEEE International Conference on Robotics
  and Automation (ICRA)}, 2021.

\bibitem{mandikal2021dexvip}
Priyanka Mandikal and Kristen Grauman.
\newblock Dexvip: Learning dexterous grasping with human hand pose priors from
  video.
\newblock In {\em Proceedings of the Conference on Robot Learning (CoRL)},
  2021.

\bibitem{mo2021where2act}
Kaichun Mo, Leonidas Guibas, Mustafa Mukadam, Abhinav Gupta, and Shubham
  Tulsiani.
\newblock Where2act: From pixels to actions for articulated 3d objects.
\newblock {\em Proceedings of the IEEE International Conference on Computer
  Vision (ICCV)}, 2021.

\bibitem{mousavian2019graspnet}
Arsalan Mousavian, Clemens Eppner, and Dieter Fox.
\newblock 6-dof graspnet: Variational grasp generation for object manipulation.
\newblock In {\em Proceedings of the IEEE International Conference on Computer
  Vision (ICCV)}, 2019.

\bibitem{nagarajan2019grounded}
Tushar Nagarajan, Christoph Feichtenhofer, and Kristen Grauman.
\newblock Grounded human-object interaction hotspots from video.
\newblock In {\em Proceedings of the IEEE Conference on Computer Vision and
  Pattern Recognition (CVPR)}, pages 8688--8697, 2019.

\bibitem{nagarajan2021shaping}
Tushar Nagarajan and Kristen Grauman.
\newblock Shaping embodied agent behavior with activity-context priors from
  egocentric video.
\newblock In {\em Advances in Neural Information Processing Systems (NeurIPS)},
  2021.

\bibitem{nagarajan2020ego}
Tushar Nagarajan, Yanghao Li, Christoph Feichtenhofer, and Kristen Grauman.
\newblock {EGO-TOPO}: Environment affordances from egocentric video.
\newblock In {\em Proceedings of the IEEE Conference on Computer Vision and
  Pattern Recognition (CVPR)}, pages 163--172, 2020.

\bibitem{ng2020you2me}
Evonne Ng, Donglai Xiang, Hanbyul Joo, and Kristen Grauman.
\newblock You2me: Inferring body pose in egocentric video via first and second
  person interactions.
\newblock In {\em Proceedings of the IEEE Conference on Computer Vision and
  Pattern Recognition (CVPR)}, pages 9890--9900, 2020.

\bibitem{salgan}
Junting Pan, Cristian Canton{-}Ferrer, Kevin McGuinness, Noel~E. O'Connor,
  Jordi Torres, Elisa Sayrol, and Xavier Gir{\'{o}}{-}i{-}Nieto.
\newblock Salgan: Visual saliency prediction with generative adversarial
  networks.
\newblock {\em CoRR}, abs/1701.01081, 2017.

\bibitem{pathakCVPR16context}
Deepak Pathak, Philipp Kr\"ahenb\"uhl, Jeff Donahue, Trevor Darrell, and Alexei
  Efros.
\newblock Context encoders: Feature learning by inpainting.
\newblock In {\em Proceedings of the IEEE Conference on Computer Vision and
  Pattern Recognition (CVPR)}, 2016.

\bibitem{petrik2020learning}
Vladim{\'\i}r Petr{\'\i}k, Makarand Tapaswi, Ivan Laptev, and Josef Sivic.
\newblock Learning object manipulation skills via approximate state estimation
  from real videos.
\newblock {\em Proceedings of the Conference on Robot Learning (CoRL)}, 2020.

\bibitem{pinto2016SupersizingSL}
Lerrel Pinto and Abhinav Gupta.
\newblock Supersizing self-supervision: Learning to grasp from 50k tries and
  700 robot hours.
\newblock {\em Proceedings of the IEEE International Conference on Robotics and
  Automation (ICRA)}, pages 3406--3413, 2016.

\bibitem{rhinehart2017first}
Nicholas Rhinehart and Kris~M Kitani.
\newblock First-person activity forecasting with online inverse reinforcement
  learning.
\newblock In {\em Proceedings of the IEEE International Conference on Computer
  Vision (ICCV)}, pages 3696--3705, 2017.

\bibitem{rogez2015understanding}
Gr{\'e}gory Rogez, James~S Supancic, and Deva Ramanan.
\newblock Understanding everyday hands in action from {RGB-D} images.
\newblock In {\em Proceedings of the IEEE Conference on Computer Vision and
  Pattern Recognition (CVPR)}, pages 3889--3897, 2015.

\bibitem{schmeckpeper2020reinforcement}
Karl Schmeckpeper, Oleh Rybkin, Kostas Daniilidis, Sergey Levine, and Chelsea
  Finn.
\newblock Reinforcement learning with videos: Combining offline observations
  with interaction.
\newblock In {\em Proceedings of the Conference on Robot Learning (CoRL)},
  2020.

\bibitem{sermanet2018time}
Pierre Sermanet, Corey Lynch, Yevgen Chebotar, Jasmine Hsu, Eric Jang, Stefan
  Schaal, Sergey Levine, and Google Brain.
\newblock Time-contrastive networks: Self-supervised learning from video.
\newblock In {\em Proceedings of the IEEE International Conference on Robotics
  and Automation (ICRA)}, pages 1134--1141, 2018.

\bibitem{sermanet2016unsupervised}
Pierre Sermanet, Kelvin Xu, and Sergey Levine.
\newblock Unsupervised perceptual rewards for imitation learning.
\newblock In {\em Robotics: Science and Systems (RSS)}, 2017.

\bibitem{shan2020understanding}
Dandan Shan, Jiaqi Geng, Michelle Shu, and David Fouhey.
\newblock Understanding human hands in contact at internet scale.
\newblock In {\em Proceedings of the IEEE Conference on Computer Vision and
  Pattern Recognition (CVPR)}, 2020.

\bibitem{shan2021contrastive}
Dandan Shan, Richard~E.L. Higgins, and David~F. Fouhey.
\newblock {COHESIV}: Contrastive object and hand embedding segmentation in
  video.
\newblock In {\em Advances in Neural Information Processing Systems (NeurIPS)},
  2021.

\bibitem{sun2020hidden}
Jin Sun, Hadar Averbuch-Elor, Qianqian Wang, and Noah Snavely.
\newblock Hidden footprints: Learning contextual walkability from 3d human
  trails.
\newblock In {\em Proceedings of the European Conference on Computer Vision
  (ECCV)}, pages 192--207. Springer, 2020.

\bibitem{tekin2019h+}
Bugra Tekin, Federica Bogo, and Marc Pollefeys.
\newblock H+ o: Unified egocentric recognition of 3d hand-object poses and
  interactions.
\newblock In {\em Proceedings of the IEEE Conference on Computer Vision and
  Pattern Recognition (CVPR)}, pages 4511--4520, 2019.

\bibitem{thermos2021affordancevideo}
Spyridon Thermos, Gerasimos Potamianos, and Petros Daras.
\newblock Joint object affordance reasoning and segmentation in rgb-d videos.
\newblock {\em IEEE Access}, 9:89699--89713, 2021.

\bibitem{Wang2020novelobj}
Suchen Wang, Kim-Hui Yap, Junsong Yuan, and Yap-Peng Tan.
\newblock Discovering human interactions with novel objects via zero-shot
  learning.
\newblock In {\em Proceedings of the IEEE Conference on Computer Vision and
  Pattern Recognition (CVPR)}, 2020.

\bibitem{wang2017binge}
Xiaolong Wang, Rohit Girdhar, and Abhinav Gupta.
\newblock Binge watching: Scaling affordance learning from sitcoms.
\newblock In {\em Proceedings of the IEEE Conference on Computer Vision and
  Pattern Recognition (CVPR)}, pages 2596--2605, 2017.

\bibitem{xiang2018posecnn}
Yu Xiang, Tanner Schmidt, Venkatraman Narayanan, and Dieter Fox.
\newblock {PoseCNN}: A convolutional neural network for 6d object pose
  estimation in cluttered scenes.
\newblock In {\em Robotics: Science and Systems (RSS)}, 2018.

\bibitem{yang2015grasp}
Yezhou Yang, Cornelia Fermuller, Yi Li, and Yiannis Aloimonos.
\newblock Grasp type revisited: A modern perspective on a classical feature for
  vision.
\newblock In {\em Proceedings of the IEEE Conference on Computer Vision and
  Pattern Recognition (CVPR)}, pages 400--408, 2015.

\bibitem{zakka2021xirl}
Kevin Zakka, Andy Zeng, Pete Florence, Jonathan Tompson, Jeannette Bohg, and
  Debidatta Dwibedi.
\newblock Xirl: Cross-embodiment inverse reinforcement learning.
\newblock In {\em Proceedings of the Conference on Robot Learning (CoRL)},
  2021.

\bibitem{zhang2017split}
Richard Zhang, Phillip Isola, and Alexei~A Efros.
\newblock Split-brain autoencoders: Unsupervised learning by cross-channel
  prediction.
\newblock In {\em Proceedings of the IEEE Conference on Computer Vision and
  Pattern Recognition (CVPR)}, pages 1058--1067, 2017.

\end{thebibliography}
}

\onecolumn
\setlength{\ptcindent}{0pt}
\mtcsettitlefont{parttoc}{\large\bfseries}
\mtcsettitle{parttoc}{Human Hands as Probes for Interactive Object Understanding -- Supplementary Material}
\renewcommand \thepart{}
\renewcommand \partname{}
\part{}
\parttoc

\renewcommand\thefigure{S\arabic{figure}}

\renewcommand\thetable{S\arabic{table}}

\renewcommand\thesection{S\arabic{section}}

\section{State Sensitive Feature Learning}
\subsection{Model Details}

\subsubsection{Baseline Models}
We compare to six baseline models: ImageNet pre-trained model without any
further training, three self-supervised models (SimCLR \cite{chen2020simple},
TCN \cite{sermanet2018time}, and SimCLR+TCN), and two supervised models (action
classification on \epic, and MIT States supervision). All models are
initialized from ImageNet pre-training. 

All models use the \RS{18} backbone. We average pool the output after the
second last layer to obtain a 512 dimensional representation. The
self-supervised models use 3 projection layers with sizes 512, 512, 128.  The
128-dimensional output from the last layer is used for computing the
similarity. The semantically supervised models (\ie those trained on MIT States
dataset, or for action classification on the \epic dataset) only use a single
linear classifier layer directly on top of ImageNet features.  These additional
layers were thrown out and just the \RS{18} backbone is used for state
classification experiment on \SD. We only train a linear classifier on top of
the learned \RS{18} backbone for downstream state-classification on \epic
dataset.

We use a batch size of 256 for pre-trainings. We optimize using Adam
optimizer with a learning rate of $10^{-4}$. All models
(with the exception of the two supervised baselines) are trained for 200 epochs
and the last checkpoint is selected as the final model, which eliminates any
dependencies on the pre-training validation set.  All models are trained on a
single NVidia GPU (RTX A40 or equivalent). We next list method specific
hyper-parameters.

\begin{enumerate}
\item \textbf{ImageNet Pre-trained Model.}
This model has no additional hyper-parameters.

\item \textbf{SimCLR.}
We use the standard SimCLR augmentations in the following order: random resized
crop with scale (0.5, 1.0), random horizontal flip, random color jitter with
parameters (0.8, 0.8, 0.8, 0.2) and 80\% probability, random grayscale with
20\% probability, Gaussian blur with 12 size kernel and sigma set to (0.1,
2.0), and finally ImageNet normalization.

\item \textbf{TCN.} 
Let $w$ denote the length of the current track (number of frames). TCN's window
size is set to $\lfloor w/4 \rfloor$ and the negative sample is guaranteed to
be at least $\lceil w/2 \rceil$ (negative window) away from the positive
examples.  Sampling the positive and negatives on opposite ends of the track
ensures a large distance between them. TCN is optimized with a triplet margin
loss. Let us reuse $o_i,o_i'$ as the positive pair and define $o_i''$ as the
negative sample. Given an arbitrary margin $\alpha$ (in practice $\alpha=2$),
the triplet margin loss is as follows. We chose the hyper-parameters as suggested by \cite{sermanet2018time}. \begin{eqnarray} ||f_o(o_i) - f_o(o_i')||_2^2 +
\alpha < ||f_o(o_i) - f_o(o_i'')||_2^2. \end{eqnarray}

\item \textbf{SimCLR+TCN.}
We combine SimCLR and TCN, by a) sampling negative from both within and across
tracks, and b) using a NT-Xent loss from SimCLR~\cite{chen2020simple}. We also
adopt the image augmentations used in SimCLR.

\item \textbf{Action Classification.}
We train ResNet-18 (initialized from ImageNet) on 32 action labels along with
their temporal extent, available as part of the \epic dataset. These include:
{take, put, wash, open, close, insert, cut, pour, mix, turn-on, move,
remove, turn-off, dry, throw, shake, squeeze, adjust, peel, scoop, empty, flip,
fill, turn, check, spray, apply, pat, fold, scrape, sprinkle, break}. The model
samples two frames (in order) and uses them jointly to classify the action.
This allows the model to disambiguate between open and close actions.  The
model is trained for 30 epochs and we select the model which performed the best
on the action classification validation set. Validation performance peaked
within 30 epochs.

\item \textbf{MIT States.}
We train ResNet-18 (initialized from ImageNet) on the MIT-States attributes
dataset \cite{isola2015discovering}. The dataset consists of 115 classes and
approximately 53K images. Examples of attributes include mossy, deflated,
dirty, \etc This model is trained for 20 epochs and we select the model which
performed the best on the MIT-States validation set. Validation performance
peaked within 20 epochs.
\end{enumerate}

\subsubsection{TSC and TSC+OHC Models}
For TSC, object crops are selected by randomly sampling $o'_i$ such
that $o'_i$ is no more than $\lfloor w/4 \rfloor$ away from $o_i$ in the track,
where $w$ is the length of the track.
    
For TSC+OHC, we use two separate models, one each for the object and the hand.
The object model itself has 2 heads, one is used for the object-object
similarity for $\Lt$, and another for object-hand similarity for $\Lh$.  The
hand model only has one head. The hand model has additional layers to produce
and combine the positional encodings that represent motion.  The positional
encoding is generated by alternating sines and cosines over 12 frequencies for
each element of $h_i^m$. It is concatenated with the output of the \RS{18}
backbone. These combined features are projected to 512 dimensions with another
linear layer and finally fed through the hand model's loss head.  Note that the
object and hand crop fed through this model are not augmented with random
horizontal flip to preserve handedness.

For the TSC+OHC model, tracks are independently sampled for computing the $\Lt$
and $\Lh$ losses. Tracks with less than 4 frames of hands were filtered out to
remove noise, which led to a pre-training dataset size of 53,661 tracks. The
evaluation scheme remains the same as TSC since we only use the features
learned by the object model. We throw out the hand model.

\noindent \textbf{Loss Functions.} 
Recall that $h^a_i$ and $h^m_i$ jointly represent the hand: $h^a_i$ describes
the appearance and $h^m_i$ describes the motion. To detail $h^m_i$, consider
the object bounding box for $o_i$ defined as $(o_{i,x1}, o_{i,y1}, o_{i,x2},
o_{i,y2})$ where $(o_{i,x1}, o_{i,y1}),(o_{i,x2},o_{i,y2})$ are the coordinates
of the top left and bottom right corner of the bounding box, respectively. We
define the hand crop bounding box similarly: $(h_{k,x1}, h_{k,y1}, h_{k,x2},
h_{k,y2})$ where $k$ is uniformly sampled at random such that $|k-i| \leq 3$.
$(*_H, *_W)$ represent the height and width of the bounding box
and $(*_{xc}, *_{yc})$ refer to the center coordinates of the bounding box. For
an arbitrary $p$, we calculate $h^m_{i,p}$ and $h^m_{i}$ as follows:
\begin{eqnarray}
h^m_{i,p} &=& \left[\frac{o_{i,xc} - h_{p,xc}}{o_{i,W}}, \frac{o_{i,yc} -
h_{p,yc}}{o_{i,H}}, \frac{h_{p,W}}{o_{i,W}}, \frac{h_{p,H}}{o_{i,H}}\right] \\
h^m_{i} &=& \left[h^m_{i,k-1}, \quad h^m_{i,k}, \quad h^m_{i,k+1}\right]
\end{eqnarray}

Assuming that all crops ($o_i, o'_i, h^a_i$) are transformed using the standard
SimCLR augmentations, we describe our application of normalized temperature scaled cross-entropy loss (NT-Xent) \cite{chen2020simple} below:
\begin{eqnarray}
s^{oo'}_{i,j} &=& \nicefrac{1}{\tau} \cdot {\Sim(f_o(o_i), f_o(o'_j))} \\
s^{oh}_{i,j} &=& \nicefrac{1}{\tau} \cdot {\Sim(f_h(o_i), g_h(h_j))} \\
s^{hh}_{i,j} &=& \nicefrac{1}{\tau} \cdot {\Sim(g_h(h_i), g_h(h_j))}
\end{eqnarray}

where $\Sim$ refers to the cosine similarity, and $\tau$ is the temperature
parameter (in practice $\tau=0.1$). Then $\Lt$ and $\Lh$ losses are computed
using $s^{oh}$, $s^{hh}$, $s^{oo}$ as follows:
\begin{eqnarray}
\Lt &=& -\sum_{i}{
\left(
  \log 
    \dfrac
      {\exp(s^{oo'}_{i,i})}
      {\sum_{j:j \neq i} \exp(s^{oo}_{i,j}) + \sum_{k} \exp(s^{oo'}_{i,k})} 
\right)}
  - \sum_{i}{
    \left(
      \log 
        \dfrac
          {\exp(s^{oo'}_{i,i})}
          {\sum_{j:j \neq i} \exp(s^{o'o'}_{i,j}) + \sum_{k} \exp(s^{oo'}_{k,i})} 
    \right)} \\
    \Lh &=& -\sum_{i}{
    \left(
      \log 
        \dfrac
          {\exp(s^{oh}_{i,i})}
          {\sum_{j:j\neq i} \exp(s^{oo}_{i,j}) + \sum_{k} \exp(s^{oh}_{i,k})} 
    \right)}
    - \sum_{i}{
    \left(
      \log 
        \dfrac
          {\exp(s^{oh}_{i,i})}
          {\sum_{j:j\neq i} \exp(s^{hh}_{i,j}) + \sum_{k} \exp(s^{oh}_{k,i})} 
    \right)}.
\end{eqnarray}

\subsection{EPIC-STATES Dataset and State Classification Task}
\seclabel{epic-states}
\subsubsection{Data Annotation}
\SD is collected on top of the {\it ground-truth} object-of-interaction tracks
and corresponding object category labels from Damen
\etal~\cite{damen2020collection}.  We filter out 13 object categories of interest:
drawer, knife, spoon, cupboard, fridge, onion, fork, egg, potato, bottle,
microwave/oven, carrot, and jar. We chose a maximum of 5 frames from each
track. These object crops are then annotated for states individually for each object category.

We used a commercial service to obtain annotations for our dataset. Each image
was annotated once and then reviewed by a high-quality annotator (determined
using the accuracy on the task). We also included an \ttt{ambiguous} class and
reject images with the \ttt{ambiguous} label, resulting in 14,346
annotated images.
\vspace{10pt}

\noindent \textbf{Annotation Instruction.} We gave the annotators the following
instructions:
\begin{quotation}
Given an image, choose all applicable
categories/states from the ones available. If the image is considerably noisy
or the object of specified category cannot be identified, annotate the image as
ambiguous. When multiple objects are visible, annotate the most dominant object
of the specified category. Note that images are captured in-the-wild and small
motion blur, therefore, should not be considered as noise.
\end{quotation}
For each object category, we specify the set of states to consider, and any
other object category specific instructions. Below we club the instructions for
multiple objects, but note that images from different object categories were
annotated \textit{separately}. 

\begin{enumerate}
\item \textbf{Microwave/Oven, Cupboard, Drawer, and Fridge}:
The applicable states are \ttt{open, close}, and \ttt{ambiguous}. From
\ttt{open/close} only one state would be applicable, \ie a drawer can not be
both, \ttt{open} and \ttt{close} at the same time. 
\item \textbf{Jar and Bottle}: The applicable states are
\ttt{inhand, outofhand, open, close}, and \ttt{ambiguous}. From
\ttt{inhand/outofhand}, and \ttt{open/close} only one state would be
applicable, \ie a bottle can not be both, \ttt{inhand} and \ttt{outofhand}; or
both, \ttt{open} and \ttt{close}. 
\item \textbf{Onion and Potato}: The applicable states are
\ttt{inhand, outofhand, raw, cooked, whole, cut, peeled, unpeeled}, and
\ttt{ambiguous}. From \ttt{inhand/outofhand}, \ttt{raw / cooked}, \ttt{peeled /
unpeeled}, and \ttt{whole / cut} only one state would be applicable. For green
onions, we asked the annotators to not label the \ttt{peeled/unpeeled}
attribute.  \item \textbf{Carrot}: The applicable states are
\ttt{inhand,
outofhand, raw, cooked, whole, cut}, and \ttt{ambiguous}. From
\ttt{inhand/outofhand}, \ttt{raw/cooked}, and \ttt{whole/cut} only one state
would be applicable.  \item \textbf{Spoon, Fork, and Knife}: The
applicable states are \ttt{inhand, outofhand}, and \ttt{ambiguous}. From
\ttt{inhand/outofhand}, only one state would be applicable. 
\item \textbf{Egg}: The applicable states are \ttt{inhand,
outofhand, raw, cooked}, and \ttt{ambiguous}. From \ttt{inhand/outofhand},
\ttt{raw/cooked}, only one state would be applicable.
\end{enumerate}

\subsubsection{Dataset Statistics}
\noindent \textbf{Splits.}
We split the dataset into train, val and test splits based on participants. See
participant assignment to the different splits in \tableref{epicstates-pids}.
Participants were assigned between val and test by minimizing the
difference in joint object state (fridge-open, onion-cut, \etc) distribution
between the sets. This ensures a good split of both objects and states.

\noindent \textbf{Novel Object Categories.} We list object categories that were
used for training and testing for the novel object category experiment in
\tableref{epicstates-novel-objects}.

\vspace{10pt}

\noindent 
\renewcommand{\arraystretch}{1.1}
\begin{minipage}[t]{0.48\textwidth}
\centering
\captionof{table}{\sd participants in each data split.}
\tablelabel{epicstates-pids}
\resizebox{1.0\linewidth}{!}{
\begin{tabular}{ll}
\toprule
\textbf{Split} & \textbf{Participants}  \\
\midrule
Train       & P01, P03, P06, P08, P13, P17, P21, P25, P26, P29 \\
Validation  & P04, P05, P07, P14, P22, P23, P27 \\
Test        & P02,  P10, P12, P15,  P16,  P19, P20, P24, P28, P30, P31 \\
\bottomrule
\end{tabular}}
\end{minipage}
\hfill
\begin{minipage}[t]{0.48\textwidth}
\centering
\captionof{table}{\textbf{Novel Category Experiment.} For the novel category
experiment, we limited the training to objects in the first row, and evaluated
on categories in the second row.}
\tablelabel{epicstates-novel-objects}
\resizebox{1.0\linewidth}{!}{
\begin{tabular}{ll}
\toprule
\textbf{Train Objects} & fridge, knife, drawer, potato, carrot, jar, egg \\
\textbf{Novel Objects} & spoon, cupboard, onion, fork, microwave / oven, bottle \\
\bottomrule
\end{tabular}}
\end{minipage}
 \vspace{10pt}

\noindent \textbf{Object and State Distributions.}
\tableref{epicstates-states} and \tableref{epicstates-objects} shows the
distribution of states and objects in \SD, respectively. We also show the joint
distribution of objects and states in \tableref{epicstates-objectstates} across
the entire dataset. As noted, many different states are applicable to the same
object. 
\vspace{10pt}

\begin{table}[h]
\renewcommand{\arraystretch}{1.1}
\centering
\caption{\textbf{Objects in \SD.} For each objects in \SD, we list the applicable states and how many instances we have for that object in each split.}
\tablelabel{epicstates-objects}
\resizebox{\linewidth}{!}{
\begin{tabular}{lccccc}
\toprule
\textbf{Object} & \textbf{Applicable States} & \textbf{Train} & \textbf{Val} & \textbf{Test} & \textbf{Total} \\
\midrule
fridge          & \ttt{open}, \ttt{close}              & 779 & 448 & 732 & 1959 \\
spoon           & \ttt{inhand}, \ttt{outofhand}        & 751 & 482 & 717 & 1950 \\
knife           & \ttt{inhand}, \ttt{outofhand}        & 749 & 551 & 729 & 2029 \\
cupboard        & \ttt{open}, \ttt{close}              & 683 & 449 & 429 & 1561 \\
drawer          & \ttt{open}, \ttt{close}              & 681 & 666 & 446 & 1793 \\
onion           & \ttt{inhand}, \ttt{outofhand}, \ttt{raw}, \ttt{cooked}, \ttt{whole}, \ttt{cut}, \ttt{peeled}, \ttt{unpeeled}                 & 487 & 337 & 474 & 1298 \\
fork            & \ttt{inhand}, \ttt{outofhand}      & 353 & 206 & 259 & 818 \\
microwave/oven  & \ttt{open},  \ttt{close}            & 306 & 118 & 179 & 603 \\
bottle          & \ttt{open}, \ttt{close}, \ttt{inhand}, \ttt{outofhand}              & 294 & 179 & 257 & 730 \\
potato          & \ttt{inhand}, \ttt{outofhand}, \ttt{raw}, \ttt{cooked}, \ttt{whole}, \ttt{cut}, \ttt{peeled}, \ttt{unpeeled}                 & 164 & 192 & 182 & 538 \\
carrot          & \ttt{inhand}, \ttt{outofhand}, \ttt{raw}, \ttt{whole}, \ttt{cut}    & 97  & 114 & 196 & 407 \\
jar             & \ttt{open}, \ttt{close}, \ttt{inhand}, \ttt{outofhand}              & 54  & 99  & 89  & 242 \\
egg             & \ttt{inhand}, \ttt{outofhand},\ttt{raw}, \ttt{cooked}               & 32  & 199 & 187 & 418 \\
\bottomrule
\end{tabular}}
\end{table}
 
\renewcommand{\arraystretch}{1.1}
\begin{table}[h]
\setlength{\tabcolsep}{10pt}
\centering
\caption{\textbf{States in \SD.} For each state in \SD, we list the object
categories it is applicable to, and how many instances we have for that state 
in each split.}
\tablelabel{epicstates-states}
\resizebox{1.0\linewidth}{!}{
\begin{tabular}{lccccc}
\toprule
\textbf{State} & \textbf{Applicable Objects} & \textbf{Train} & \textbf{Val} & \textbf{Test} & \textbf{Total} \\
\midrule
\ttt{inhand}          & ~~~~~~~~~~~~bottle, carrot, egg, fork, jar, knife, onion, potato, spoon~~~~~~~~~~ & 1861                                                        & 1236 & 1699 & 4796 \\
\ttt{open}            & bottle, jar, cupboard, drawer, fridge, microwave / oven     & 2099 & 1440 & 1459    & 4998 \\
\ttt{outofhand}       & bottle, carrot, egg, fork, jar, knife, onion, potato, spoon & 1280 & 1112 & 1227    & 3619 \\
\ttt{raw}             & carrot, egg, onion, potato                                  & 623  & 561  & 783     & 1967 \\
\ttt{close}           & bottle, jar, cupboard, drawer, fridge, microwave / oven     & 686  & 496  & 633     & 1815 \\
\ttt{cut}             & carrot, onion, potato                                       & 477  & 459  & 499     & 1435 \\
\ttt{peeled}          & onion, potato                                               & 450  & 423  & 465     & 1338 \\
\ttt{whole}           & carrot, onion, potato                                       & 270  & 178  & 351     & 799 \\
\ttt{cooked}          & egg, onion, potato                                          & 152  & 273  & 245     & 670 \\
\ttt{unpeeled}        & onion, potato                                               & 197  & 104  & 186     & 487 \\
\bottomrule
\end{tabular}}
\end{table}

\subsubsection{State Classification Task}
We construct binary state classification tasks by considering all {\it
non-ambiguous} crops from object categories that afford the particular state
label, as noted in \tableref{epicstates-states}. We throw out images that
were \ttt{ambiguous} overall for all categories, or were \ttt{ambiguous} for
the specific state category under consideration.

\renewcommand{\arraystretch}{1.1}
\begin{table*}[!h]
\centering
\caption{Joint object and state distribution for \SD. Note that multiple states
are applicable to objects.}
\tablelabel{epicstates-objectstates}
\resizebox{1.0\linewidth}{!}{
\begin{tabular}{lcccccccccc}
\toprule
 & \textbf{\ttt{open}} & \textbf{\ttt{close}} & \textbf{\ttt{inhand}} & \textbf{\ttt{outofhand}} & \textbf{\ttt{raw}} & \textbf{\ttt{cooked}} & \textbf{\ttt{whole}} & \textbf{\ttt{cut}} & \textbf{\ttt{peeled}} & \textbf{\ttt{unpeeled}} \\
\midrule
bottle               & 286    & 358    & 521    & 205    & \xmark & \xmark & \xmark & \xmark & \xmark & \xmark \\
carrot               & \xmark & \xmark & 223    & 184    & 399    & \xmark & 272    & 132    & \xmark & \xmark \\
egg                  & \xmark & \xmark & 118    & 300    & 210    & 204    & \xmark & \xmark & \xmark & \xmark \\
fork                 & \xmark & \xmark & 525    & 293    & \xmark & \xmark & \xmark & \xmark & \xmark & \xmark \\
jar                  & 136    & 103    & 154    & 78     & \xmark & \xmark & \xmark & \xmark & \xmark & \xmark \\
knife                & \xmark & \xmark & 1331   & 699    & \xmark & \xmark & \xmark & \xmark & \xmark & \xmark \\
onion                & \xmark & \xmark & 411    & 887    & 1000   & 296    & 290    & 1005   & 992    & 303 \\
potato               & \xmark & \xmark & 236    & 300    & 358    & 170    & 237    & 294    & 346    & 184 \\
spoon                & \xmark & \xmark & 1277   & 673    & \xmark & \xmark & \xmark & \xmark & \xmark & \xmark \\
cupboard             & 1262   & 310    & \xmark & \xmark & \xmark & \xmark & \xmark & \xmark & \xmark & \xmark \\
drawer               & 1479   & 315    & \xmark & \xmark & \xmark & \xmark & \xmark & \xmark & \xmark & \xmark \\
fridge               & 1505   & 456    & \xmark & \xmark & \xmark & \xmark & \xmark & \xmark & \xmark & \xmark \\
microwave/oven~~~~~~ & 330    & 273    & \xmark & \xmark & \xmark & \xmark & \xmark & \xmark & \xmark & \xmark \\
\bottomrule
\end{tabular}}
\end{table*}

\begin{figure*}
\setlength{\tabcolsep}{2pt}
\centering
\resizebox{1\linewidth}{!}{
\begin{tabular}{rcccccccccc}

\rotatebox[origin=l]{90}{\ttt{open}}  &
\includegraphics[width=0.15\textwidth]{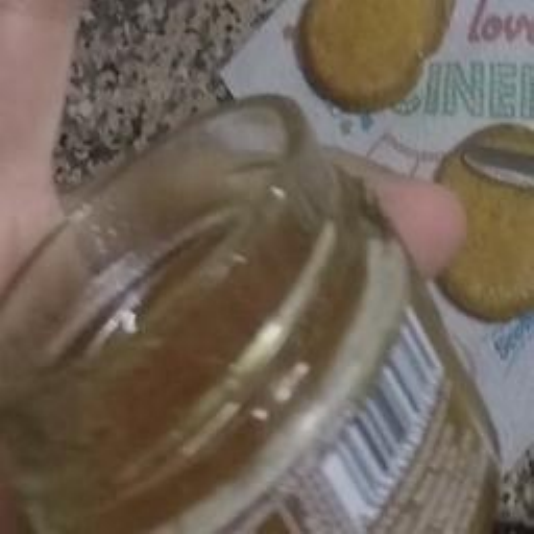} &
\includegraphics[width=0.15\textwidth]{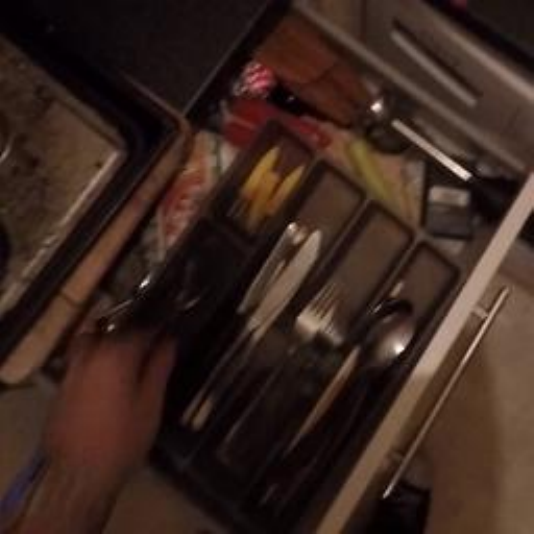} &
\includegraphics[width=0.15\textwidth]{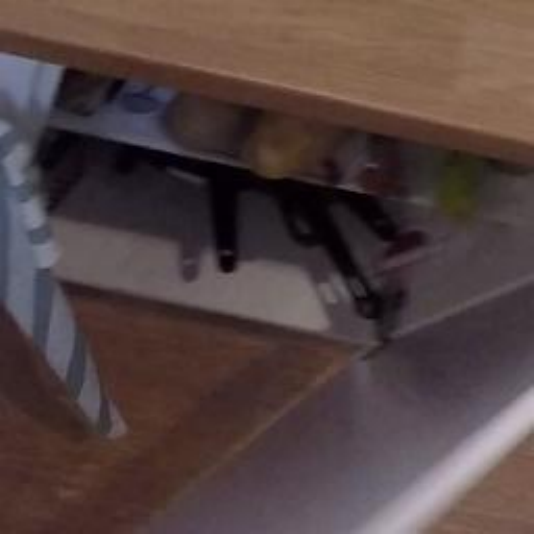} &
\includegraphics[width=0.15\textwidth]{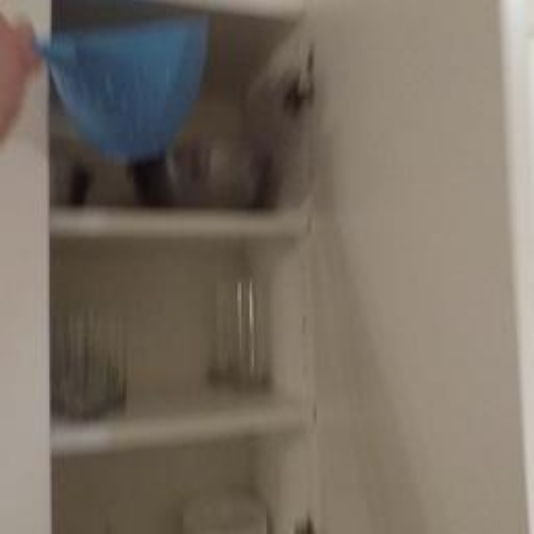} &

\rotatebox[origin=l]{90}{\ttt{close}}  &
\includegraphics[width=0.15\textwidth]{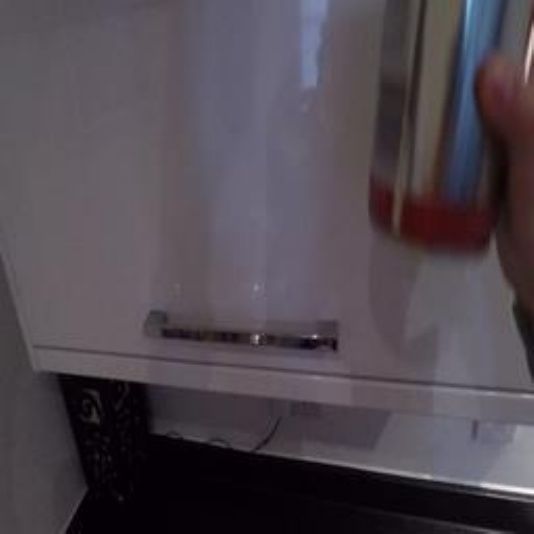} &
\includegraphics[width=0.15\textwidth]{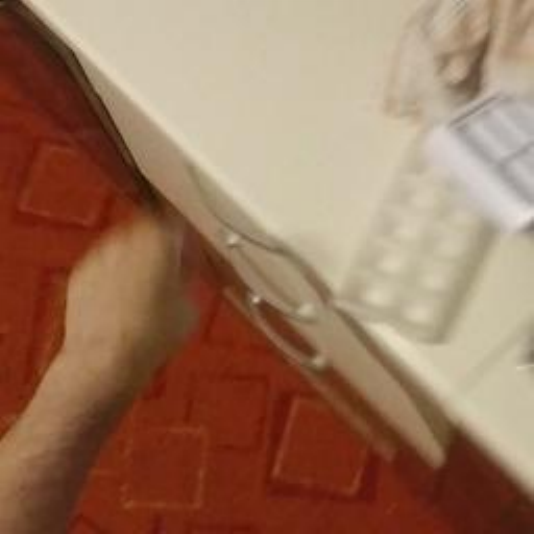} &
\includegraphics[width=0.15\textwidth]{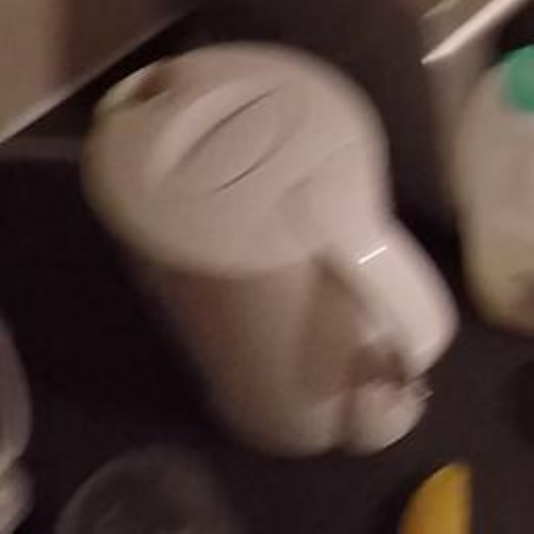} &
\includegraphics[width=0.15\textwidth]{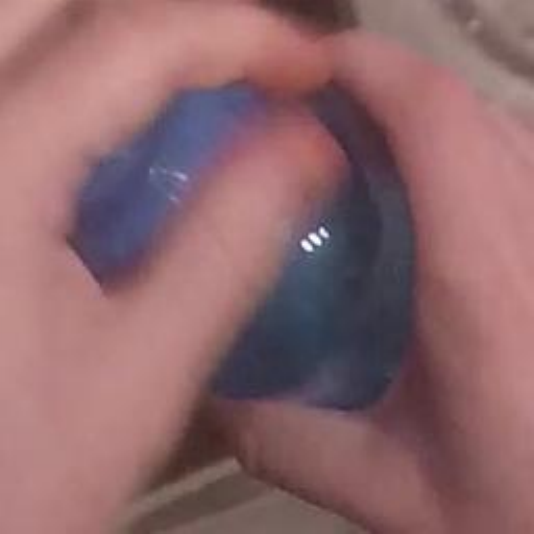} \\

\rotatebox[origin=l]{90}{\ttt{peeled}}  &
\includegraphics[width=0.15\textwidth]{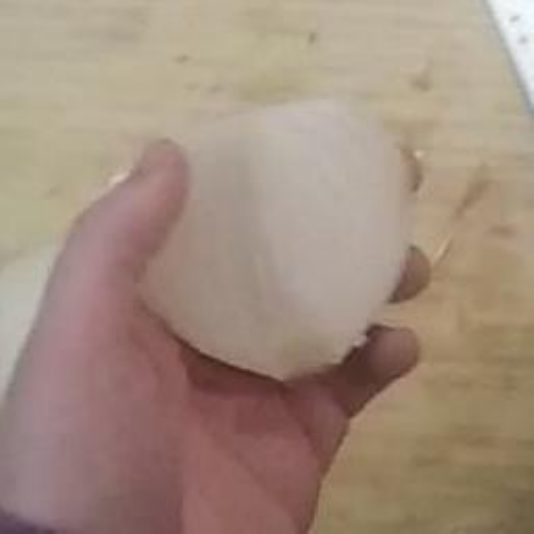} &
\includegraphics[width=0.15\textwidth]{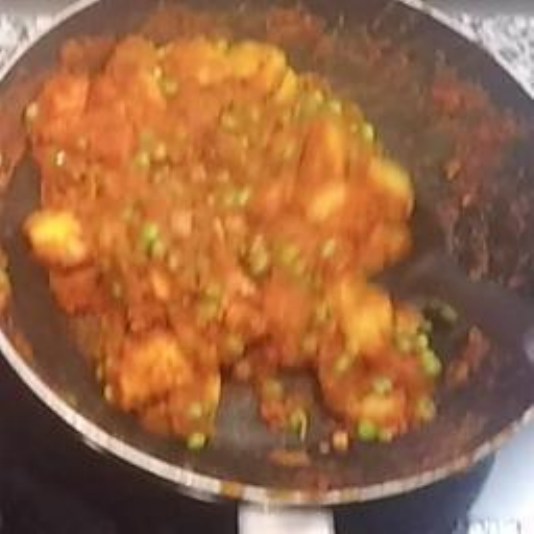} &
\includegraphics[width=0.15\textwidth]{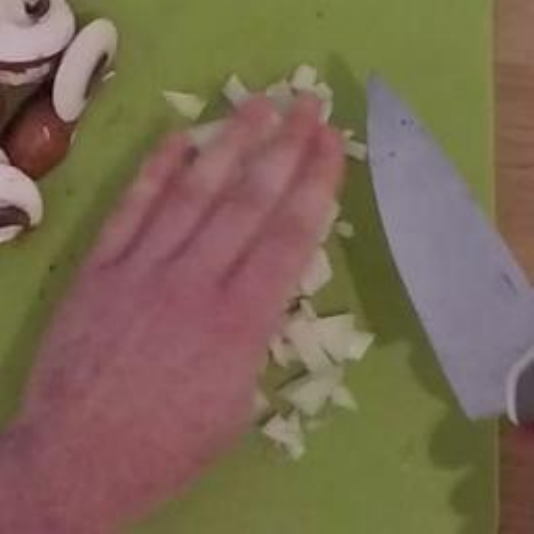} &
\includegraphics[width=0.15\textwidth]{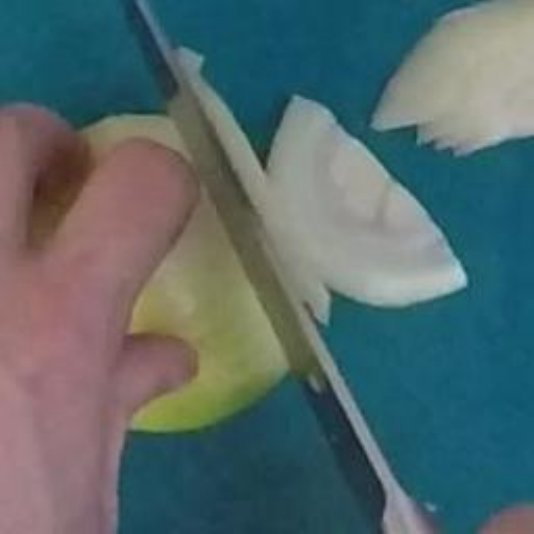} &

\rotatebox[origin=l]{90}{\ttt{unpeeled}}  &
\includegraphics[width=0.15\textwidth]{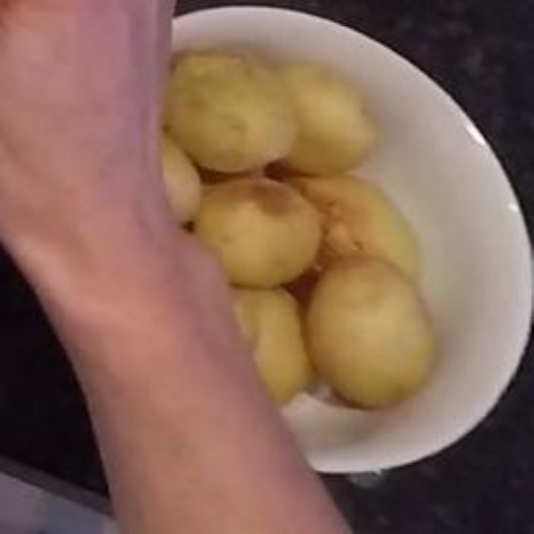} &
\includegraphics[width=0.15\textwidth]{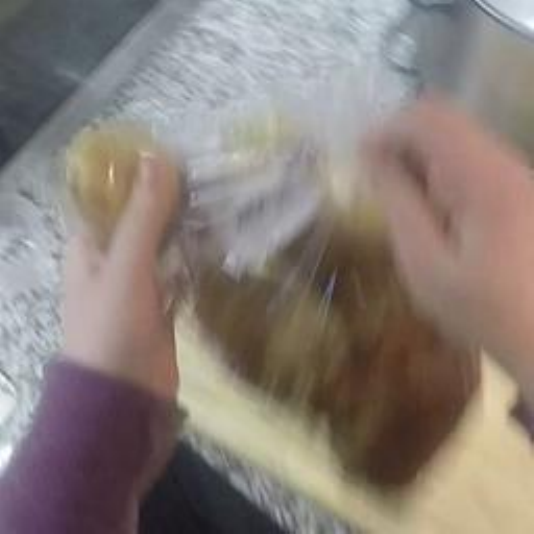} &
\includegraphics[width=0.15\textwidth]{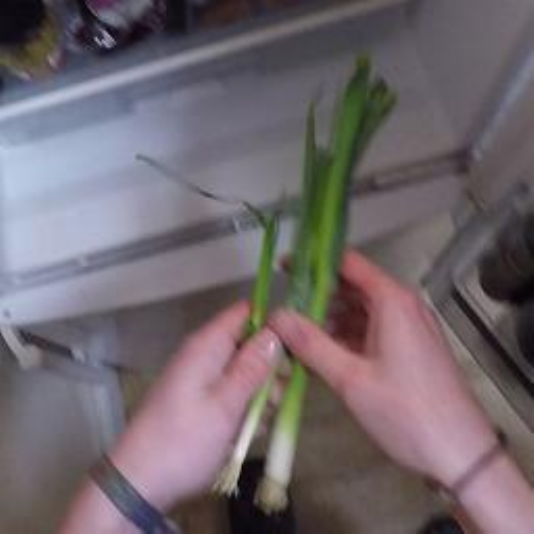} &
\includegraphics[width=0.15\textwidth]{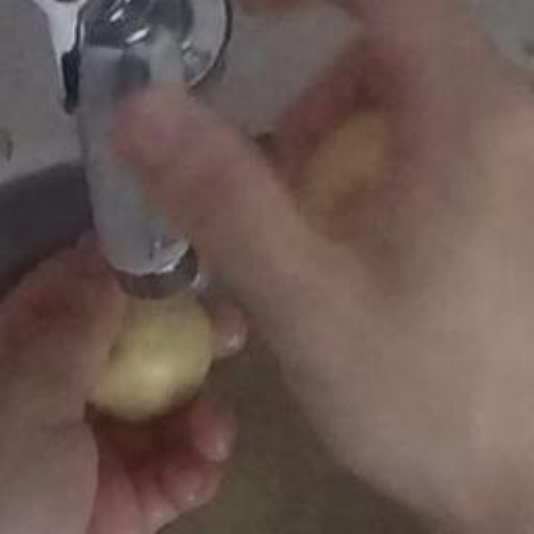} \\

\rotatebox[origin=l]{90}{\ttt{raw}}  &
\includegraphics[width=0.15\textwidth]{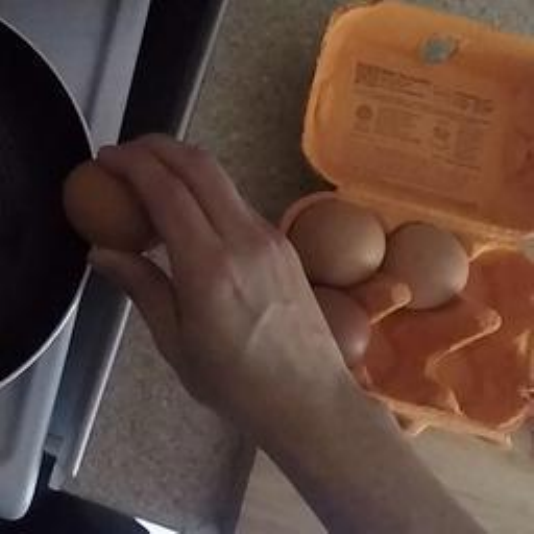} &
\includegraphics[width=0.15\textwidth]{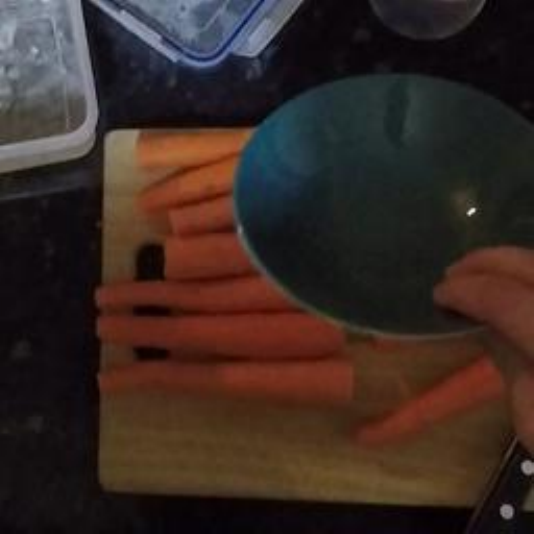} &
\includegraphics[width=0.15\textwidth]{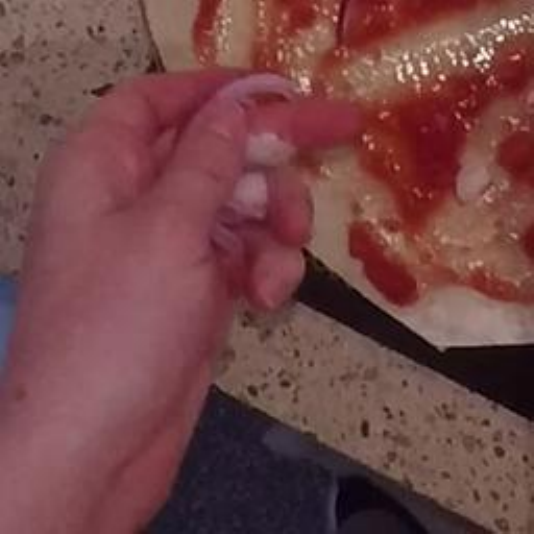} &
\includegraphics[width=0.15\textwidth]{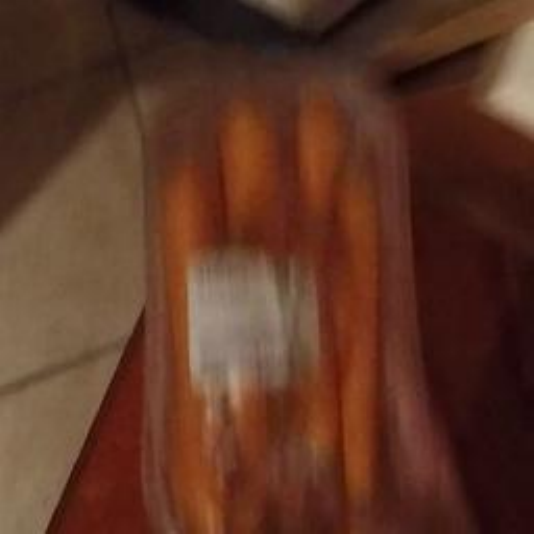} &

\rotatebox[origin=l]{90}{\ttt{cooked}}  &
\includegraphics[width=0.15\textwidth]{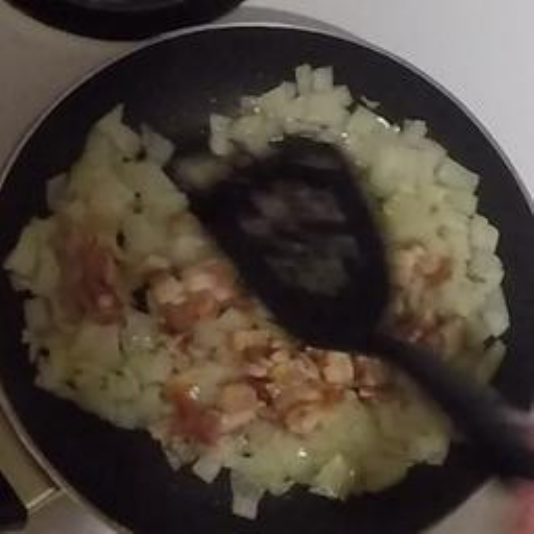} &
\includegraphics[width=0.15\textwidth]{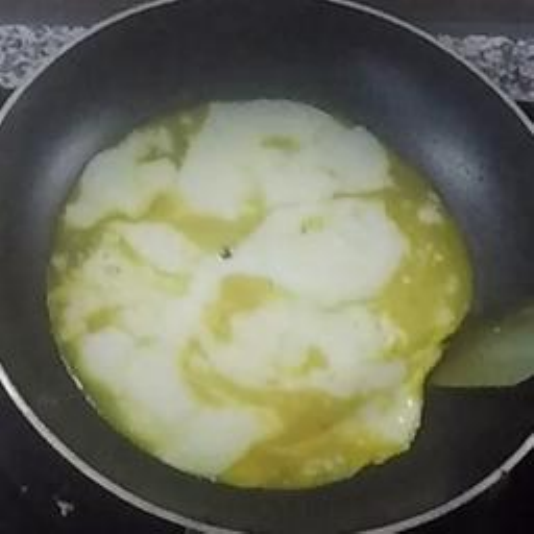} &
\includegraphics[width=0.15\textwidth]{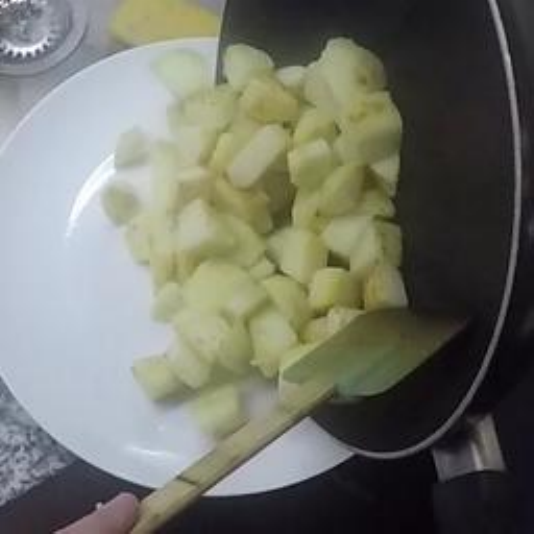} &
\includegraphics[width=0.15\textwidth]{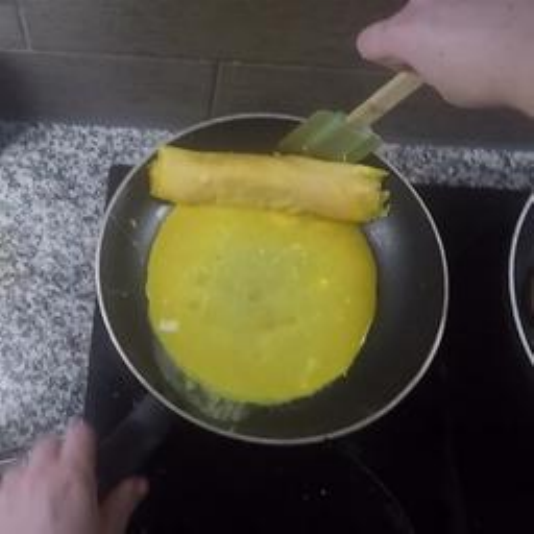} \\

\rotatebox[origin=l]{90}{\ttt{whole}}  &
\includegraphics[width=0.15\textwidth]{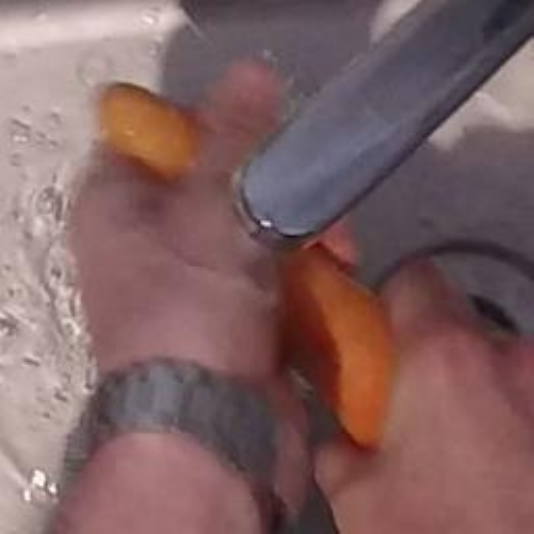} &
\includegraphics[width=0.15\textwidth]{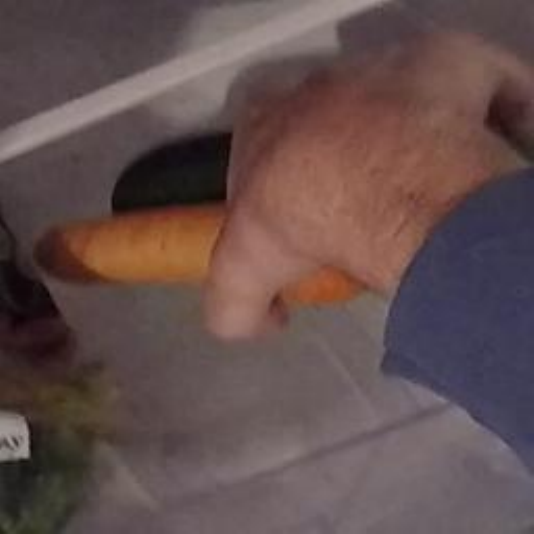} &
\includegraphics[width=0.15\textwidth]{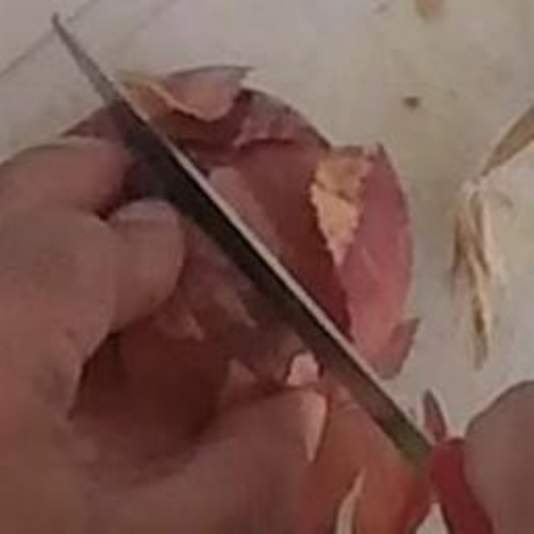} &
\includegraphics[width=0.15\textwidth]{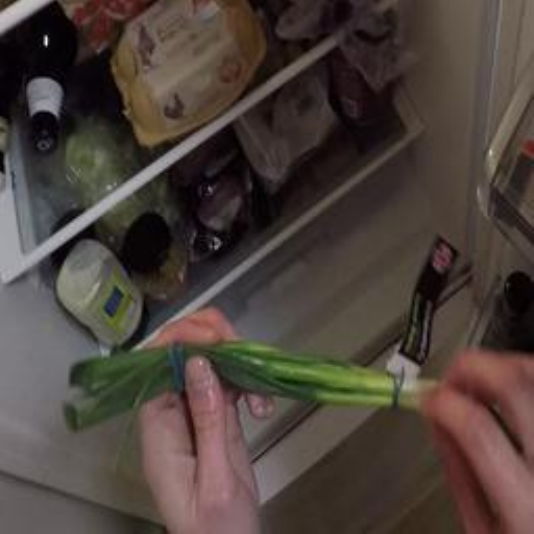} &

\rotatebox[origin=l]{90}{\ttt{cut}}  &
\includegraphics[width=0.15\textwidth]{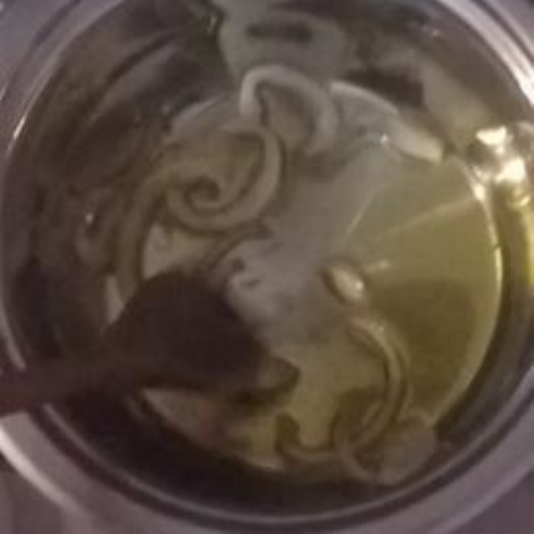} &
\includegraphics[width=0.15\textwidth]{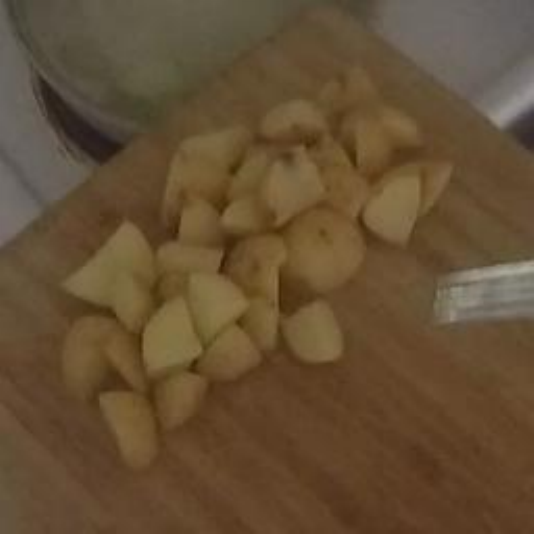} &
\includegraphics[width=0.15\textwidth]{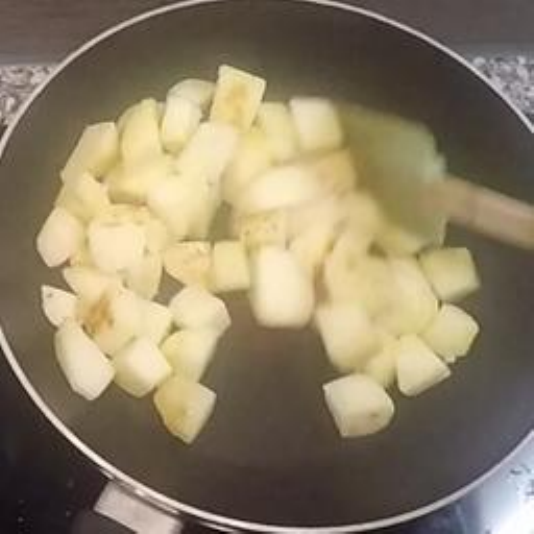} &
\includegraphics[width=0.15\textwidth]{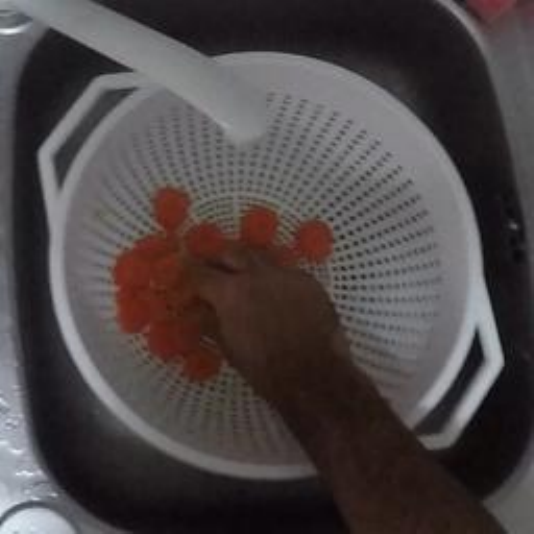} \\

\rotatebox[origin=l]{90}{\ttt{inhand}}  &
\includegraphics[width=0.15\textwidth]{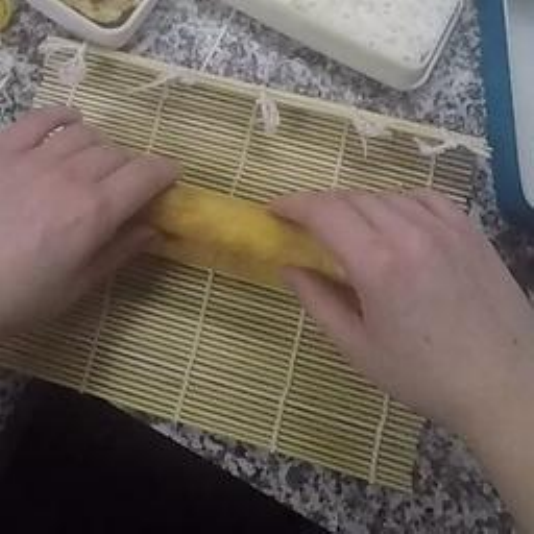} &
\includegraphics[width=0.15\textwidth]{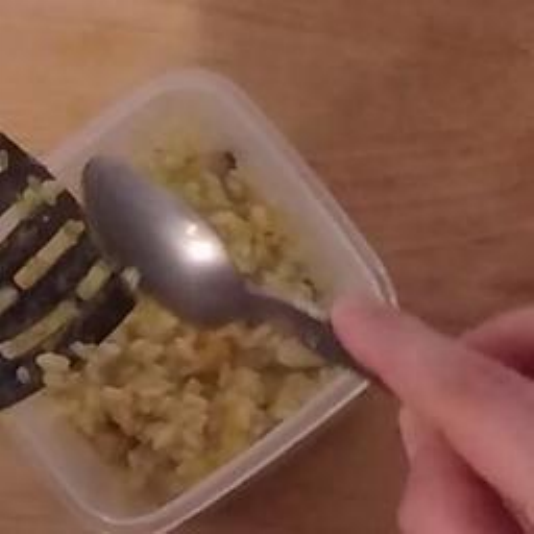} &
\includegraphics[width=0.15\textwidth]{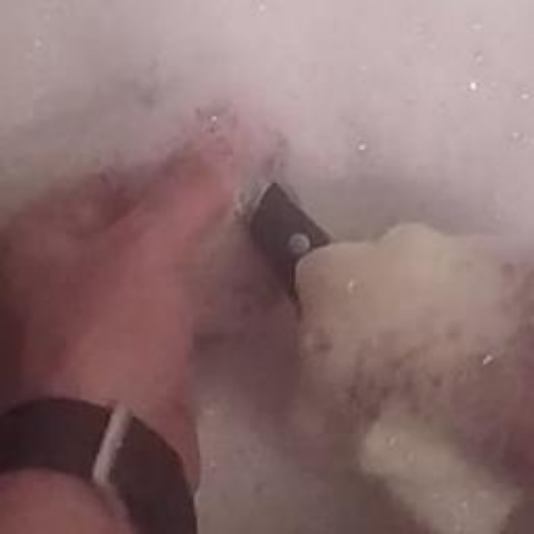} &
\includegraphics[width=0.15\textwidth]{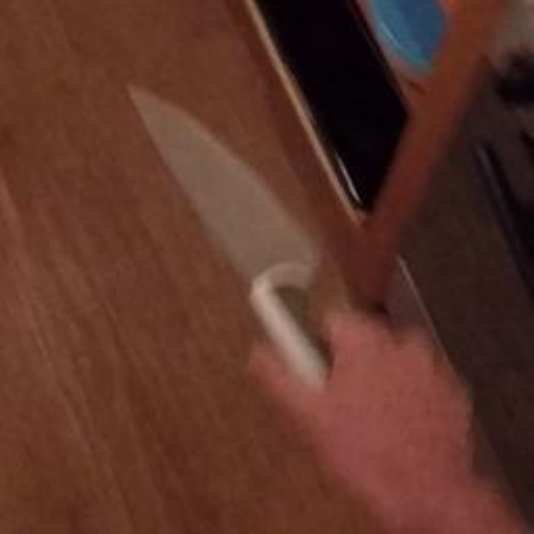} &

\rotatebox[origin=l]{90}{\ttt{outofhand}}  &
\includegraphics[width=0.15\textwidth]{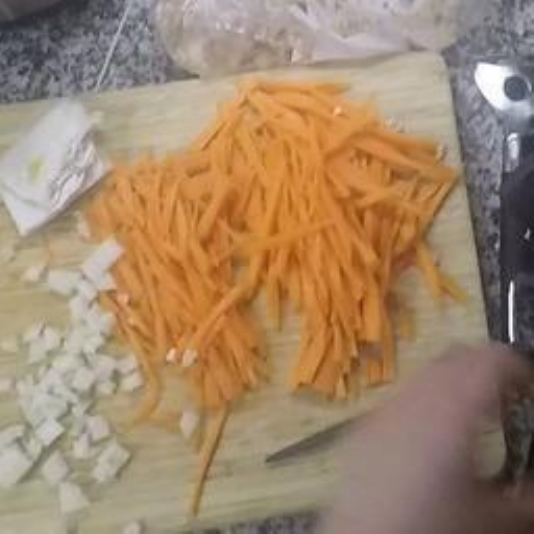} &
\includegraphics[width=0.15\textwidth]{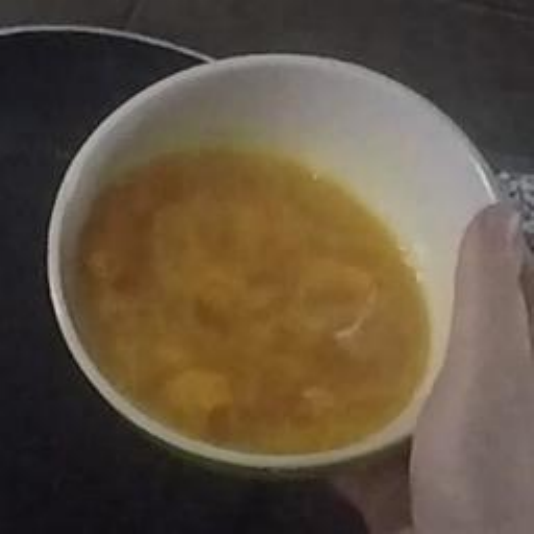} &
\includegraphics[width=0.15\textwidth]{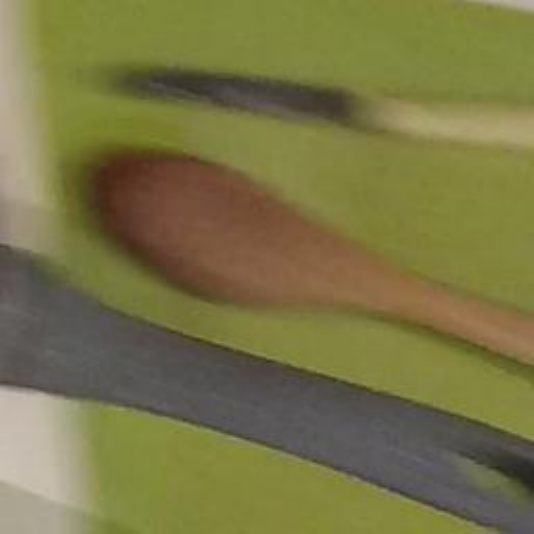} &
\includegraphics[width=0.15\textwidth]{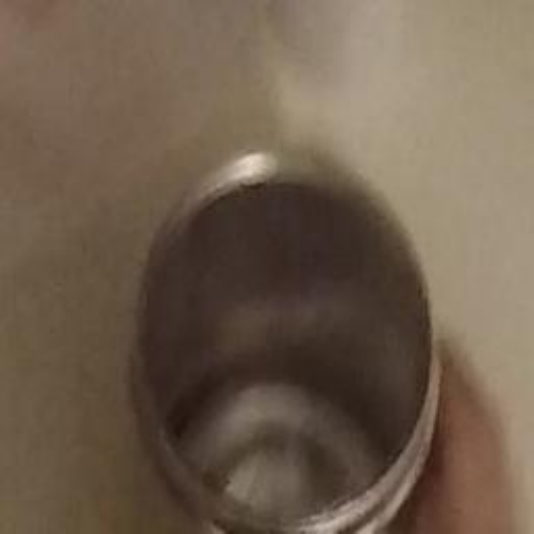} \\

\end{tabular}
}
\caption{Image samples corresponding to the different states in the \SD dataset.}
\label{epic-states}
\end{figure*}
  
\cleardoublepage
\subsection{Detailed Results and Ablations}

\subsubsection{State-wise Performance}
State-wise performance of considered methods for state classification is shown
in \tableref{epicstates-classwise-results}. In particular, we see that
\RS{18} without any feature learning on our tracks already performs well on
the \ttt{open}, \ttt{raw}, \ttt{cooked}, \ttt{peeled}, and \ttt{cut}
categories for most settings. Nonetheless, our methods improve performance
across the board. TSC+OHC improves upon TSC on all but one category in the
challenging setting of novel categories with limited data.

\vspace{5pt}
\renewcommand{\arraystretch}{1.3}
\begin{table*}[!h]
\centering
\caption{State-wise performance of models on \SD test set in the different
settings: novel objects with 12.5\% training data, novel objects with 100\%
training data, all objects with 12.5\% training data, and all objects with
100\% training data.}
\tablelabel{epicstates-classwise-results}

\resizebox{\linewidth}{!}{
\begin{tabular}{lccccccccccc}
\toprule
\bf Novel Objects [12.5\%] & \textbf{\ttt{open}} & \textbf{\ttt{close}} & \textbf{\ttt{inhand}} & \textbf{\ttt{outofhand}} & \textbf{\ttt{raw}} & \textbf{\ttt{cooked}} & \textbf{\ttt{whole}} & \textbf{\ttt{cut}} & \textbf{\ttt{peeled}} & \textbf{\ttt{unpeeled}} & \textbf{Mean}\\
\midrule

ImageNet Pre-trained & 71.9 \std 0.0 & 53.5 \std 0.0 & 76.5 \std 0.0 & 62.8 \std 0.0 & 97.6 \std 0.0 & 87.1 \std 0.0 & 46.6 \std 0.0 & 82.1 \std 0.0 & 78.1 \std 0.0 & 45.9 \std 0.0 & 70.2 \std 0.0 \\

TCN~\cite{sermanet2018time} & 67.4 \std 4.6 & 50.9 \std 5.2 & 63.9 \std 2.5 & 46.7 \std 2.4 & 75.8 \std 1.3 & 32.9 \std 1.6 & 36.6 \std 2.6 & 76.6 \std 0.8 & 72.7 \std 1.7 & 37.3 \std 0.9 & 56.1 \std 1.9 \\
SimCLR~\cite{chen2020simple} & 77.4 \std 2.8 & 61.7 \std 2.6 & 72.0 \std 1.1 & 53.6 \std 1.4 & 96.0 \std 1.0 & 84.6 \std 2.0 & 55.2 \std 5.2 & 82.3 \std 3.3 & 81.7 \std 1.0 & 54.2 \std 3.0 & 71.9 \std 0.2 \\
SimCLR+TCN & 73.7 \std 2.3 & 57.2 \std 2.8 & 68.9 \std 1.1 & 51.8 \std 2.5 & 90.5 \std 0.7 & 65.3 \std 1.3 & 42.2 \std 2.3 & 76.0 \std 1.9 & 72.9 \std 3.7 & 38.5 \std 3.7 & 63.7 \std 0.3 \\

Semantic supervision\\
$\;$ via EPIC action classification & 76.0 \std 2.1 & 57.7 \std 1.5 & 76.4 \std 0.4 & 64.0 \std 1.4 & 95.3 \std 2.5 & 73.6 \std 9.0 & 56.3 \std 4.7 & 81.6 \std 2.0 & 79.4 \std 3.3 & 48.6 \std 4.2 & 70.9 \std 2.0 \\
$\;$ via MIT States dataset~\cite{isola2015discovering} & 75.0 \std 0.9 & 56.2 \std 4.3 & 73.7 \std 0.8 & 60.4 \std 2.6 & 94.5 \std 2.7 & 77.5 \std 9.1 & 51.4 \std 1.3 & 87.1 \std 1.6 & 78.1 \std 2.8 & 46.8 \std 1.5 & 70.1 \std 1.5 \\

\midrule
Ours [TSC] & 79.5 \std 1.1 & \textbf{63.5 \std 3.0} & 74.8 \std 1.9 & 54.5 \std 2.5 & 96.8 \std 1.3 & 81.7 \std 7.7 & 64.2 \std 3.3 & 87.0 \std 1.9 & 83.9 \std 1.6 & 58.8 \std 3.2 & 74.5 \std 1.0 \\
Ours [TSC+OHC] & \textbf{81.1 \std 0.5} & 62.5 \std 1.1 & \textbf{82.9 \std 0.9} & \textbf{67.5 \std 1.6} & \textbf{98.3 \std 0.6} & \textbf{88.1 \std 4.6} & \textbf{67.1 \std 2.6} & \textbf{90.4 \std 1.4} & \textbf{89.1 \std 1.8} & \textbf{70.4 \std 2.1} & \textbf{79.8 \std 0.6} \\

\bottomrule
\end{tabular}}

\vspace{10pt}
\resizebox{\linewidth}{!}{
\begin{tabular}{lccccccccccc}
\toprule
\bf Novel Objects [100\%] & \textbf{\ttt{open}} & \textbf{\ttt{close}} & \textbf{\ttt{inhand}} & \textbf{\ttt{outofhand}} & \textbf{\ttt{raw}} & \textbf{\ttt{cooked}} & \textbf{\ttt{whole}} & \textbf{\ttt{cut}} & \textbf{\ttt{peeled}} & \textbf{\ttt{unpeeled}} & \textbf{Mean}\\
\midrule

ImageNet Pre-trained & 74.8 \std 0.0 & 57.6 \std 0.0 & 78.5 \std 0.0 & 70.3 \std 0.0 & 98.6 \std 0.0 & 92.0 \std 0.0 & 62.1 \std 0.0 & 91.4 \std 0.0 & 74.5 \std 0.0 & 45.4 \std 0.0 & 74.5 \std 0.0 \\

TCN~\cite{sermanet2018time} & 70.2 \std 3.9 & 52.6 \std 3.3 & 71.0 \std 1.6 & 53.5 \std 1.8 & 87.0 \std 3.9 & 49.9 \std 10.6 & 50.6 \std 3.5 & 84.6 \std 1.9 & 73.6 \std 3.5 & 45.6 \std 4.6 & 63.9 \std 1.1 \\
SimCLR~\cite{chen2020simple} & 77.9 \std 2.2 & 62.2 \std 1.9 & 77.4 \std 1.3 & 63.1 \std 1.1 & 97.4 \std 1.3 & 94.0 \std 3.4 & 65.4 \std 4.3 & 90.3 \std 0.5 & 82.0 \std 0.5 & 61.6 \std 3.9 & 77.1 \std 1.0 \\
SimCLR+TCN & 76.1 \std 2.8 & 61.0 \std 2.9 & 75.2 \std 0.7 & 60.3 \std 2.0 & 92.7 \std 0.7 & 77.6 \std 5.9 & 52.5 \std 4.0 & 82.9 \std 2.8 & 69.6 \std 3.0 & 35.7 \std 5.2 & 68.4 \std 1.6 \\

Semantic supervision\\
$\;$ via EPIC action classification & 80.0 \std 2.4 & 59.8 \std 3.1 & 82.4 \std 1.4 & 73.6 \std 1.0 & 97.1 \std 0.3 & 83.8 \std 8.2 & 59.6 \std 4.5 & 87.7 \std 2.1 & 84.6 \std 2.4 & 62.0 \std 7.9 & 77.0 \std 0.9 \\
$\;$ via MIT States dataset~\cite{isola2015discovering} & 77.2 \std 1.7 & 58.3 \std 4.4 & 78.8 \std 2.1 & 67.9 \std 0.9 & 95.6 \std 2.4 & 85.6 \std 6.8 & 48.6 \std 2.2 & 89.2 \std 0.9 & 82.8 \std 4.5 & 54.9 \std 10.9 & 73.9 \std 0.7 \\

\midrule
Ours [TSC] & 80.2 \std 1.2 & \textbf{63.8 \std 3.7} & 80.5 \std 1.6 & 63.5 \std 2.5 & 98.8 \std 0.4 & 94.1 \std 1.7 & \textbf{73.1 \std 1.6} & \textbf{93.1 \std 0.7} & \textbf{87.9 \std 0.2} & \textbf{67.2 \std 2.4} & 80.2 \std 0.4 \\
Ours [TSC+OHC] & \textbf{81.2 \std 0.4} & 63.2 \std 1.7 & \textbf{87.9 \std 0.7} & \textbf{77.3 \std 1.6} & \textbf{99.2 \std 0.4} & \textbf{94.7 \std 3.0} & 69.7 \std 3.3 & 92.9 \std 0.8 & 87.7 \std 1.6 & 64.5 \std 5.0 & \textbf{81.8 \std 0.4} \\

\bottomrule
\end{tabular}}

\vspace{10pt}
\resizebox{\linewidth}{!}{
\begin{tabular}{lccccccccccc}
\toprule
\bf All Objects [12.5\%] & \textbf{\ttt{open}} & \textbf{\ttt{close}} & \textbf{\ttt{inhand}} & \textbf{\ttt{outofhand}} & \textbf{\ttt{raw}} & \textbf{\ttt{cooked}} & \textbf{\ttt{whole}} & \textbf{\ttt{cut}} & \textbf{\ttt{peeled}} & \textbf{\ttt{unpeeled}} & \textbf{Mean}\\
\midrule

ImageNet Pre-trained & 90.1 \std 0.0 & 62.3 \std 0.0 & 83.1 \std 0.0 & 73.4 \std 0.0 & 97.4 \std 0.0 & 83.8 \std 0.0 & 68.7 \std 0.0 & 85.1 \std 0.0 & 87.2 \std 0.0 & 51.2 \std 0.0 & 78.2 \std 0.0 \\

TCN~\cite{sermanet2018time} & 83.0 \std 1.7 & 48.6 \std 2.2 & 73.9 \std 2.1 & 55.9 \std 2.2 & 85.0 \std 0.8 & 40.4 \std 1.4 & 53.0 \std 2.2 & 72.7 \std 1.3 & 78.6 \std 0.3 & 33.6 \std 1.7 & 62.5 \std 0.8 \\
SimCLR~\cite{chen2020simple} & 91.0 \std 0.8 & 65.3 \std 1.5 & 79.5 \std 0.9 & 64.6 \std 2.2 & 97.1 \std 0.6 & 82.0 \std 3.7 & 70.0 \std 1.8 & 82.1 \std 1.6 & 85.8 \std 2.3 & 57.2 \std 0.5 & 77.4 \std 1.0 \\
SimCLR+TCN & 87.9 \std 1.2 & 59.7 \std 1.6 & 76.8 \std 0.4 & 60.7 \std 2.2 & 95.4 \std 0.7 & 76.1 \std 4.9 & 63.3 \std 1.8 & 80.3 \std 0.6 & 82.3 \std 3.5 & 46.3 \std 3.5 & 72.9 \std 1.3 \\

Semantic supervision\\
$\;$ via EPIC action classification & 88.5 \std 1.1 & 55.0 \std 1.6 & 83.1 \std 0.9 & 72.2 \std 0.2 & 95.9 \std 0.7 & 68.8 \std 3.7 & 60.0 \std 1.8 & 73.4 \std 2.2 & 79.7 \std 3.3 & 44.5 \std 1.5 & 72.1 \std 0.8 \\
$\;$ via MIT States dataset~\cite{isola2015discovering} & 89.9 \std 0.1 & 58.5 \std 1.6 & 81.1 \std 1.2 & 70.0 \std 3.1 & 96.3 \std 1.2 & 78.9 \std 3.3 & 66.5 \std 2.0 & 86.3 \std 1.3 & 86.6 \std 1.3 & 50.4 \std 3.9 & 76.4 \std 0.6 \\

\midrule
Ours [TSC] & \textbf{91.9 \std 0.5} & \textbf{65.6 \std 1.0} & 83.6 \std 0.6 & 69.8 \std 2.3 & \textbf{98.2 \std 0.0} & 86.7 \std 2.6 & \textbf{74.3 \std 2.7} & 88.3 \std 1.5 & \textbf{90.3 \std 0.1} & \textbf{65.2 \std 0.6} & 81.4 \std 1.0 \\
Ours [TSC+OHC] & \textbf{91.9 \std 0.6} & 63.5 \std 1.6 & \textbf{88.7 \std 0.4} & \textbf{77.9 \std 0.2} & \textbf{98.2 \std 0.1} & \textbf{88.5 \std 1.1} & 73.5 \std 0.9 & \textbf{89.5 \std 0.7} & 89.8 \std 0.6 & 64.8 \std 2.2 & \textbf{82.6 \std 0.2} \\

\bottomrule
\end{tabular}}

\vspace{10pt}
\resizebox{\linewidth}{!}{
\begin{tabular}{lccccccccccc}
\toprule
\bf All Objects [100\%] & \textbf{\ttt{open}} & \textbf{\ttt{close}} & \textbf{\ttt{inhand}} & \textbf{\ttt{outofhand}} & \textbf{\ttt{raw}} & \textbf{\ttt{cooked}} & \textbf{\ttt{whole}} & \textbf{\ttt{cut}} & \textbf{\ttt{peeled}} & \textbf{\ttt{unpeeled}} & \textbf{Mean}\\
\midrule

ImageNet Pre-trained & 92.9 \std 0.0 & 68.3 \std 0.0 & 85.3 \std 0.0 & 78.2 \std 0.0 & 98.3 \std 0.0 & 88.5 \std 0.0 & 73.0 \std 0.0 & 89.4 \std 0.0 & 91.4 \std 0.0 & 65.1 \std 0.0 & 83.1 \std 0.0 \\

TCN~\cite{sermanet2018time} & 87.0 \std 1.0 & 56.7 \std 0.6 & 80.6 \std 1.0 & 67.2 \std 0.8 & 94.2 \std 0.9 & 68.5 \std 2.3 & 65.8 \std 2.3 & 82.7 \std 1.5 & 85.1 \std 1.5 & 46.4 \std 4.7 & 73.4 \std 1.4 \\
SimCLR~\cite{chen2020simple} & 92.1 \std 0.8 & 67.6 \std 2.3 & 82.9 \std 0.5 & 71.3 \std 1.2 & 97.8 \std 0.6 & 86.0 \std 3.1 & 72.8 \std 1.6 & 86.9 \std 0.8 & 88.9 \std 2.2 & 64.1 \std 3.0 & 81.0 \std 0.9 \\
SimCLR+TCN & 90.5 \std 1.2 & 65.5 \std 1.3 & 81.3 \std 0.3 & 69.1 \std 1.5 & 96.9 \std 0.9 & 78.3 \std 6.3 & 68.9 \std 3.1 & 84.2 \std 1.8 & 84.9 \std 2.6 & 54.2 \std 2.5 & 77.4 \std 1.2 \\

Semantic supervision\\
$\;$ via EPIC action classification & 91.1 \std 0.7 & 63.2 \std 1.2 & 86.6 \std 0.7 & 79.2 \std 1.3 & 97.0 \std 0.3 & 73.4 \std 4.4 & 65.2 \std 1.4 & 80.9 \std 1.3 & 87.1 \std 1.7 & 56.0 \std 9.9 & 77.9 \std 1.3 \\
$\;$ via MIT States dataset~\cite{isola2015discovering} & 92.0 \std 0.1 & 66.9 \std 1.7 & 83.3 \std 0.5 & 73.9 \std 1.3 & 96.7 \std 1.8 & 82.5 \std 4.6 & 75.6 \std 2.3 & 89.3 \std 1.3 & 90.3 \std 1.7 & 64.4 \std 7.8 & 81.5 \std 1.3 \\

\midrule
Ours [TSC] & \textbf{93.0 \std 0.3} & \textbf{69.6 \std 2.1} & 86.9 \std 0.6 & 75.8 \std 2.1 & \textbf{98.6 \std 0.1} & 89.0 \std 2.5 & \textbf{77.5 \std 2.6} & 89.8 \std 1.5 & 92.1 \std 0.6 & \textbf{69.3 \std 4.2} & 84.2 \std 1.0 \\
Ours [TSC+OHC] & 92.4 \std 0.8 & 66.1 \std 2.5 & \textbf{90.5 \std 0.5} & \textbf{83.1 \std 1.4} & 98.5 \std 0.1 & \textbf{91.1 \std 0.8} & 76.3 \std 2.9 & \textbf{91.2 \std 1.1} & \textbf{92.5 \std 1.1} & 66.7 \std 4.1 & \textbf{84.8 \std 0.4} \\

\bottomrule
\end{tabular}}

\end{table*}
 
\subsubsection{TSC+OHC Ablations}
\setlength{\intextsep}{8pt}\setlength{\columnsep}{10pt}\renewcommand{\arraystretch}{1.1}
\begin{wraptable}[9]{r}{10.5cm} \setlength{\tabcolsep}{8pt}
\centering
\caption{We perform ablations on the components of TSC+OHC on the validation
set. Both hand motion and appearance contribute to the performance over TSC,
with motion being more important.}
\tablelabel{hand-ablations}
\resizebox{\linewidth}{!}
{
\begin{tabular}{lcccc}
\toprule
& \multicolumn{2}{c}{\textbf{Novel Objects}} & \multicolumn{2}{c}{\textbf{All Objects}}\\
\cmidrule(lr){2-3} \cmidrule(lr){4-5}
Linear classifier training data & 12.5\% & 100\% & 12.5\% & 100\% \\ 
\midrule
TSC                           & 72.3 \std 1.3 & 77.8 \std 0.4 & 78.3 \std 0.3 & 81.2 \std 0.3 \\
TSC+OHC (appearance)          & 73.8 \std 0.8 & 77.3 \std 0.5 & 78.4 \std 0.5 & \textbf{82.5 \std 0.6} \\
TSC+OHC (motion)              & 74.6 \std 1.6 & \textbf{78.7 \std 0.6} & \textbf{78.8 \std 0.4} & 82.4 \std 0.2 \\
TSC+OHC (motion + appearance) & \textbf{75.1 \std 0.4} & 78.2 \std 0.2 & 78.6 \std 0.2 & 81.7 \std 0.3 \\
\bottomrule
\end{tabular}}
\end{wraptable}
 We analyse the individual contribution of hand motion ($h^m_{i}$) and hand
appearance ($h^a_i$) towards the performance of TSC+OHC.
\tableref{hand-ablations} shows that both components, by themselves, improve
upon just TSC.  Motion information gives larger boosts than appearance
information; and both together lead to the best performance in the challenging
setting of novel categories with limited data.

\subsubsection{Track Ablations}
Mining object-level tracks from in-the-wild videos presents two challenges: a)
how to select a {\it useful} patch to track, and b) how to successfully track
it in the given ego-centric video.

Egocentric videos showcase objects that are undergoing {\it non-trivial}
transformations (deformations, state changes, occlusion by hands). Furthermore,
use of hand context could aid with tracking in egocentric videos that have
large amounts of egomotion.  We test the extent to which these advantages of
working with egocentric videos contributes to performance. We generate several
sets of tracks that ablate the two aforementioned factors. Visualizations for
these tracks are shown by \figref{tracks}, and the quantitative results are
presented in \tableref{state-ablations}. \\

\noindent \textbf{Source for Starting Patches.} We experiment with the
following sources for the starting patch.
\begin{enumerate}
\item \textbf{Object-agnostic Starting Patch.} Here, we consider an
arbitrary starting patch source, either a random crop or center crop in a
frame. Random crops vary in scale and location, while the center crop is always
a 256 $\times$ 256 crop from the $456 \times 256$ image.
\item \textbf{Starting Patch on Background Object.} We detect background
objects (\ie not overlapping with objects of interaction as detected by the
model from \cite{shan2020understanding}) using Mask~RCNN~\cite{he2017mask} with
a ResNet101-FPN backbone trained on MS-COCO 2017 instance segmentation dataset.
We only detections for categories commonly found in kitchens and remove classes
like car, train, \etc We only consider the 10 highest scoring detections, and
sample a detection that doesn't overlap with the object-of-interaction as the
starting patch.
\item \textbf{Starting Patch on Object-of-Interaction (Ours).} We use the
object-of-interaction detections from Shan \etal~\cite{shan2020understanding}.
As noted in the main paper, we use leave one out predictions
from~\cite{shan2020understanding}: we split the train set into 5 parts by
participants, retrain~\cite{shan2020understanding} on 4, use predictions on the
5\textsuperscript{th} (\ie unseen participants); and repeat this 5 times over.
\item \textbf{Ground Truth Objects-of-Interaction (Ceiling).} For reference, we
also report performance on using ground truth objects of interaction as
annotated in the \epic dataset. We use these with ground truth
tracking (see below).
\end{enumerate}

\noindent \textbf{Tracking Algorithm.} We experiment with the following
tracking algorithms.

\begin{enumerate}
\item \textbf{No Tracking.} Here, we don't do any tracking and copy over the
box from the previous frame, to the same location in the current frame.
\item \textbf{Off-the-Shelf Tracker.} We use SiamRPN++ tracker
from~\cite{li2019siamrpn++} to track the object from one frame to the next.
Given a starting crop, the tracker produces bounding boxes for crop in
consecutive frames. In practice, we only consider a tracker-produced bounding
box to be valid if it has a score above 0.1. We allow for up to two frames of
either missing or invalid detections, or a max of 256 frames tracked, before
sub-sampling and saving the track.  
\item \textbf{Hand-context (Ours).} To construct our tracks, we
focus on objects-of-interaction
detected by~\cite{shan2020understanding}
along with information about what
hands do they correspond to. We do this jointly with
the object-of-interaction starting patches described above. 
In more detail, we utilize hand-object detections for both, finding the starting
patch and tracking it. Specifically, we start with a frame and find all
interacted objects with a score above 0.2 and start tracking them
independently. At the next frame, we receive another set of valid objects
bounding boxes and try to match them with the previous frame's detections by
posing the problem as a linear sum assignment in a bipartite graph where the
cost is the intersection over union (IoU) over the two bounding boxes (provided
that IoU $>$ 0.4). Boxes still not matched with previous boxes start their own
track. Tracks also have an 8 frame buffer with no matches before they are
closed. We subsample the tracks to 10fps. We cap the
track length at 25.6s, and split longer tracks. We get a total of 61.3K tracks. 

\item \textbf{Ground Truth (Ceiling).} Here we use the ground truth
object-of-interaction tracks provided in the \epic dataset (used in conjunction
with ground truth object-of-interaction above). We use the bounding box
annotations for the object-of-interaction from Damen
\etal~\cite{damen2020collection} on \epic dataset. Since these annotations are
provided at 0.5 fps, we interpolate the bounding box for the intermediate
frames to get dense tracks. This gives us 16,474 tracks with an average length
of 66 frames. 
\end{enumerate}

\renewcommand{\arraystretch}{1.1}
\begin{wraptable}[17]{r}{9cm} \setlength{\tabcolsep}{8pt}
\centering
\caption{Comparison on validation set for various tracking approaches used for
learning state sensitive features with Temporal SimCLR. Tracks obtained with the
hand-object-interaction detector perform the best and come close to 
hand annotated tracks~\cite{damen2020collection} in performance. Italicized rows
correspond to our proposal in this paper.}
\tablelabel{state-ablations}
\resizebox{1.0\linewidth}{!}{
\begin{tabular}{lccc}
\toprule
\textbf{Starting Patch Source} & \textbf{Tracking Algorithm} & \textbf{TSC Val mAP} \\
\midrule
Center crop                                                      & None                             & 72.9 \\
Random crop                                                      & None                             & 77.1 \\
Object-of-interaction~\cite{shan2020understanding}               & None                             & 79.2 \\
Random Background Crop                                           & SiamRPN++~\cite{li2019siamrpn++} & 77.6 \\
Random Background Object                                         & SiamRPN++~\cite{li2019siamrpn++} & 80.6 \\
Object-of-interaction~\cite{shan2020understanding}               & SiamRPN++~\cite{li2019siamrpn++} & 79.7 \\
\it Object-of-interaction~\cite{shan2020understanding}           & \it Hand context                 & \it 81.2 \\
GT Object-of-interaction                                         & Ground Truth                     & 83.5 \\
\midrule
\it Object-of-interaction~\cite{shan2020understanding}           & \it Hand context                 & \it 81.7 (TSC+OHC) \\
\bottomrule
\end{tabular}}
\end{wraptable}
 \noindent \textbf{Results.} 
\tableref{state-ablations} shows the performance of TSC on the various tracks.
As noted in the main paper, use of object-of-interaction tracks offers two
advantages: they stabilize for the large egomotion in egocentric videos, and
focus on aspects of the scene that are undergoing interesting (non-viewpoint)
transformations. No stabilization performs poorly. Stabilization using
off-the-shelf tracker SiamRPN++~\cite{li2019siamrpn++} also works well.
However, tracking with hand context enables use of Object-Hand Consistency
which aids performance. Ground truth tracks annotated in \epic dataset lead to
better learning, indicating that better detection and tracking of
objects-of-interaction can improve performance further.

\begin{figure*}[t]
\setlength{\tabcolsep}{2pt}
\centering
\resizebox{1\linewidth}{!}{
\begin{tabular}{rcc}
\rotatebox[origin=l]{90}{\parbox{2cm}{\small Random+ No~tracking}} &
\includegraphics[width=1\textwidth]{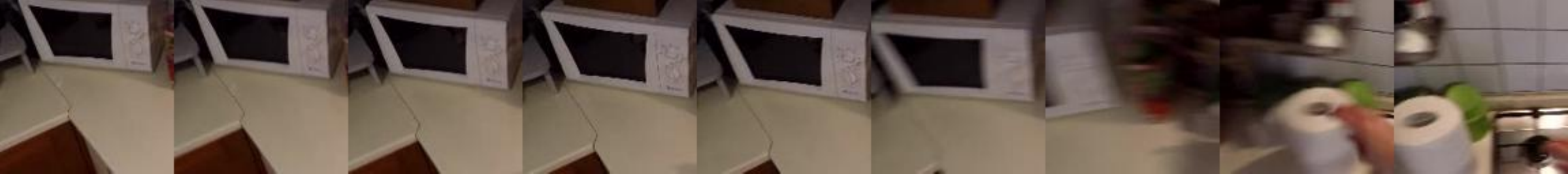} \\ 
\rotatebox[origin=l]{90}{\parbox{1.8cm}{\small MaskRCNN+ SiamRPN++}} &
\includegraphics[width=1\textwidth]{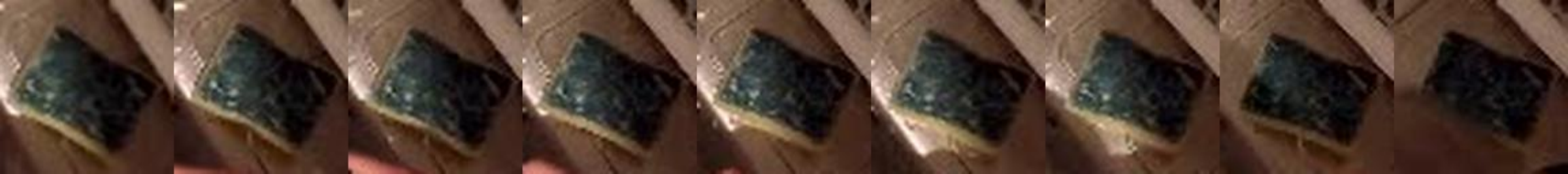} \\
\rotatebox[origin=l]{90}{\parbox{2.6cm}{\small InteractionObj+ HandContext~[Ours]}} &
\includegraphics[width=1\textwidth]{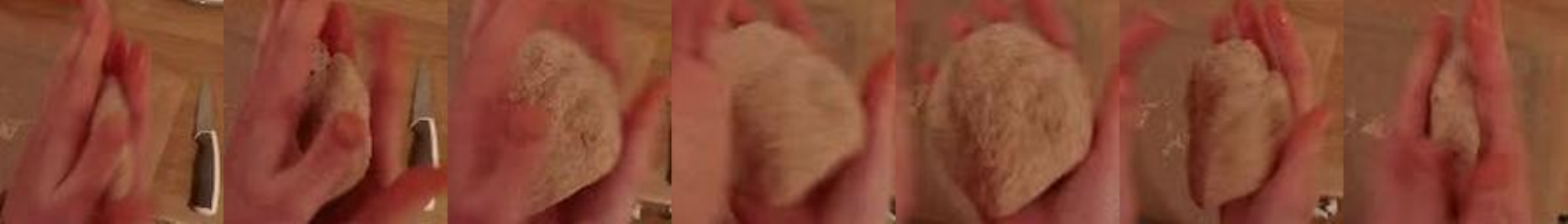} \\ 
\end{tabular}
}
\caption{Sample track from the random crop with no tracking (Random), the
background object crop with MaskRCNN + tracking with SiamRPN++ (MaskRCNNN +
SiamRPN++), and our tracks that use objects-of-interaction and track using hand
context. We see that ego-motion in egocentric videos leads to large drift by
the end without any tracking. We see that the background object tracks fail to
capture meaningful appearance changes. For our tracks, we see the object in a
variety of poses with distinct appearances.}
\figlabel{tracks}
\end{figure*}
  
\cleardoublepage

\section{Object Affordance Prediction}
\subsection{EPIC-ROI Dataset and Task}
\seclabel{epic-roi}
\noindent \textbf{Task Definition.} The ROI (Region of Interaction) task is to
predict regions where human hands {\it frequently} touch in everyday
interaction. Specifically, image regions that afford any of the most frequent
actions: \ttt{take, open, close, press, dry, turn, peel} are considered as
positive.

\noindent \textbf{Data Sampling.} We randomly sampled 500 images from 9
participants: P01, P08, P11, P02, P32, P18, P04, P09, P03. From these
participants, we only use videos present in the test set of \epic dataset (2018
version). We annotated frames at 1920 $\times$ 1080 resolution.  The images may
or may not contain participant hands (if they were present, they were annotated
and ignored during evaluation) .  We manually filter out images to minimize
motion blur, out of distribution frames (for example, completely dark frames at
the starting of some videos or rare views such as picking a spoon that fell
onto the floor). We made sure to minimize redundancy among frames by
selecting the most diverse 7 -- 15 frames from each participant.

\noindent \textbf{Annotation Procedure.} To determine where participants
frequently interact in the scene, we manually watched the videos from these
participants and created a list of objects that underwent interaction objects
and also identified the interacted regions. Then, for every considered action, we annotated applicable regions of
interaction using polygons for larger objects (such as bottles, jars \etc), and
lines for thin regions (wires, rims and object edges). The lines for the rims
and edges of objects were converted to regions by dilating them by 25 pixels to
convert them to strips. Annotation for 10 images from 1 participant took 120
minutes on average. Lastly, we aggregate the annotations across all actions to generate the EPIC-ROI ground truth segmentation mask.

\begin{figure}[t]
\centering
  \insertWL{0.33}{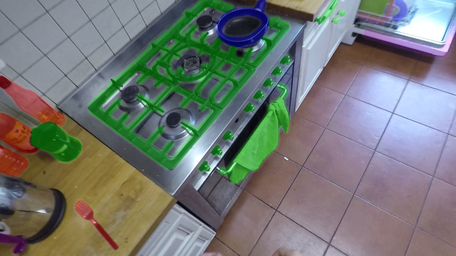}
  \insertWL{0.33}{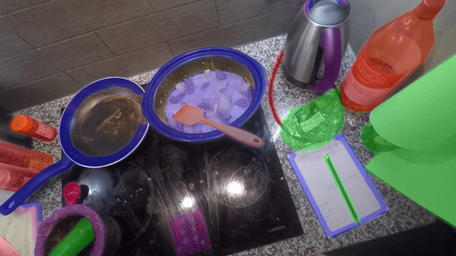}
  \insertWL{0.33}{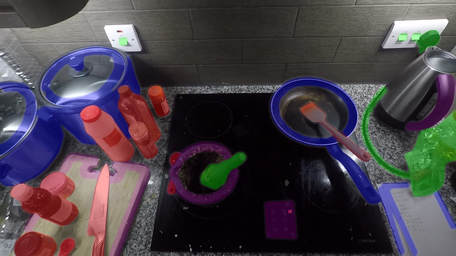}
  \\
  \insertWL{0.33}{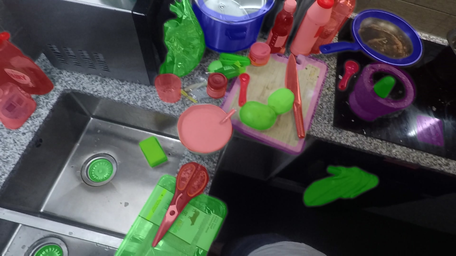}
  \insertWL{0.33}{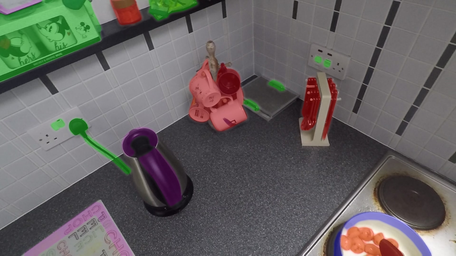}
  \insertWL{0.33}{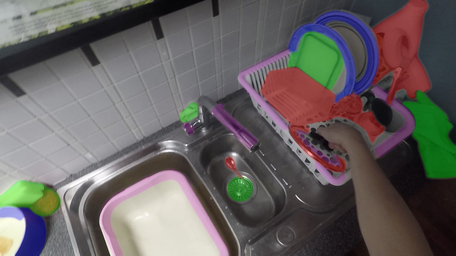}
  \\
  \insertWL{0.33}{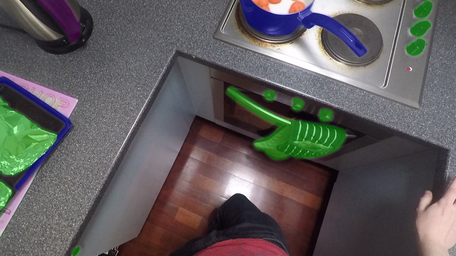}
  \insertWL{0.33}{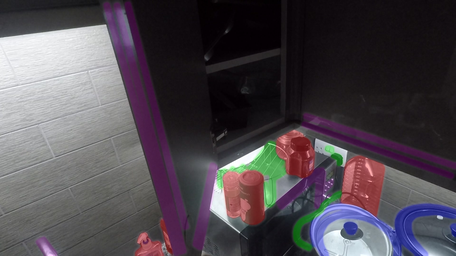}
  \insertWL{0.33}{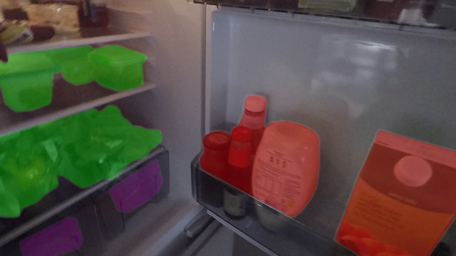}
  \\
  \insertWL{0.33}{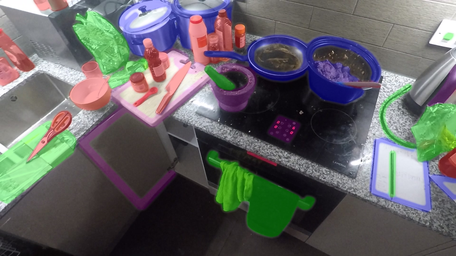}
  \insertWL{0.33}{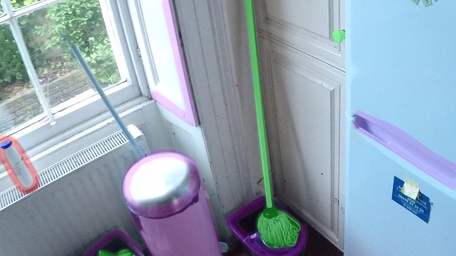}
  \insertWL{0.33}{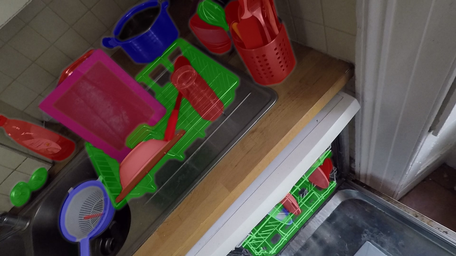}
  \\
\caption{{\bf Annotated images from \RoID dataset.} We show some sample images
from the dataset annotated for evaluating region of interaction predictions.
Each annotated region is attributed with one of the four labels: COCO objects
(red), Non-COCO objects (green), COCO parts (blue), and Non-COCO parts
(magenta).}
\figlabel{roi-samples}
\end{figure}
 
To enable detailed analysis, every annotation is also assigned one of the four
labels: COCO objects, Non-COCO objects, COCO parts, and Non-COCO parts.  To
determine the set of COCO objects, we first select only the relevant classes
removing categories like cat, dog, bird etc. This leaves us with the following
categories: \ttt{backpack, umbrella, handbag, tie, suitcase, sports ball,
baseball bat, baseball glove, tennis racket, bottle, wine glass, cup, fork,
knife, spoon, bowl, banana, apple, sandwich, orange, broccoli, carrot, hot dog,
pizza, donut, cake, chair, mouse, remote, keyboard, cell phone, toaster, book,
vase, scissors, hair drier, toothbrush, microwave, oven, sink,} and
\ttt{refrigerator}. We further observed that removing \ttt{microwave, oven,
sink,} and \ttt{refrigerator} from the relevant categories improves the
performance of Mask~RCNN on the validation split (see
\tableref{maskrcnn-ablations}). Thus, we don't include objects from these 4
categories into COCO objects. \figref{roi-samples} shows some annotated images
from our validation split where different categories (out of above four) are
assigned different colors.

\renewcommand{\arraystretch}{1.0}
\begin{wraptable}[4]{r}{7cm} \setlength{\tabcolsep}{8pt}
\centering
\caption{Val and test sets for \RoID dataset.}
\tablelabel{acp-pids}
\resizebox{1.0\linewidth}{!}{
\begin{tabular}{llc}
\toprule
\textbf{Split} & \textbf{Participants} & \textbf{\# Frames} \\
\midrule
Validation & P03, P04, P09                & 32 \\
Test       & P01, P02, P08, P11, P18, P32 & 71\\
\bottomrule
\end{tabular}}
\end{wraptable}
 \noindent \textbf{Dataset Splits.} We split the collected dataset into
validation and testing sets based on the participants.
P03, P04, and P09 are in the validation set with a total of 32 frames.
P01, P02, P08, P11, P18, and P32 are in the test set with a total of 71 images.

 \subsection{Grasps Afforded by Objects (GAO) Task}
\seclabel{GAO}

\noindent \textbf{Task Definition.} The task is to predict the hand-grasps
afforded by objects present in the scene where each object can afford multiple
grasps. The task also requires reasoning about the occlusion between objects
which can leave some of the hand-grasps inapplicable.

\noindent
\textbf{Datasets.} We utilize \ycb dataset~\cite{corona2020ganhand}
that builds upon YCB-Videos~\cite{xiang2018posecnn} (sample frames shown in
\figref{ycb-samples}) to set up GAO Task. The dataset annotates each object
with the afforded hand grasps (see \figref{ycb-samples} (right)). 

For our methods and the baseline alike, we assume that objects have already
been localized (we use the object masks provided with the dataset). This
side-steps the detection problem and allows us to focus on the task of
predicting afforded grasps. For the baselines, predictions are made on a crop
around the object of interest. For our method, dense predictions from our model 
are aggregated over the segmentation mask to obtain the final classification
(more below).

\noindent
\textbf{Splits.} We divide the \ycb dataset into three parts
for training, validation and testing. We make sure that the training,
  validation and test sets do not overlap in the objects. Further,
  there is no overlap in the videos from training, validation and test sets.
  This results in a training set consisting of 77 videos, validation set with 6
  videos and testing set with 9 videos. We further only use 15 objects (out of
  21) from the training set to train the supervised baseline. For validation
  and testing, we use the remaining 6 objects to compute the metrics. Since the
  scene is static, and the camera motion is slow, we sub-sample 60 frames (10
  from each video) from the validation videos to create the validation set, and
  180 frames (20 from each video) to create the testing set. We use all the
  frames (110K) from training videos to create the training set for supervised
  ceiling. The splits ensure that we test generalization to novel objects.

\begin{figure}[!h]
\centering
\insertH{0.213}{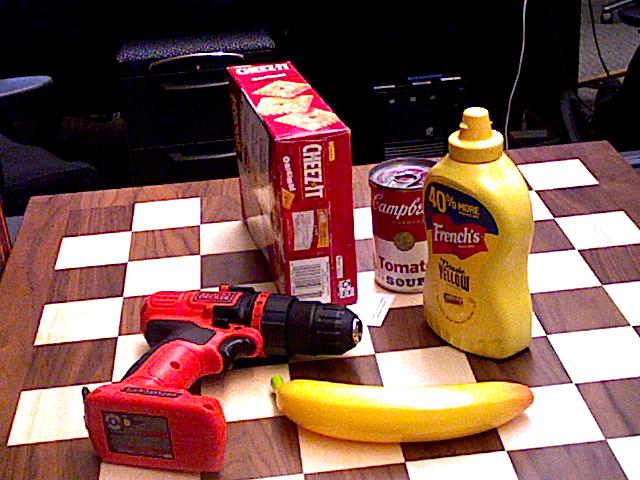} \hfill
  \insertH{0.213}{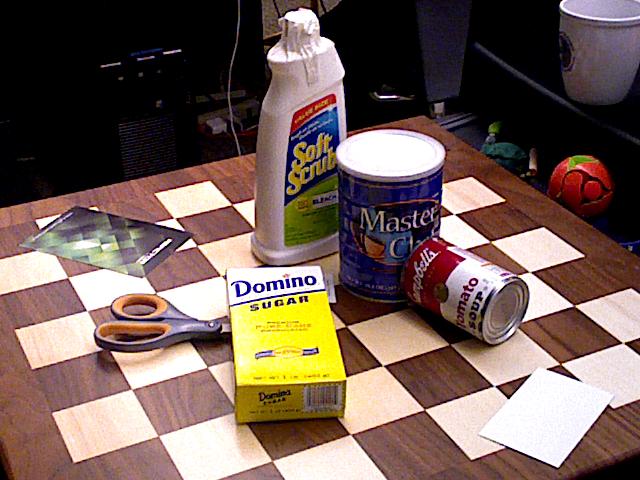} \hfill
  \insertH{0.213}{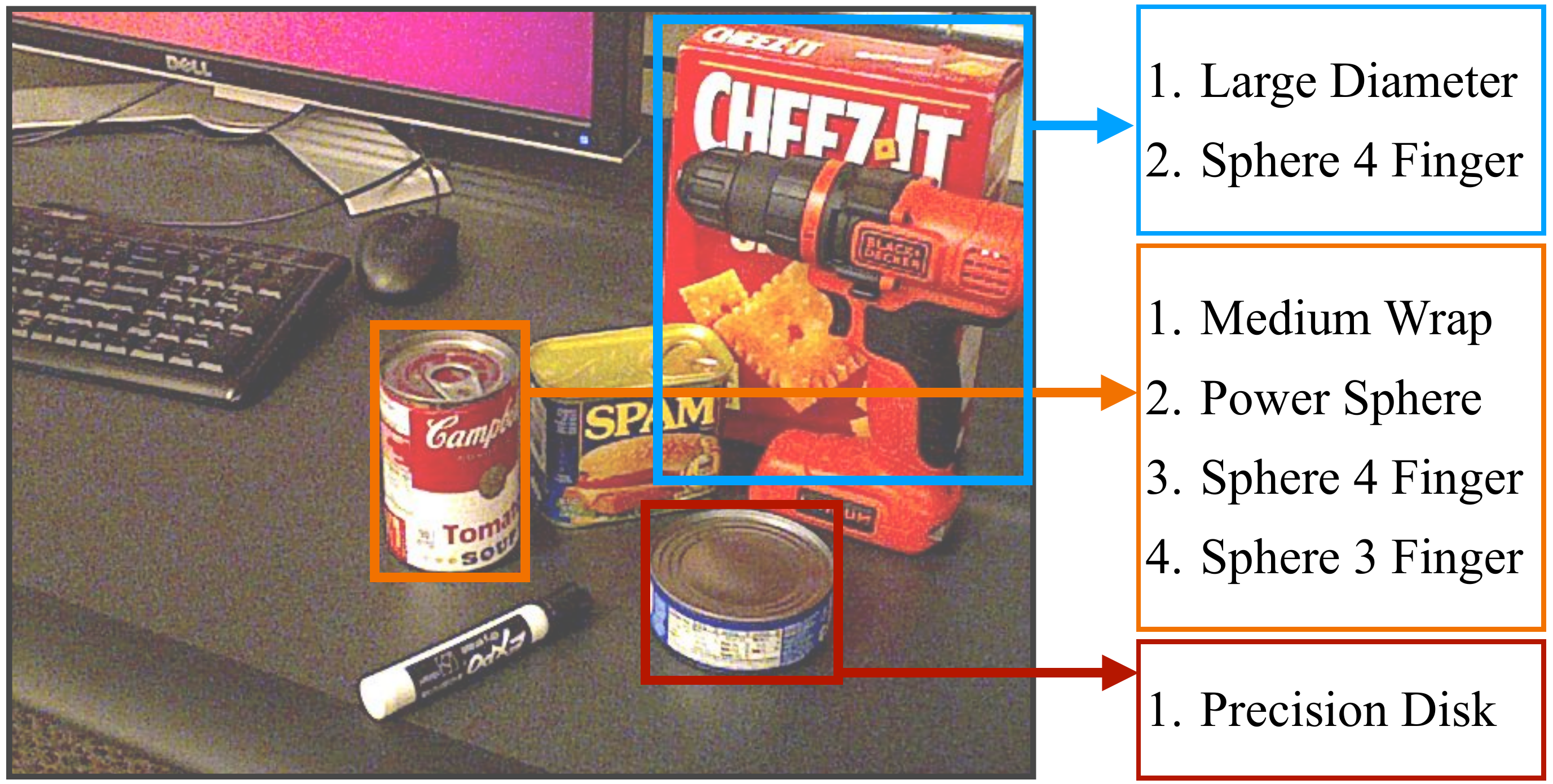}
\caption{{\bf Sample frames from YCB-Videos~\cite{xiang2018posecnn}.} We use
annotations from Corona \etal~\cite{corona2020ganhand} on the
YCB-Videos~\cite{xiang2018posecnn} to setup the GAO task. The task is to
predict the hand-grasps afforded by the objects present in the scene (see right
figure). This requires reasoning about object shape and occlusion patterns that
can render some grasps inapplicable.} \figlabel{ycb-samples}
\end{figure}
  \subsection{Model Details}
\subsubsection{Affordances via Context Prediction (ACP) Details}
\seclabel{acp-approach}

\noindent
\textbf{Architecture.} We use the \RS{50} backbone as the encoder. The region
of interaction branch uses a decoder that consists of 4 deconvolution layers
with $4 \times 4$ kernels and stride length of 2 and a padding of 1. Lastly, we have a $5 \times 5$ average pooling layer with padding 2 and stride 1 which outputs the final ROI-prediction.
The grasp prediction branch, uses one fully-connected layer followed
by 33 binary classifiers on top of the output from the encoder.

\noindent
\textbf{Training Splits.} We use participants P05, P06, P07, P10, P12, P13, P14,
P15, P16, P17, P19, P20, P21, P22, P23, P24, P25, P26, P27, P28, P29, P30, P31
for training the model. Note that these are disjoint from the participants used
for validation and testing in \RoID, as listed in \tableref{acp-pids}.

\noindent \textbf{Data Sampling.} For training our ACP model, we extract
patches from $456 \times 256$ \epic frames (2018 version). We use the hand and
object-of-interaction detections to generate the ground-truth segmentation mask
(by pasting the detections). We only use object-of-interaction detections that have a confidence
score $\geq 0.8$, and that are smaller than 150 pixels in width and height.
Next, we sample positives around the detected hands and detected
objects-of-interaction. For hands, we randomly select a square patch 1 to 1.3
times the size of the detected hand box, centered at the hand (positive) or
randomly located elsewhere in the image (negative). For sampling around the
objects, we only consider object-of-interaction detections that have width and height greater than 20
pixels, and we sample a square patch inside the object-of-interaction box
(positive). The width of the sampled patch is randomly varied between 0.5 to
0.75 times the size of the detected object-of-interaction box. We train on all
participants except the ones in the \RoID validation and testing sets.
For training the grasp prediction branch, we only use the positive patches
sample centered at the hand.

Patches are resized to $128\times 128$ for training. Note that the bottom
center $64\time 64$ region is masked out before feeding into the networks both
at train and test times.

\noindent
\textbf{Loss function.} Our loss contains two terms $\Ls$ and $\Lg$,
the former training the region of interaction branch and the latter training
the grasp prediction branch. Both losses encourage the network to focus on the
surrounding context to make prediction about the hidden hand.

$\Ls$ computes the binary cross-entropy between
the predicted segmentation mask and the ground truth segmentation mask (as
derived by pasting detection boxes). We weight the positive pixels by a factor
of 4. 

$\Lg$ is trained on predictions from a classifier trained on the GUN-71
dataset~\cite{rogez2015understanding}. We only consider scores for the 33 hand
grasps (that are annotated in \ycb dataset) from the GUN-71 classifier. 
We use the highest scoring class as the positive class. We create a set of
negatives which consists of the least scoring $K=15$ classes. This
generates both the positive and negative data for training the grasp prediction
head.

$\Ls$ is trained on all positive and negative patches described above. $\Lg$ is
only trained on positive patches (\ie those that are around the hands).

We train both the segmentation head and the grasp prediction branch jointly by
combing the two loss functions as, 
\begin{equation}
    L = L_{seg} + 0.5 \cdot L_{grasp}.
\end{equation}

\noindent
\textbf{Training Details.} We use a batch size of 64 and Adam optimizer with a
learning rate of $10^{-4}$. During training, we also perform horizontal flips,
motion blur and color jitter augmentations on the input image. The masked
context region is resized to $128 \times 128$ before being input to the model.
We train for a total of 400 epochs (each epoch consisting of 256 iterations on
randomly sampled batches with batch size of 64) and then validate checkpoints
at epoch 300, 350, and 400 to select the best model for evaluation. Training
took 6 hours on a single modern GPU (RTX 2080 Ti or equivalent).

\noindent
\textbf{Supervision for Grasp Prediction Branch.} Here, we provide more details
about the GUN-71 classifier that is used to generate the necessary supervision
for training ACP. 

This GUN-71 classifier is trained on hands cropped from the GUN-71 dataset from
Rogez \etal~\cite{rogez2015understanding}. We
use a hand detector~\cite{shan2020understanding} to crop out hands from the
GUN-71 dataset resulting in 8403 crops for training (Subjects 1, 2, 3, 4, 5,
and 6) and 1655 crops for validation (Subject 7). The classifier uses a \RS{18}
backbone with two fully connected layers (512 and 128 units), followed by a
71-way classification layer. This model is trained using hand grasp annotations
in the GUN-71 dataset with a cross-entropy loss.

In addition to this grasp classification layer, we also have another head
consisting of one linear layer (128-dimensional output) that is trained using
$\Lt$ on the \epic dataset to {\it adapt} the GUN-71 classifier to work well on
\epic dataset. This $\Lt$ uses the hand tracks obtained using detector from 
\cite{shan2020understanding} on the \epic dataset, as used by the other
parts of our paper. 
  
This network is trained jointly by minimizing $L_c + \Lt$ using Adam optimizer
with a learning rate of $10^{-4}$ and 0.05 weight decay. We use a batch size
128.  We perform random crops, horizontal flips, and color jitter as
augmentations.  We train for a maximum of 60 epochs where we early stop based
on the validation performance. For $\Lt$, we use a window of length 10 to sample
positive hand crops. We only train on tracks with a minimum length of 15 frames. We save the model after every 3 epochs, and select the snapshot based on the validation performance on the GAO task. Typically, training for 24-30 epochs resulted in the best performance where each epoch consisted of training for 64 iterations.

\noindent \textbf{ACP (no $\Lt$).} This model is trained similarly to ACP but
only for minimizing $L_c$ cross entropy loss for classification.

\vspace{10pt}
\noindent \textbf{Inference.} At test time, we uniformly sample square context
regions of size 60, 100, and 160 from $1920 \times 1080$ images. We sample 4000
regions for each size, resulting in a total of 12000 regions. We resize the
sampled patches to $128 \times 128$ and mask out their bottom center $64\times 64$ region,
before feeding them into our learned models to obtain the 12000 $64\times 64$
predictions.

  These predictions are spatially aggregated, individually for both the
  afforded hand grasps and regions of interaction, by resizing and pasting at
  the corresponding locations.  For GAO task, we only use the spatially
  aggregated grasp predictions, and for the ROI-prediction task, we only use
  the spatially aggregated ROI prediction.  For evaluation on \RoID, we also
  smooth our predictions using a Gaussian kernel (with standard deviation of 25)
  to suppress high frequencies.  To generate affordances (for example in
  \figref{vis-hotspots}), we simply multiply the two spatial predictions.

\noindent \textbf{Inference on \ycb}.
We do inference over 800 $160 \times 160$ patches to obtain pixel-wise grasp
predictions for each of the 33-hand grasps. We then compute the average score
within each object mask and use that to compute 33 scores, one each for each
grasp type.

 \subsubsection{Region of Interaction (RoI) Baselines}
\noindent \textbf{Mask~RCNN.} 
We use an FPN-based (Feature Pyramid Network) Mask~RCNN model trained on MSCOCO
with a \RS{101} backbone for implementing this baseline. For inference, we
predict 1000 detections per image with an NMS threshold of $0.7$. To get the
RoI prediction, we multiply the class-score with the soft instance segmentation
mask, and paste it at the corresponding detection locations. 

\vspace{2pt}
\renewcommand{\arraystretch}{1.0}
\begin{wraptable}[15]{r}{10cm}
\setlength{\tabcolsep}{8pt}
\centering
\caption{\textbf{Selecting Relevant COCO Categories to Maximize Mask~RCNN
Performance.} We observe that using predictions for microwave, oven, sink,
refrigerator or all four, reduces the performance of Mask~RCNN on validation
set. This is because the region-of-interaction task requires localizing the
regions of interaction on these objects and not segmenting them out as a whole.
Consequently, we remove these 4 object classes from the relevant categories. We
compare to this stronger Mask~RCNN baseline.}
\tablelabel{maskrcnn-ablations}
\resizebox{1.0\linewidth}{!}{
\begin{tabular}{lcc}
\toprule
& \multicolumn{2}{c}{\small{\bf Overall AP}}
\\
\cmidrule(lr){2-3}
Slack at segment boundaries & {0\%} & {1\%}  \\
\midrule
Mask~RCNN [relevant]                                        & \bf 65.7 & 72.1\\
Mask~RCNN [relevant] w/ oven                                & 50.0     & 58.1\\
Mask~RCNN [relevant] w/ microwave                           & 61.7     & \bf 73.1\\
Mask~RCNN [relevant] w/ sink                                & 54.4     & 60.7\\
Mask~RCNN [relevant] w/ refrigerator                        & 52.6     & 57.7\\
Mask~RCNN [relevant] w/ oven, microwave, sink, refrigerator & 47.3     & 53.3\\
\bottomrule
\end{tabular}}
\end{wraptable}
 \noindent \textbf{Mask~RCNN [relevant].} Before pasting the predicted
segmentations, we filter out detections corresponding to the relevant
categories. Specifically, we consider the following object categories that we
selected so as to maximize the AP on the validation set (see ablation in
\tableref{maskrcnn-ablations}): \ttt{ backpack, umbrella, handbag, tie,
suitcase, sports ball, baseball bat, baseball glove, tennis racket, bottle,
wine glass, cup, fork, knife, spoon, bowl, banana, apple, sandwich, orange,
broccoli, carrot, hot dog, pizza, donut, cake, chair, mouse, remote, keyboard,
cell phone, toaster, book, vase, scissors, hair drier,} and \ttt{toothbrush}.

\vspace{5pt}

\vspace{2pt}
\noindent \textbf{Interaction Hotspots.} We use the pre-trained model (with a
dilated \RS{50} backbone) provided by Nagarajan
\etal~\cite{nagarajan2019grounded} to predict interaction hotspots on \RoID.
Specifically, we uniformly sample 800 patches of size $400 \times 400$, resize
to $224 \times 224$ to feed into their model, get $28 \times 28$ predictions
from their model, upsample and paste these predictions at the corresponding
location. We selected the ($400 \times 400$) patch size based on the validation
set performance. We did not observe any improvement in performance on
increasing the number of patches sampled. Note that the model
from~\cite{nagarajan2019grounded} is a action-specific model. We convert their
predictions into per-pixel interaction probability by taking the max score
across actions at each pixel.

\vspace{2pt}
\noindent \textbf{DeepGaze2.} We use the predictions from
DeepGaze2~\cite{deepgazeii} model to compute the AP on the RoI-prediction task.

\vspace{2pt}
\noindent \textbf{SalGAN.} We use the predictions from SalGAN~\cite{salgan}
model to compute the AP on the RoI-prediction task.

\vspace{2pt}
\noindent \textbf{Mask~RCNN + X.} We combine predictions, $P_\text{X}$ from
models (DeepGaze2, Ours) with predictions $P_\text{Mask~RCNN}$ from Mask~RCNN
[relevant] to obtain combined predictions which are denoted as Mask~RCNN +
DeepGaze2 and Mask~RCNN+ACP in the main paper. This is done by a pixel-wise
combination with scalar weights:
\begin{equation}
P^{\text{comb}}_X = w \cdot P_\text{Mask~RCNN} + (1-w) \cdot P_\text{X}
\end{equation}

We set $w$ to $2/3$ when combining with ACP, and to $1/2$ when combining
with DeepGaze2. This scalar weight was obtained through validation on the
validation set. We additionally found it useful to smooth the output from our
model (Gaussian filtering with standard deviation of $25$ pixels, image size
was $1920 \times 1080$). Such blurring wasn't useful for predictions from
DeepGaze2. 

\subsubsection{Grasps Afforded by Objects (GAO) Baselines}
\noindent \textbf{Chance.} As chance performance, we report the fraction of
positive data for each grasp in the dataset. This corresponds to a flat
precision recall plot.
\vspace{2pt}
\\ \noindent \textbf{Supervised Ceiling.} To train the supervised ceiling, we use
the training split with 15 objects and 77 videos. We use a \RS{50} backbone
with a classifier head containing one fully-connected layer, followed by 33
binary classifiers. We train this network by sampling square patches centered
at the object bounding boxes and use a binary cross entropy loss for training.
We also use color jitter, horizontal flips and random crops on the sampled
patches as data augmentation during training. We validate on the validation
split on the held-out 6 objects.

 \subsection{Detailed Results, Ablations, and Visualizations}

\subsubsection{ACP Ablations}
We study the effect of the different choices regarding supervision, data
preparation and network input, and network architecture, made in the design of
ACP. We conduct these experiments on the validation sets and report performance
on the ROI task and the GAO task (where applicable). For the ROI task, we
report the performance in isolation, and upon combination with Mask~RCNN.
Results are presented in \tableref{affordances-ablations}. 

\noindent
\textbf{Data preparation and network input.} Our full model masks out the hand
before feeding in patches to the network for training, and uses an asymmetrical
context window around the masked region. Furthermore, we only make predictions
for objects and hands when they are in contact with the hand. We ablate these
choices, and find that all three of these choices contribute to the performance
of the full ACP model.

\noindent
\textbf{Supervision and data sampling.} Our full model uses the regions for
both the hand and the object as target and for sampling data during training.
We see a large drop in performance on the ROI task when not using the object
regions for data sampling or as target (denoted as `no object'). Not
using the hand regions for data sampling or as targets (denoted as `no hand')
leads to a small drop in performance for the ROI task but additionally renders
it impossible to train for the GAO task. The role of hands is further
emphasized when we switch to using hand segmentation masks rather than box
masks (as used in all other experiments). Richer understanding of the hands
leads to improved performance on the ROI task. 

\noindent
\textbf{Network architecture.} Our ACP model as used in the main paper takes in
a $2s\times 2s$ input and produces a $s\times s$ output. We also experimented
with a symmetric architecture ($2s\times 2s$ input and output). This can lead
to better spatial alignment and ease learning. We report metrics with two such
architectures, (i) where we put the loss on the bottom center patch, and (ii)
where we put the loss in the entire output window. We observe slight
improvements in performance from these architectural modifications. \\ \\

\renewcommand{\arraystretch}{1.1}
\begin{table}[!h]
	\setlength{\tabcolsep}{12pt}
	\centering
	\caption{\textbf{Variations of ACP}. Average precision for
		Region-of-Interaction prediction and mean average precision for GAO task, each
		on the respective validation sets. For ROI prediction task, we report
    results using raw ACP predictions as well as when combined with Mask~RCNN.
    We train each ablation three times and report the mean and the standard
    deviation ($\mu$\std$\sigma$).}
	\tablelabel{affordances-ablations}
	\resizebox{\linewidth}{!}{
		\begin{tabular}{lccccc}
			\toprule
			& \multicolumn{2}{c}{\small{\bf ROI (Overall AP)}}
			& \multicolumn{2}{c}{\small{\bf ROI (Overall AP) [+Mask~RCNN]}}
			& \small{\bf GAO (mAP)}
			\\
			\cmidrule(lr){2-3}
			\cmidrule(lr){4-5}
Methods                                 & {0\% Slack}  & {1\% Slack}  & {0\% Slack} & {1\% Slack} & \\
			\midrule
			ACP (full model)                             & 61.4 \std 0.3  & 73.3 \std 0.5  & 70.9 \std 0.1 & 79.5 \std 0.2 & 42.2 \std 2.6\\
			\midrule
			\textit{Data preparation and network input}\\
			$\;$ ACP (no hand hiding)                    & 60.8 \std 0.3  & 72.4 \std 0.4  & 70.3 \std 0.1 & 78.8 \std 0.1 & 42.0 \std 5.2 \\
			$\;$ ACP (no contact filtering)              & 59.9 \std 0.7  & 71.4 \std 0.8  & 70.5 \std 0.2 & 79.0 \std 0.3 & 43.0 \std 1.2 \\
			$\;$ ACP (symmetric context)                 & 60.2 \std 0.2  & 72.4 \std 0.2  & 70.0 \std 0.2 & 78.5 \std 0.1 & 39.1 \std 1.0 \\
			\midrule
			\textit{Supervision and data sampling}\\
			$\;$ ACP (no object)              & 53.6 \std 1.0  & 65.1 \std 1.3  & 69.5 \std 0.2 & 77.3 \std 0.3 & 41.4 \std 5.9 \\
			$\;$ ACP (no hand)                & 60.8 \std 0.8  & 72.8 \std 0.7  & 70.7 \std 0.3 & 79.3 \std 0.2 & N/A \\
			$\;$ ACP (hand segmentation masks as opposed to box-masks) & 62.1 \std 0.5  & 74.0 \std 0.4  & 71.1 \std 0.4 & 79.7 \std 0.4 & 42.5 \std 2.7 \\
			\midrule
			\textit{Network architecture}\\
			$\;$ ACP ($2s\times2s$ output, loss everywhere) & 61.5 \std 0.4  & 73.7 \std 0.5  & 70.6 \std 0.2 & 79.2 \std 0.1 & 40.7 \std 2.6 \\
			$\;$ ACP ($2s\times2s$ output, loss on bottom center) & 61.7 \std 0.3  & 73.4 \std 0.1  & 71.1 \std 0.2 & 79.7 \std 0.2 & 41.6 \std 5.0 \\
			
			\bottomrule 
	\end{tabular}}
	
\end{table}

\subsubsection{GAO Category-wise Performance}
The test set only contains 7 (out of 33) grasps, we report the mean
average precision over these 7 categories.  The class-wise performance for ACP
is shown in \tableref{gao-grasp-fixed}. We also report the chance performance
along with a supervised ceiling. \\ \\

\renewcommand{\arraystretch}{1.1}
\begin{table}[!h]
\setlength{\tabcolsep}{10pt}
\centering
\caption{{\bf Class-wise performance on the GAO test set.} We report average
precision for each of the 7 hand grasp type contained in the test set. For ACP
(no $\Lt$) and ACP, we conducted the experiment three times and report the
mean performance. We also report the standard deviation over three runs for ACP and ACP (no $\Lt$).}
\tablelabel{gao-grasp-fixed}
\resizebox{1.0\linewidth}{!}{
\begin{tabular}{lcccg}
\toprule
\textbf{Grasp Type}     
& \textbf{Chance} 
& \textbf{ACP} (no $\Lt$) {\it{[Ours]}} 
& \textbf{ACP {\it{[Ours]}}} 
& \textbf{Supervised Ceiling} \\
\midrule
Large Diameter     & 55.6 & 56.3 \std 5.7 & 45.2 \std 3.6  & 80.2\\
Medium Wrap        & 27.8 & 26.6 \std 5.8 & 20.4 \std 1.5  & 67.2\\
Power Sphere       & 27.8 & 23.6 \std 0.7 & 36.0 \std 3.6 & 68.4\\
Precision Disk     & 22.2 & 15.1 \std 0.8 & 14.9 \std 1.5 & 94.7\\
Parallel Extension & 11.1 & 11.0 \std 0.4 & 28.1 \std 5.6 & 14.8\\
Sphere 4 Finger    & 50.0 & 68.8 \std 6.9 & 64.6 \std 1.9 & 56.6\\
Sphere 3 Finger    & 16.7 & 39.3 \std 6.2 & 57.3 \std 0.2 & 15.4\\
\midrule
Mean~~~~~~~~~~~~~~~~~~~~~~~~~~~~~~~~~~~~~~~~~~~~~~~               
                   & 30.2 & 34.3 \std 0.8 & 38.1 \std 0.2 & 56.8\\
\bottomrule
\end{tabular}}
\end{table}
 
\subsubsection{Qualitative Results}
\begin{enumerate}
\item  We provide additional visualizations for ROI predictions made by ACP on the
validation split of \RoID dataset (see \figref{roi-results}).  We observe that
our method can locate regions that afford interaction such as drawer handles,
knobs and buttons which are not typically annotated in object segmentation
datasets. We also see that predictions are localized to object regions that
afford interaction \eg edges of plates.
\begin{figure}[!h]
\centering
  \insertWL{0.33}{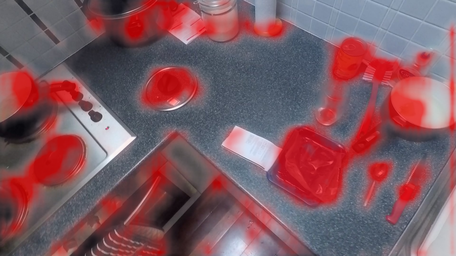}
  \insertWL{0.33}{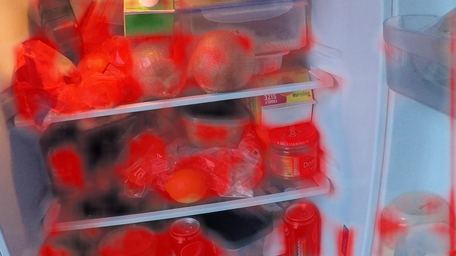}
  \insertWL{0.33}{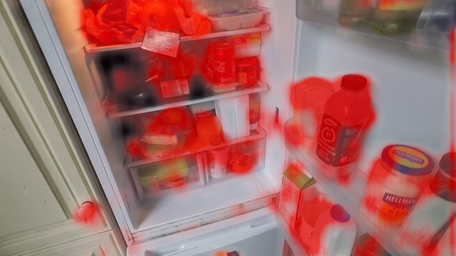}\\
  \insertWL{0.33}{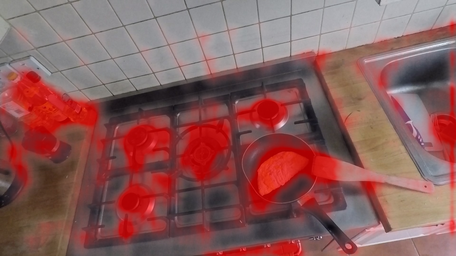}
  \insertWL{0.33}{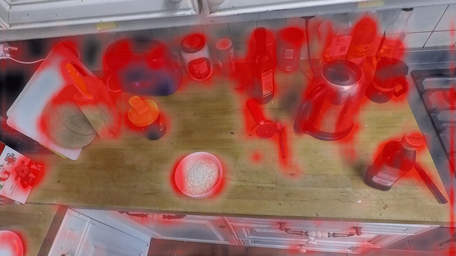}
  \insertWL{0.33}{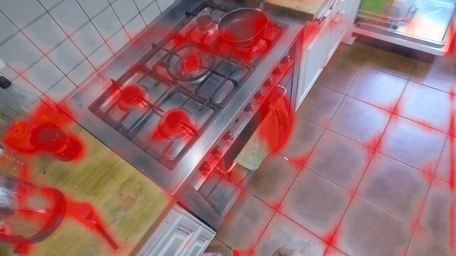}\\
  \insertWL{0.33}{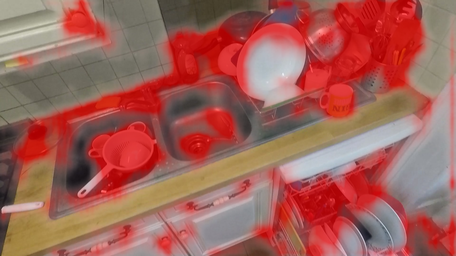}
  \insertWL{0.33}{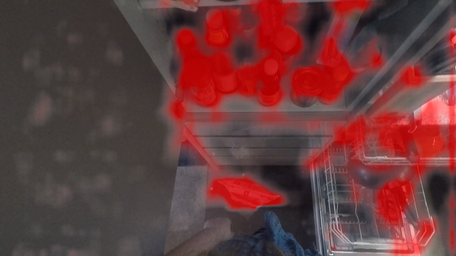}
  \insertWL{0.33}{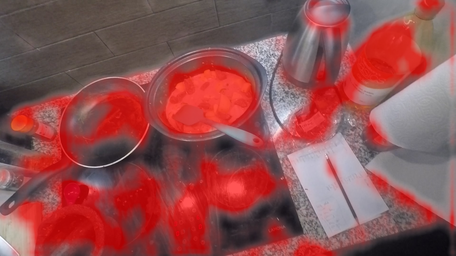}\\
  \insertWL{0.33}{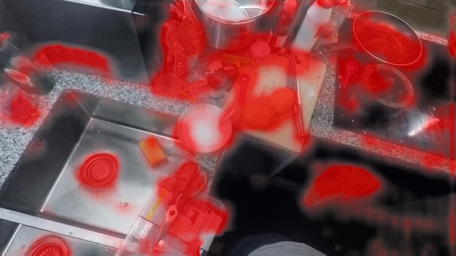}
  \insertWL{0.33}{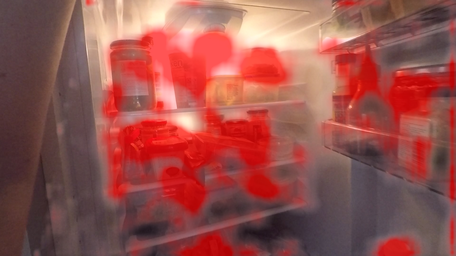}
  \insertWL{0.33}{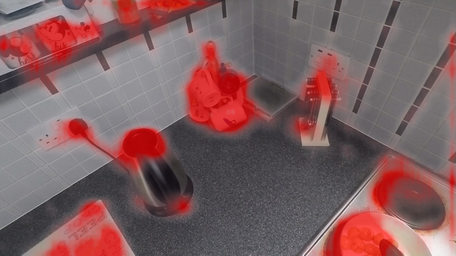}\\
  \insertWL{0.33}{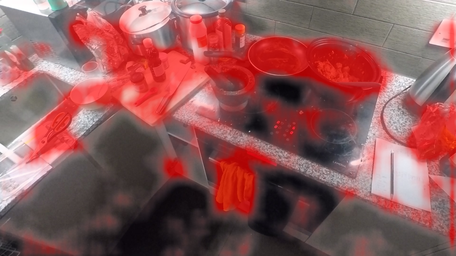}
  \insertWL{0.33}{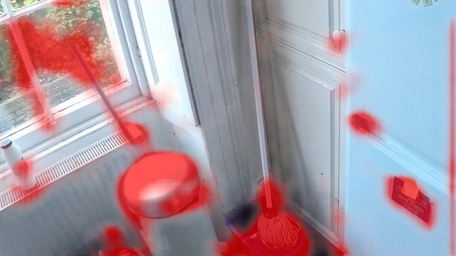}
  \insertWL{0.33}{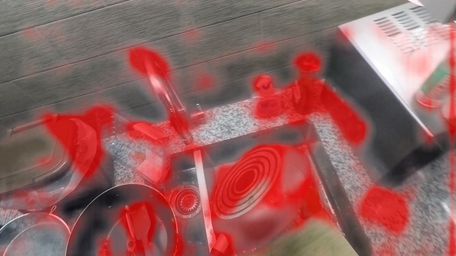}\\
  \insertWL{0.33}{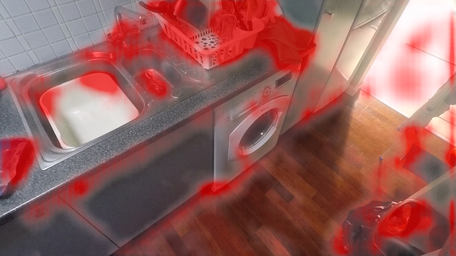}
  \insertWL{0.33}{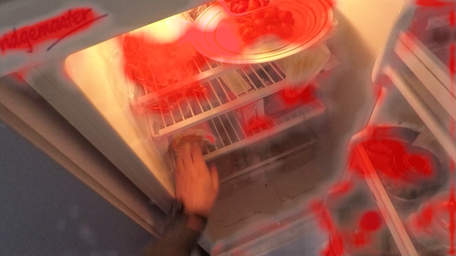}
  \insertWL{0.33}{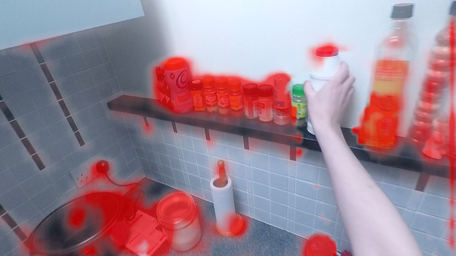}\\
\caption{{\bf Regions-of-Interaction (ROI) predictions on \RoID dataset.} We
show ROI predictions on 18 images from the validation dataset. We observe that
our method can locate regions that afford interaction: drawer handles, knobs
and buttons (not typically annotated in object segmentation datasets). 
We also see that predictions are localized to object regions that
afford interaction \eg edges of plates.}
\figlabel{roi-results}
\end{figure}
 
\item We also visualize predictions for afforded grasps on the \epic dataset.
We convert predicted grasp-specific heatmaps into detections (by finding {\it
scale-space blobs} in the heatmaps) and visualize the top scoring detections
across the validation dataset in \figref{obj-aff-retrievals}.

\newcommand{\insertGAO}[3]{
  \rotatebox[origin=l]{90}{\centering \parbox{1.5cm}{\tiny #3}}  &
\includegraphics[width=0.105\textwidth]{supp/figures/aff_vis/grasps/#2.png} &
	\includegraphics[width=0.105\textwidth]{supp/figures/aff_vis/#1/001_im.jpg} &
	\includegraphics[width=0.105\textwidth]{supp/figures/aff_vis/#1/002_im.jpg} &
	\includegraphics[width=0.105\textwidth]{supp/figures/aff_vis/#1/003_im.jpg} &
	\includegraphics[width=0.105\textwidth]{supp/figures/aff_vis/#1/004_im.jpg} &
	\includegraphics[width=0.105\textwidth]{supp/figures/aff_vis/#1/005_im.jpg} &
	\includegraphics[width=0.105\textwidth]{supp/figures/aff_vis/#1/006_im.jpg} &
	\includegraphics[width=0.105\textwidth]{supp/figures/aff_vis/#1/007_im.jpg} &
	\includegraphics[width=0.105\textwidth]{supp/figures/aff_vis/#1/008_im.jpg} 
}

\begin{figure*}
	\setlength{\tabcolsep}{1pt}
	\centering
\begin{tabular}{rc|cccccccc}
	\insertGAO{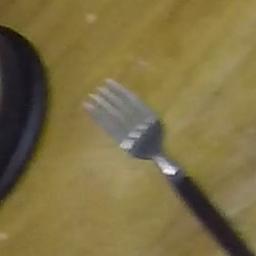}{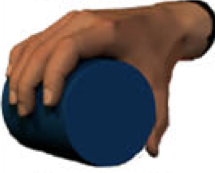}{Large Diameter} \\
	\insertGAO{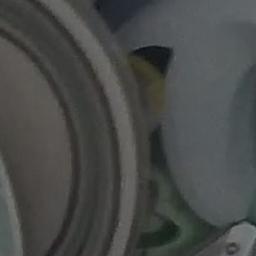}{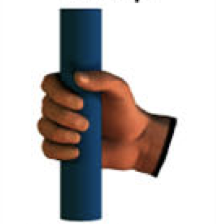}{Medium Wrap} \\
	\insertGAO{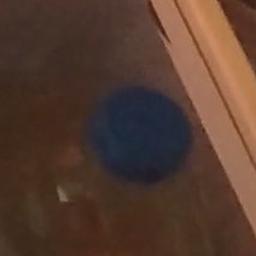}{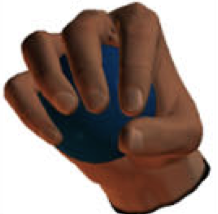}{Power Sphere} \\
	\insertGAO{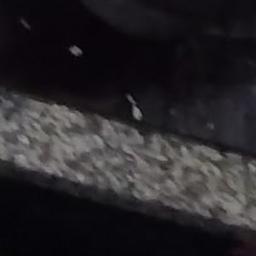}{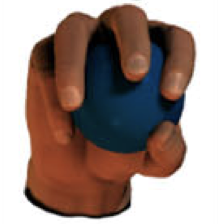}{Sphere 4 Finger} \\
	\insertGAO{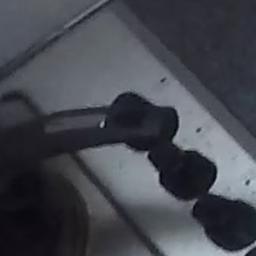}{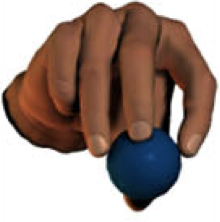}{Quadpod} \\
	\insertGAO{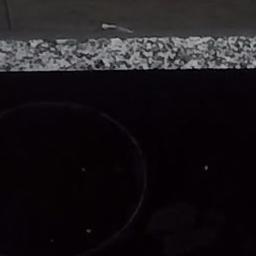}{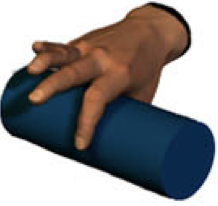}{Ring} \\
	\insertGAO{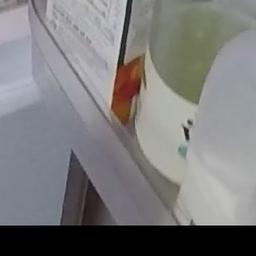}{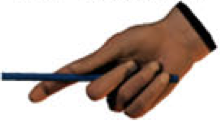}{Ventral} \\
	\insertGAO{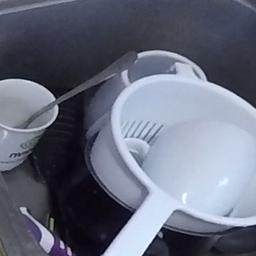}{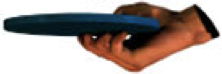}{Extension Type} \\
	\insertGAO{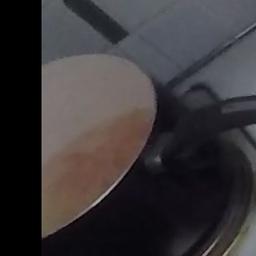}{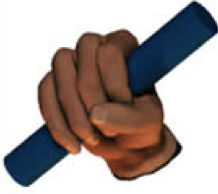}{Small Diameter} \\
	\insertGAO{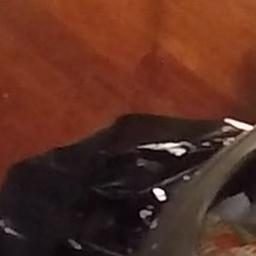}{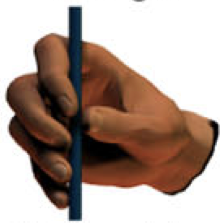}{Prismatic 4 Finger} \\
	\end{tabular}
  \caption{\textbf{Visualizations for Afforded Grasps.} Top detections for
  selected hand grasp types on the validation split of EPIC-ROI dataset. We
  convert predicted grasp-specific heatmaps into detections (by finding
  scale-space blobs in the heatmaps) and visualize the top scoring detections
  across the validation dataset. Many of these detections are plausible, \eg
  lid handles for power sphere, and sphere 4 finger grasps; bottle
  caps and stove knobs for quadpod grasp.}
	\figlabel{obj-aff-retrievals}
\end{figure*}
 
\item In \figref{ycb-res}, we show the top detections for each of the 7 grasps made
by ACP on the validation set of GAO benchmark. The images are colored green if
the corresponding grasp is applicable to the highlighted object or else colored
in red. We observe that for many hand-grasp types, the top scoring objects
actually afford the corresponding hand-grasp type.
\begin{figure*}[!h]
\centering
\insertWL{1.0}{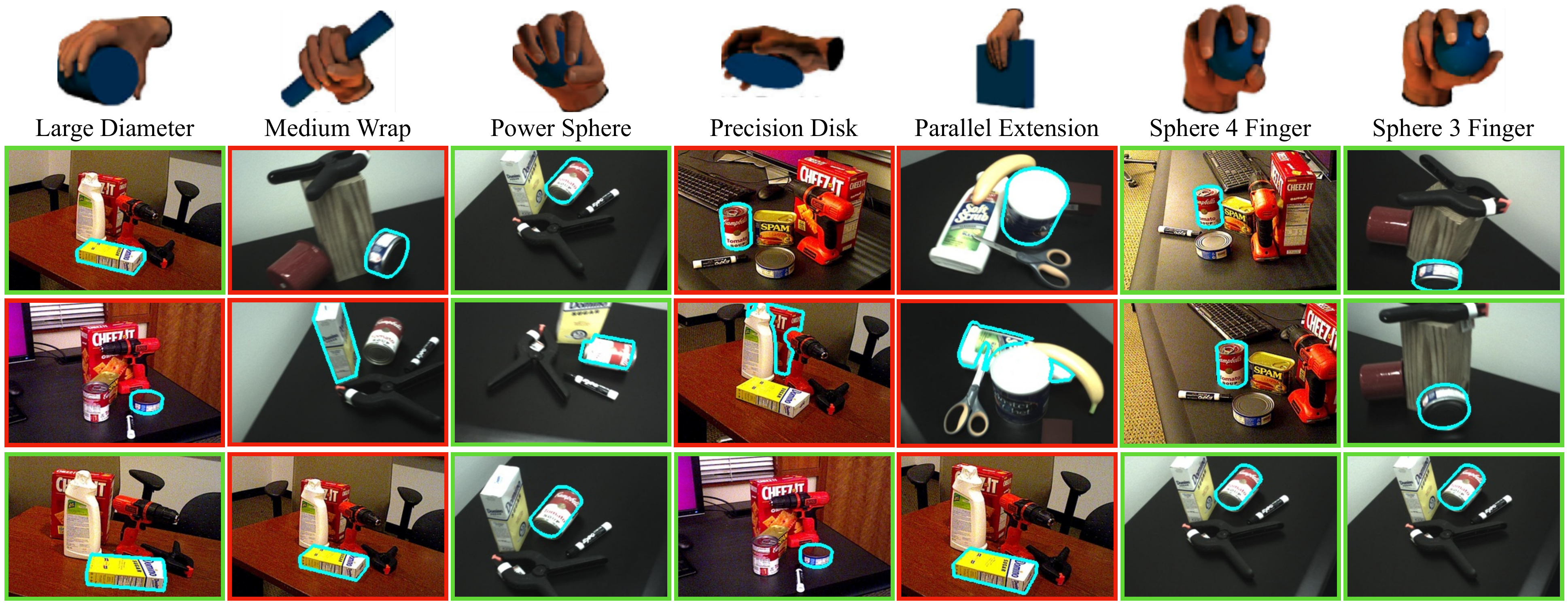}
\caption{{\bf Visualizations of predictions for GAO task on \ycb dataset.} Here
we show the top predictions (object-wise) made by ACP for the 7 grasps
contained in the validation set. For each of the 7 grasps, we visualize the
grasp (reproduced from~\cite{feix2015grasp}) in the top row, and show the top
three predictions (after removing images that were very similar). We highlight
the object for which the prediction is being made for (in cyan). We color the
image frame to indicate correctness of prediction based on ground truth from
Corona \etal~\cite{corona2020ganhand} (green indicates correct, red indicates
incorrect).}\figlabel{ycb-res} 
\end{figure*}
 
\end{enumerate}

\end{document}